\documentclass[letterpaper]{article} %
\usepackage[preprint]{aaai2027}  %
\usepackage[hyphens]{url}  %
\usepackage{graphicx} %
\usepackage{natbib}  %
\usepackage{caption} %
\usepackage{algorithm}
\usepackage{algorithmic}
\usepackage{amsmath}
\usepackage{tcolorbox}
\usepackage{amssymb}
\usepackage{array}
\usepackage{enumitem}
\usepackage{newfloat}
\usepackage{listings}
\DeclareCaptionStyle{ruled}{labelfont=normalfont,labelsep=colon,strut=off} %
\floatstyle{ruled}
\newfloat{listing}{tb}{lst}{}
\floatname{listing}{Listing}

\usepackage{booktabs}

\usepackage{xcolor}

\title{Spontaneous Context Restoration: How Language Models Recover from Corrupted Inputs}

\title{Spontaneous Context Restoration: How Language Models Recover from Corrupted Inputs}
\author {
    Pranjal Garg\textsuperscript{\rm 1},
    Jacob Beck\textsuperscript{\rm 2}
}
\affiliations {
    \textsuperscript{\rm 1}Independent Researcher\\
    \textsuperscript{\rm 2}ML Collective (MLC)\\
    pranjaldun@gmail.com, 
    jacob\_beck@alumni.brown.edu
}

\begin{document}

\maketitle
\begin{abstract}
Language models sometimes produce correct outputs even when their inputs are corrupted by deletion, replacement, or misspelling. We study the internal processes accompanying this behavior, which we call context restoration, in controlled attention-only transformers and five pretrained LLMs (1B--32B parameters) across arithmetic, reading comprehension, and multiple-choice reasoning tasks. In the attention-only transformers, restoration emerges spontaneously despite training exclusively on clean sequences, without corruption training or an explicit denoising objective. We find that context restoration follows a two-phase process: early layers localize effects associated with repair at corrupted positions, while later layers accumulate these effects at uncorrupted positions through the residual stream and ultimately concentrate them at the output position. Repair outcome is predictable from hidden states: cosine alignment with the clean state is highly predictive in attention-only models, while linear probes recover additional information in pretrained LLMs. A linear probe using only the corrupted prompt's first-block hidden state predicts failure with mean ROC-AUC 0.78. This enables failure triage under matched or even partially shifted deployment conditions and may reduce unnecessary verification or computation. Failed examples also show substantially greater nonlinearity along corruption directions. Moderate-corruption finetuning increases corruption tolerance while simultaneously reducing displacement-normalized linearization error, associating improved robustness with a more nearly linear response to corruption.
\end{abstract}

\section{Introduction}
When an input is corrupted, a key word is deleted, a number in a sequence is replaced, or a sentence is interrupted by irrelevant text, a language model sometimes behaves as if it had seen the uncorrupted input. It may still produce the same answer as it would have produced from the original prompt. Whether accuracy survives the perturbation, measured at the output alone, does not capture what is happening here: the model must preserve the task-relevant computation despite receiving damaged input, and matching outputs alone do not show whether it did. This creates an oversight problem: when robustness fails, the failure may be silent, so a deployed system needs a way to identify inputs at elevated risk before acting on the model's output. This matters in practice because real inputs are often affected by OCR errors, speech-transcription errors, formatting artifacts, tool-call failures, copy-paste noise, and adversarial edits~\citep{singh2024robustness, wang2024comprehensive, mccoy2019right, petroni2020context, zhou2024robust}. Understanding how models preserve correct behavior under such corruption can help distinguish deployed systems that are genuinely robust from those that will fail silently under distributional shift and inform training procedures that promote reliable recovery~\citep{wang2022measure}. A mechanism that estimates failure risk before generation could support selective clarification or verification rather than treating every corrupted input as unreliable.

We call this phenomenon \emph{spontaneous context restoration}: the preservation of correct task-relevant computation under label-preserving corruption of input content. In our attention-only transformer models, this behavior emerges despite the model being trained exclusively on clean sequences and never trained to recover from corrupted inputs. We trace \emph{where} corrupted information first matters inside the network, \emph{how} effects associated with repair move across positions, and \emph{what geometric properties} of internal representations separate examples that restore from examples that fail. We study controlled attention-only transformers (4- and 10-layer) trained on arithmetic sequences alongside five pretrained instruction-tuned LLMs spanning 1B--32B parameters, evaluated on ARC-Challenge ~\citep{allenai:arc}, factual QA ~\citep{rajpurkar2016squad}, and CommonsenseQA ~\citep{talmor-etal-2019-commonsenseqa} under four corruption modes (dropout, replacement, spelling, distractor injection).
 
Our central finding is that behaviorally context restoration follows a \textbf{two-phase
spatial process} shared between architectures and tasks: 
early layers localize the influence of replacing corrupted-run activations with their clean counterparts at corrupted positions, while with depth this effect shifts toward clean context positions and ultimately concentrates at the prediction position. Ablating attention heads leaves repair largely intact, indicating that it draws on the accumulated residual stream rather than any single specialized head. The repair outcome is predicted by the \emph{geometry} of the residual stream: in controlled attention-only transformer models, the cosine similarity between the corrupted and the clean states reaches AUC 0.80--0.95 at fixed corruption severity, while in pretrained models linear probes recover features invisible to directional metrics alone. Using only the corrupted prompt's first-block hidden state, a linear probe achieves a mean AUC of 0.78 across models, tasks, corruption modes, and corruption levels. Flagging the 10\% of inputs assigned the highest failure risk yields a mean $2.6\times$ recall lift over random selection, allowing selective clarification or verification before full generation. A linearization analysis (i.e. how closely a first-order Taylor approximation predicts the true effect of corruption on the output) shows that failed examples exhibit greater nonlinearity along corruption directions than repaired examples. Consistent with this account, moderate-corruption finetuning in the controlled model reduces curvature along corruption directions and the corruption level tolerated at 50\% accuracy increases from 40\% to 90\% without measurable loss of clean accuracy.

\begin{figure}[htbp!]
    \centering
    \includegraphics[width=0.95\columnwidth]
    {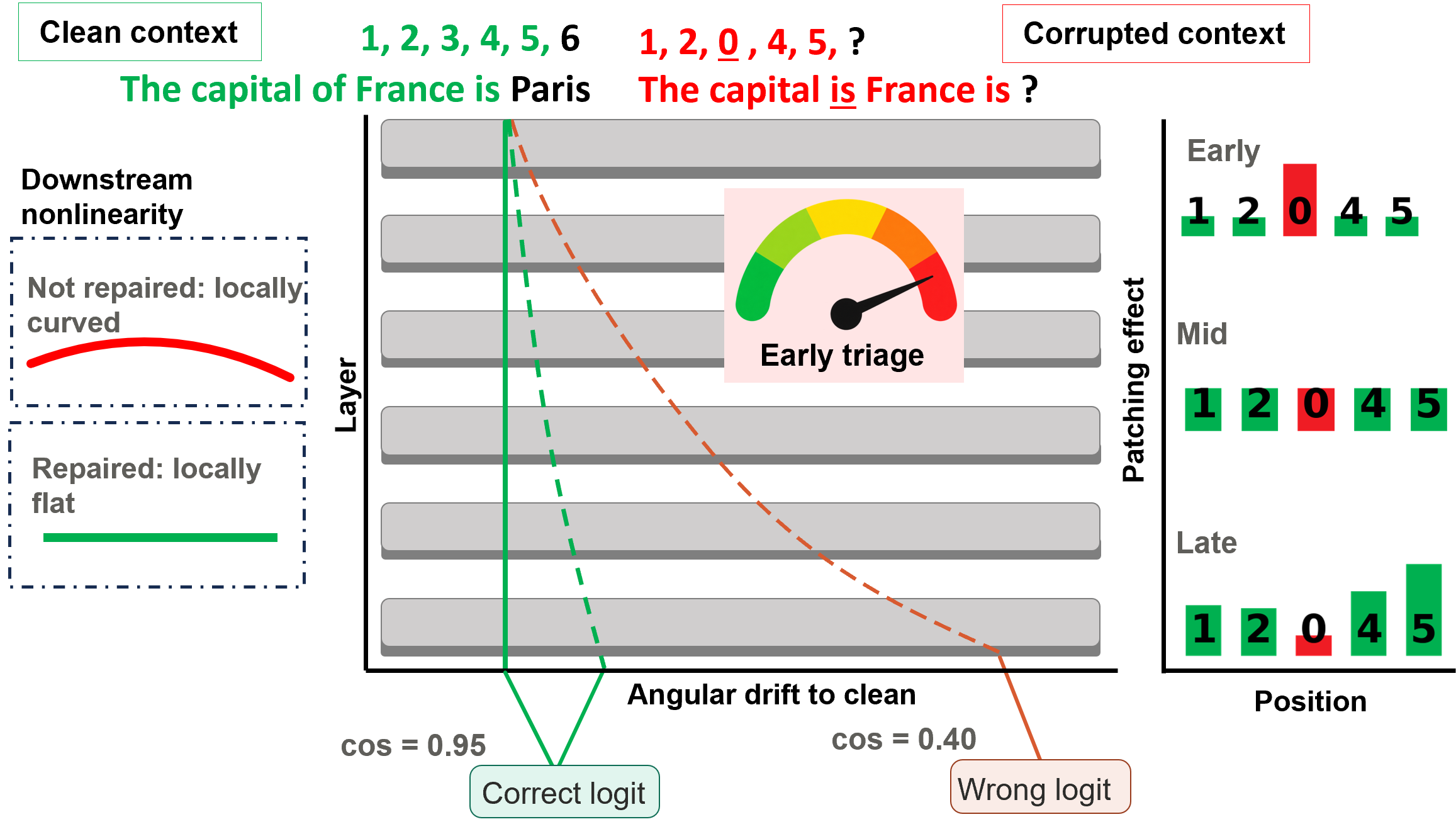}
    \caption{Overview of spontaneous context restoration. Patching effects shift across positions with depth, hidden states distinguish repaired from failed inputs, and early representations support failure triage before generation.}
    \label{fig:overview}
\end{figure}

\textbf{Our contributions are}: (1) we show that context restoration emerges spontaneously in models trained only on clean sequences and characterize its two-phase spatial organization across attention-only and pretrained transformers, localizing where corrupted information matters and how effects associated with repair shift to the prediction position; (2) we show that repair outcome is predictable from residual-stream geometry and linearly decodable from the corrupted prompt's early hidden state before output generation, enabling selective verification of inputs most likely to fail rather than treating all corrupted inputs as unreliable; (3) we show that failed examples exhibit greater nonlinearity along corruption directions, and that corruption finetuning jointly increases robustness and reduces linearization error.

\section{Related Work}

\paragraph{Context robustness and corruption.}
\citet{liu2024lost} show that model performance degrades when relevant evidence is positioned in the middle of long contexts. \citet{min2022rethinking} decompose prompts into components and find that some can be semantically corrupted without destroying performance. \citet{abedin2025arithmattack} show that punctuation-only corruption, which preserves all lexical content, still degrades math reasoning. Noisy-rationale and prompt-perturbation studies~\citep{zhou2024can,anantheswaran2024cutting, chatziveroglou2025exploring} report smooth degradation with noise and partial mitigation via prompting or finetuning. A separate line examines context faithfulness versus parametric priors: ~\citet{neeman2023disentqa} formalize persuasion and susceptibility metrics, and ~\citet{ortu2024competition} trace how factual recall and in-context counterfactual repetition compete internally. All of these are primarily behavioral; they do not characterize the internal processes associated with successful performance under corruption.

\paragraph{Mechanistic interpretability and signal routing.}
\citet{elhage2021mathematical} frame the residual stream as a shared communication channel; this view underpins our patching analyses. Circuit analyses for IOI~\citep{wang2022interpretability}, greater-than~\citep{hanna2023does}, and arithmetic~\citep{stolfo2023mechanistic,yu2024interpreting,zhang2024interpreting} localize specific behaviours to compact sets of heads and MLPs, but assume clean inputs. Activation patching~\citep{meng2022locating, goldowsky2304localizing, heimersheim2024use,zhang2023towards} and sparse autoencoders~\citep{bricken2023towards,olah2020zoom} provide the causal and feature-level tools we use. \citet{lad2024remarkable} report that deleting or swapping middle layers preserves 72--95\% of predictions and hypothesise four stages of inference; this is consistent with our finding that middle layers are where effects associated with repair redistribute.

\paragraph{Self-repair and compensation.}
The Hydra Effect~\citep{mcgrath2023hydra} shows that ablating one layer induces compensation elsewhere. \citet{rushing2024explorations} expand this across model families, identifying LayerNorm scaling and anti-erasure neurons as partial mechanisms. These studies examine compensation after \emph{internal} ablation (removed heads or layers). We study recovery from \emph{external}, label-preserving input corruption as a related but distinct problem.

\paragraph{Depth-wise readout and representation geometry.}
The tuned lens~\citep{belrose2023eliciting} decodes every layer in vocabulary space, treating the transformer computation as an iterative refinement. \citet{marks2023geometry} and \citet{shai2024transformers} show that task-relevant information can form linear structures in the residual-stream. \citet{engels2024not} show that some features occupy subspaces of higher dimensions rather than single directions. Our linearization analysis measures functional curvature along the corruption direction, linking representation geometry to repair outcome.

\paragraph{Finetuning and robustness to corruption}
Training on noisy or perturbed inputs has been shown to improve robustness across NLP models and LLMs~\citep{gupta2024evaluating,alajrami2025fine,liu2026improving,zhang2026gsm}. Our work differs by varying the severity of corruption during finetuning and relating the resulting robustness gains to changes in linearization error along corruption directions. We also compare the setting in which models are trained only on clean inputs with one in which they are further finetuned on corrupted inputs.

\section{Experimental Setup}
\label{sec:setup}

We study \emph{spontaneous context restoration}: the ability of transformer models to preserve the intended task-relevant computation when part of the input context is corrupted, without being explicitly instructed to denoise it. Given a clean input $x$, a label-preserving corrupted input $\tilde{x}=c_r(x)$, and target output $y$, we ask when $f_\theta(\tilde{x})=y$ and what internal representations distinguish successful restoration from failure. We examine this in two complementary settings described below. For full details see Supplementary Content 1 \& 2.

\begin{table}[t]
\centering
\small
\begin{tabular}{ll}
\toprule
\textbf{Task family} & \textbf{Example sequence} \\
\midrule
Constant increment & $2,\,4,\,6,\,8,\,10,\ldots$ \\
Constant decrement & $10,\,8,\,6,\,4,\,2,\ldots$ \\
Add--subtract & $2,\,5,\,3,\,6,\,4,\,7,\ldots$ \\
Variable arithmetic & $2,\,4,\,7,\,16,\,22,\,29,\ldots$ \\
\bottomrule
\end{tabular}
\caption{Examples of the arithmetic sequence task families used in the controlled setting.}
\label{tab:arithmetic_examples}
\end{table}

\subsection{Synthetic Symbolic Setting: Controlled Attention-only transformers on arithmetic}
\label{sec:setup_toy}
We generate four task families from four types of arithmetic sequences: constant increment, constant decrement, add-subtract, variable arithmetic, rendered as comma-separated token lists (Table~\ref{tab:arithmetic_examples}). In variable-arithmetic sequences, the gap between successive elements increases by one at each step, unlike the fixed or alternating step sizes used in the other sequence families. Each sequence is of length $L \in \{40,60,80,100,120,140\}$, with values bounded by $S_{\max}=13{,}000$. For the corruption rate $r \in \{10,\ldots,90\}\%$, we corrupt a total of $\lceil r L / 100 \rceil$ positions in each sequence, excluding the prediction target with one of three corruption modes: zero ablation, in-range substitution or out-of-range substitution. All corruptions are label-preserving. 

We train two attention-only decoder-only transformers from scratch in clean sequences (Supplementary Figure 1.1): a \textsc{small} 4-layer model (1.75M parameters) and a \textsc{large} 10-layer model (4.01M parameters), both with 8 heads and a vocabulary of 13{,}008 tokens. Integer token indices are randomly permuted in tokeniser construction, forcing the model to learn the numerical structure from context. The 10-layer model is used for main analyses; the 4-layer model verifies depth-independence and replicates main findings. For each experimental condition, we evaluated 100 held-out disjoint sequences; accuracy is the exact match to the clean target. The architecture and training details are in the Supplementary Content 1 \& 2.

\subsection{Natural-Language Setting: Pretrained language models on NLP tasks}
\label{sec:setup_nlp}

We evaluated five instruction-tuned LLMs: Gemma-3-1B-IT~\citep{gemma3_1b_it_hf}, OLMo-2-7B-Instruct~\citep{olmo2_7b_instruct_hf}, Llama-3-8B-Instruct~\citep{llama3_8b_instruct_hf}, Qwen3-32B~\citep{qwen3_32b_hf}, and Gemma-4-31B-IT~\citep{gemma4_31b_it_hf}, using greedy decoding without system prompt. The task families are ARC-Challenge (multiple-choice science), Factual QA (SQuAD-derived extractive reading comprehension) and CommonsenseQA (multiple-choice commonsense reasoning), with up to 500 examples per task filtered to $\leq$250 tokens.

Each prompt is split into (\texttt{user\_prefix}, \texttt{corruption\_region}, \texttt{user\_suffix}); only the corruption region is modified while supporting tokens (such as "Question", "Answer", "Context", etc.), answer choices, and target answer span within the context passage are protected, ensuring label-preserving corruption. We apply four corruption modes at rates $r \in \{10,30,50,70,90,100\}\%$: \textbf{dropout} (word deletion), \textbf{replacement} (substitution with function words), \textbf{spelling} (internal character shuffling) and \textbf{distractor} (interleaving irrelevant sentences). The corruption rate is defined relative to the token count of the clean corruption region.

Accuracy degrades monotonically with corruption rate in both settings; dropout and replacement are the most destructive, spelling is milder, distractor corruption is weakest, and factual QA is the most fragile task (Supplementary Figure 5.7).

We evaluate using exact match, F1 score (the harmonic mean of token precision and recall after text normalization, taking the maximum  over all the available phrasings of the gold answers) and token agreement (whether clean and corrupted prompts produce the same sequence of generated tokens). We term an instance \emph{repaired} when the corrupted input produces the correct answer (operationalized as exact match, $F1 \geq 0.5$, or token agreement depending on the analysis). When repair occurs in models trained on clean sequences without corruption priors, as in our controlled setting, we call it \emph{spontaneous context restoration} as opposed to context restoration in pretrained models where parametric knowledge may also contribute.

\section{Analysis Methods}
\label{sec:methods}
This section defines the tools applied in both settings. In the controlled attention-only transformer models, we cache activations at two sites per layer: attention output before the skip connection and  after residual addition. In pretrained models, we hook at the output of each transformer block. All analyses read from the last (prediction) token position unless otherwise noted. Full procedural details are in Supplementary Content 1 \& 2.

\paragraph{Activation patching.}
To localize which positions carry repair-relevant information at each layer, we run the corrupted input and replace activations at a subset of positions with their clean counterparts, measuring the resulting change in the target-token logit~\citep{elhage2021mathematical}. In the controlled model, patching is performed at both hook sites; in pretrained models, at the layer output. Positions are partitioned into \emph{corrupted} and \emph{clean} groups, patched separately. Because this requires token-aligned sequences, patching in the pretrained setting uses only replacement and spelling corruptions.

\paragraph{Cosine similarity and repair probes.}
At each layer $\ell$, we compute the cosine similarity $\cos_\ell$ between the corrupted and clean activations at the prediction position before and after residual addition in the controlled model; layer output in pretrained models, measuring the angular alignment with the clean trajectory. To assess whether the repair outcome is predictable from intermediate representations, we train two probes: (1)~a \emph{cosine probe} using $\cos_\ell$ as a single scalar feature and (2)~a \emph{logistic regression linear probe} on the full $d$-dimensional corrupted activation with 5-fold stratified cross-validation. A separate \emph{is-corrupted} probe predicts whether an input is corrupted vs.\ clean, measuring when corruption identity becomes linearly decodable.

\begin{figure}[!htbp]
\centering
\includegraphics[width=0.75\columnwidth]{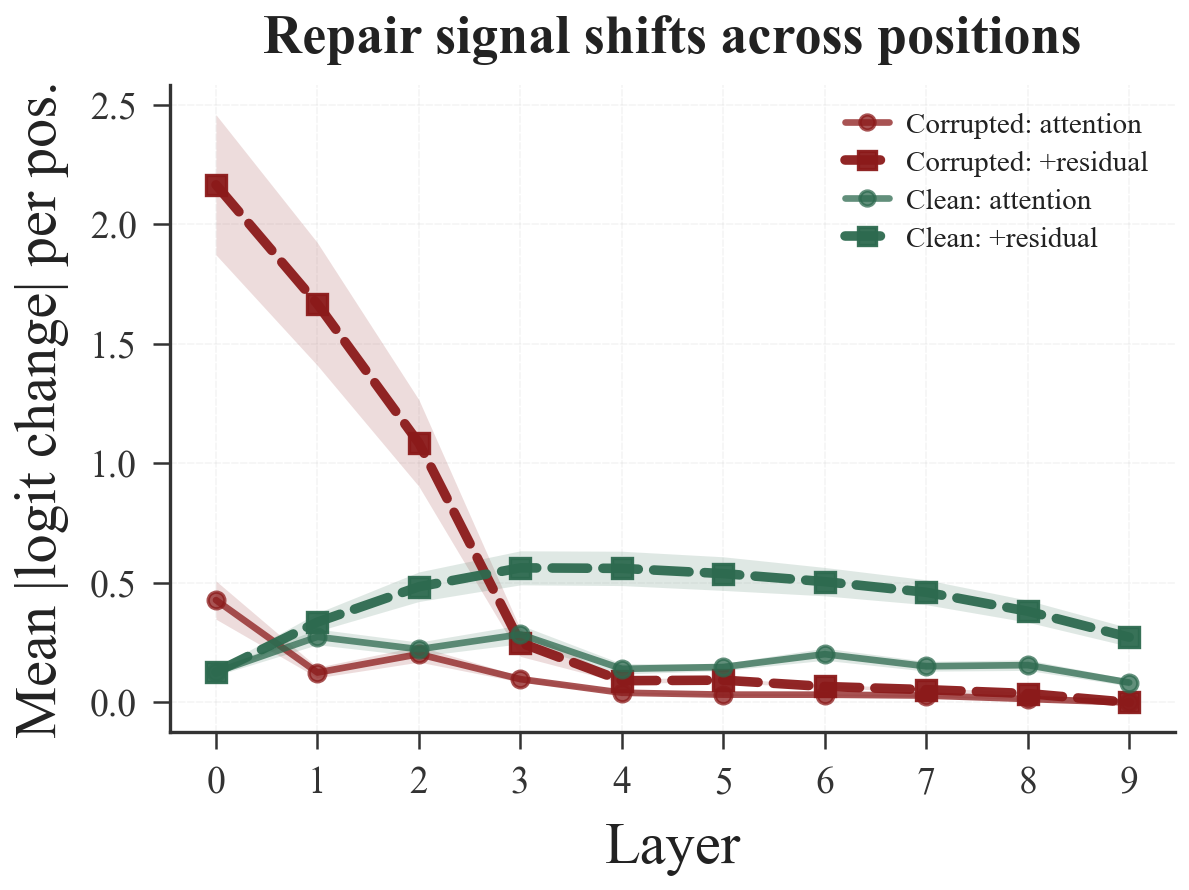}
\par
\includegraphics[width=0.75\columnwidth]{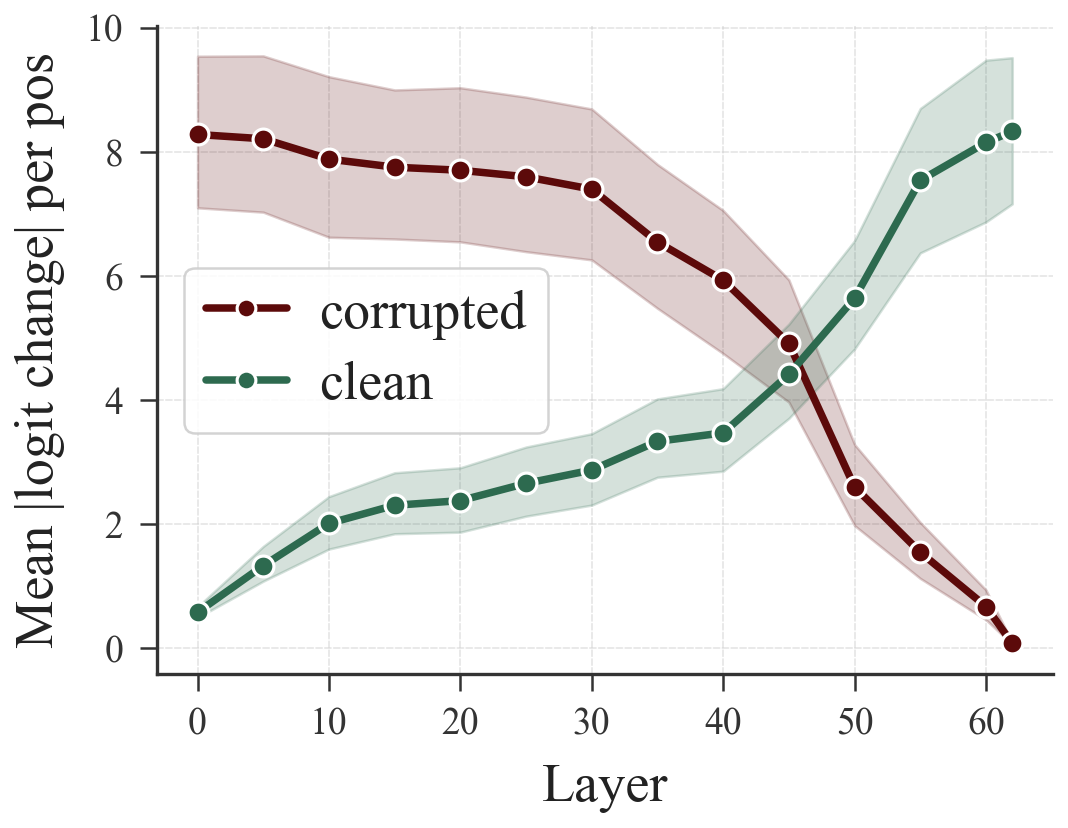}
\caption{Top: Residual stream decomposition and mean patching effect in the controlled (large) model for zero corruption mode. Bottom: Mean per-position patching effect for Qwen3-32B for factual tasks.}
\label{fig:toy_residual_decomposition}
\end{figure}

\paragraph{Linearization test.}
To probe functional curvature along the corruption direction, we define $\mathbf{e} = \mathbf{h}^{\text{corr}}_\ell - \mathbf{h}^{\text{clean}}_\ell$ at the prediction position and compare the true logit change when patching $\mathbf{h}^{\text{clean}}_\ell + \alpha\mathbf{e}$ against the first-order Taylor prediction $\alpha \, \nabla g^\top \mathbf{e}$, for $\alpha \in \{0.25, 0.5, 0.75, 1.0\}$, where $g$ denotes the downstream mapping from the hidden state at layer $\ell$ to the output logit being analyzed. The residual $r(\alpha) \approx \tfrac{1}{2}\alpha^2 |\mathbf{e}^\top H \mathbf{e}|$ quantifies departure from linearity: small residuals indicate the representation lies in a locally linear region of the logit map; large residuals indicate greater departure from locally linear behavior along the corruption direction. The quadratic approximation holds at small $\alpha$; at $\alpha = 1.0$ the residual captures total nonlinearity beyond second order.
\section{Key Findings}
\label{sec:findings}

\subsection{\emph{Where does repair happen?} Repair Signals Shift from Corrupted to Clean Positions}
\label{sec:finding_patching}

Activation patching at individual positions reveals two distinct phases. In the 10-layer model at layer~0, corrupted positions produce approximately $20\times$ larger patching effects than clean positions (Figure~\ref{fig:toy_residual_decomposition},  top); this effect decays sharply with distance downstream of the nearest corrupted tokens. The patching effect is defined as the mean absolute change in the correct-token logit per patched position. By layers~3--4 the pattern reverses: clean positions carry the dominant effect, and corrupted positions approach the zero effect. Very few positions carry the maximum of the effect, and by the late layers, nearly all effects concentrate at the prediction position: the model routes information from distributed positions to clean positions especially at the single readout. Because the readout position is clean by construction, we analyzed it separately from other clean positions. The crossover persists after excluding the prediction position; however, the final prediction position has an effect $17 \times$ larger than the other clean positions  (Supplementary Figure 5.9, 5.10, 5.11). %
This structure replicates in pretrained models: corrupted positions dominate early layers on all three tasks, while clean positions take over later (Figures~\ref{fig:toy_residual_decomposition} bottom, Supplementary Figure 5.12). The depth of the crossover varies by task, but the qualitative pattern is preserved.

Decomposing the patching effect into the current layer's attention output (before residual addition) and the full accumulated state (after residual addition) indicates that effects associated with successful repair accumulate progressively in the residual stream (Figure~\ref{fig:toy_residual_decomposition} top, Supplementary Figure 5.11). 
At corrupted positions in layer~0, the skip connection accounts for most of the effect, showing that the clean patch is carried forward primarily through the residual path rather than being modified by additional attention computation. At clean positions, each layer's attention contributes a roughly uniform increment, while the skip-connection contribution grows steadily from layers~0 through~4.

In the controlled model, ablating 4, 6, or 8 of 8 heads at a single target layer leaves accuracy close to the no-ablation baseline when averaged over target layers (Supplementary Table 4.4), so repair does not depend on any individual layer-local head subset at this scale.

\subsection{\emph{Why does repair fail?} Failures Encounter High Curvature Along Corruption Trajectories}
\begin{figure}[!htbp]
    \centering
    \includegraphics[width=0.75\columnwidth]{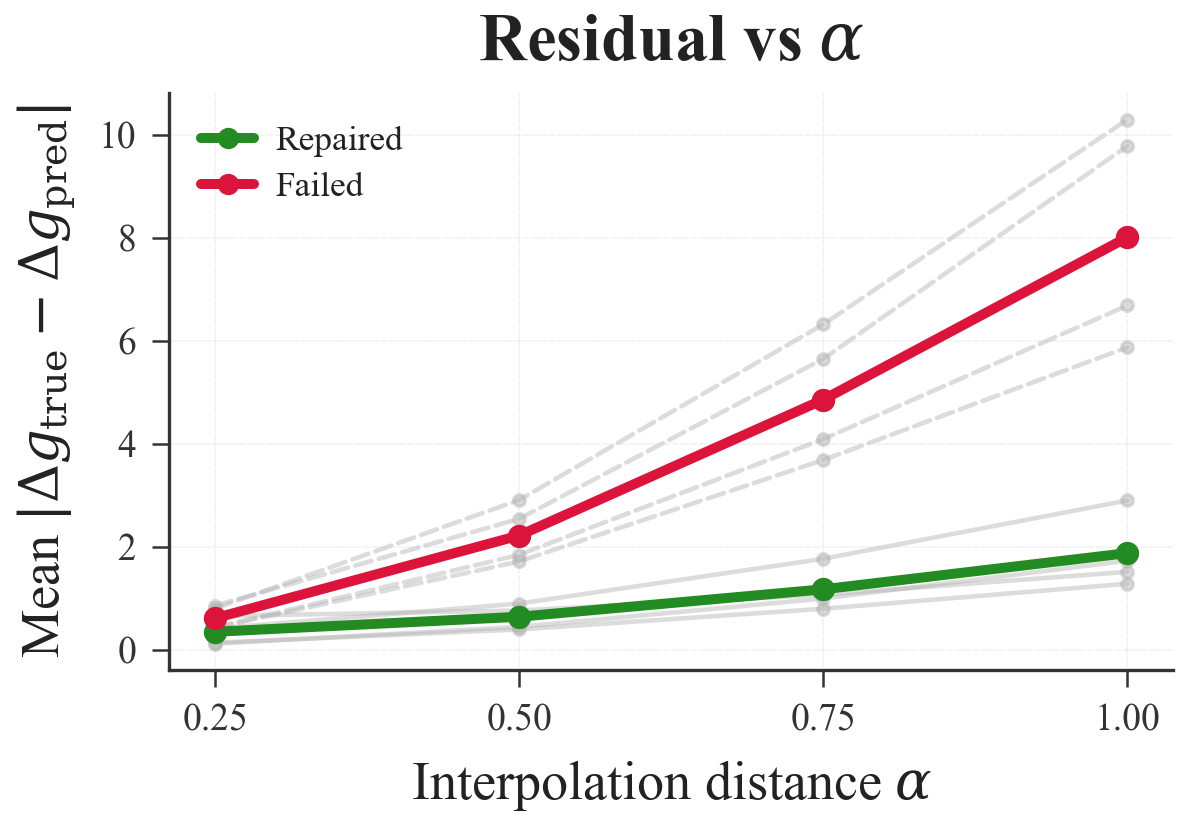}
    \par\vspace{0.2em}
    \includegraphics[width=0.75\columnwidth]{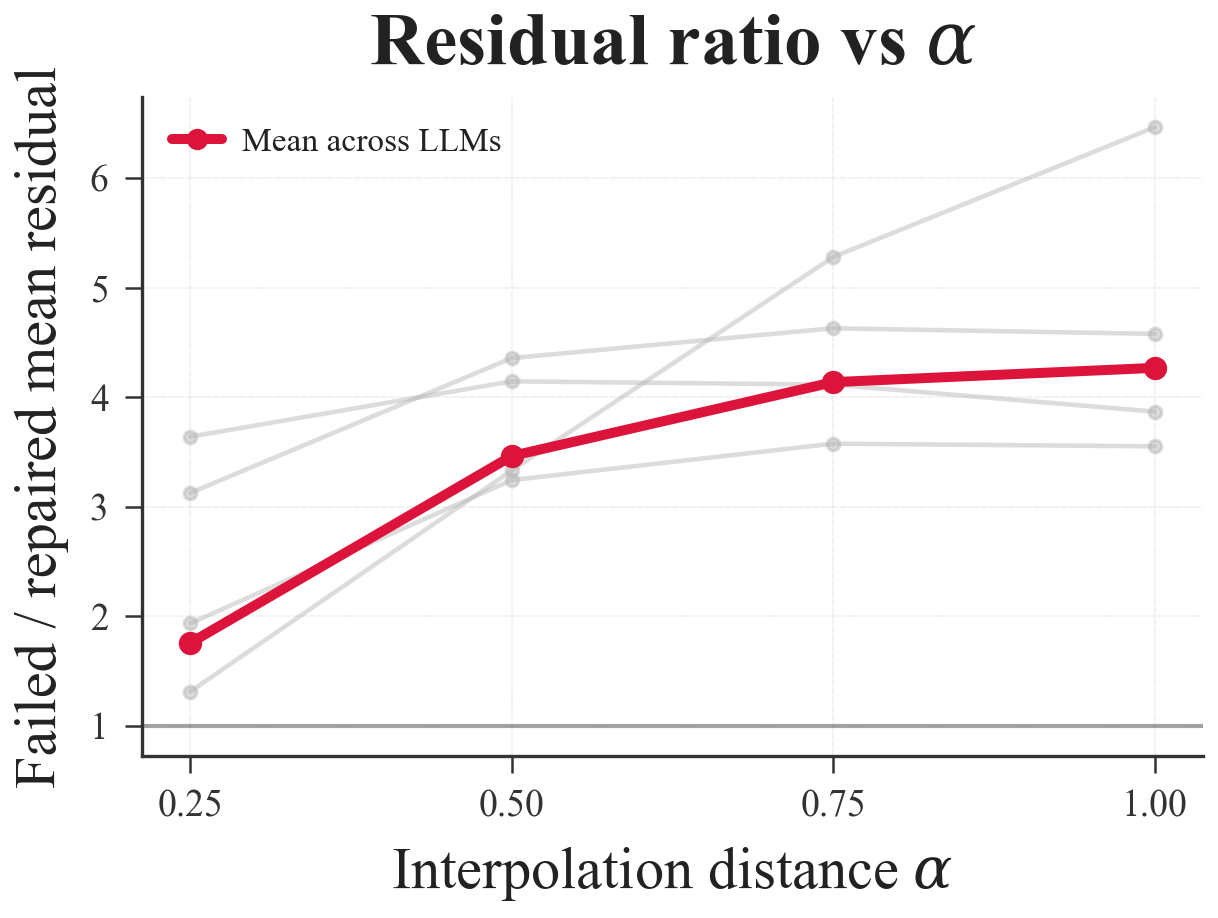}
    \caption{Linearization residuals in the pretrained models. Repaired examples stay close to locally linear behavior near the clean trajectory, while failed examples show much larger residuals at larger interpolation steps.}
    \label{fig:linearization_main}
\end{figure}

\label{sec:finding_linearization}
First-order predictions ($\Delta g_{pred.}$) closely match the true logit change ($\Delta g_{true.}$) at $\alpha = 0.25$ for both repaired and failed examples, indicating that the logit map is locally approximately linear near the clean representation (Figure~\ref{fig:linearization_main} top). At $\alpha = 1.0$, the two groups separate sharply: failed examples show approximately $8\times$ larger linearization residuals than repaired examples. The residual grows with both corruption rate and interpolation distance, producing a trumpet-shaped fan as the examples move farther from the clean trajectory (Supplementary Figure 5.17). This pattern appears in all tested LLMs and controlled models (Supplementary Figure 5.18, 5.19).
At layer~0, the repaired and failed examples are nearly indistinguishable between step sizes. By the final layer, failed examples exhibit approximately $6-8\times$ larger residuals than repaired examples at $\alpha = 1.0$, while the gap remains small at $\alpha = 0.25$. The linearization residual increases with depth for both repaired and failed examples, but substantially faster for failed examples across all four pretrained models (Supplementary Table 4.7).

A natural alternative is that failed examples simply undergo larger or more sensitive displacements rather than entering more curved regions. The interpolation sweep suggests that displacement magnitude or first-order sensitivity alone is insufficient to explain the difference: the ratio of failed to repaired linearization residuals grows with step size, from ${\approx}2\times$ at $\alpha = 0.25$ to ${\approx}4\times$ at $\alpha = 1.0$ (Figure~\ref{fig:linearization_main}). A first-order sensitivity or displacement difference would produce a constant ratio across $\alpha$; curvature produces a higher-order effect, scaling approximately as $\alpha^2$, and therefore predicts this growth. Together, these results associate repair failure with increasing departure from locally linear behavior as corrupted representations move away from the clean state.

\subsection{\emph{What distinguishes repair from failure?} Repair Outcome Is Decodable from Residual-State Representations}

\label{sec:finding_cosine}
\begin{figure}[!t]
    \centering
    {\textbf{Repair success by layer}\par}
    \vspace{0.25em}
    \includegraphics[width=0.7\columnwidth]{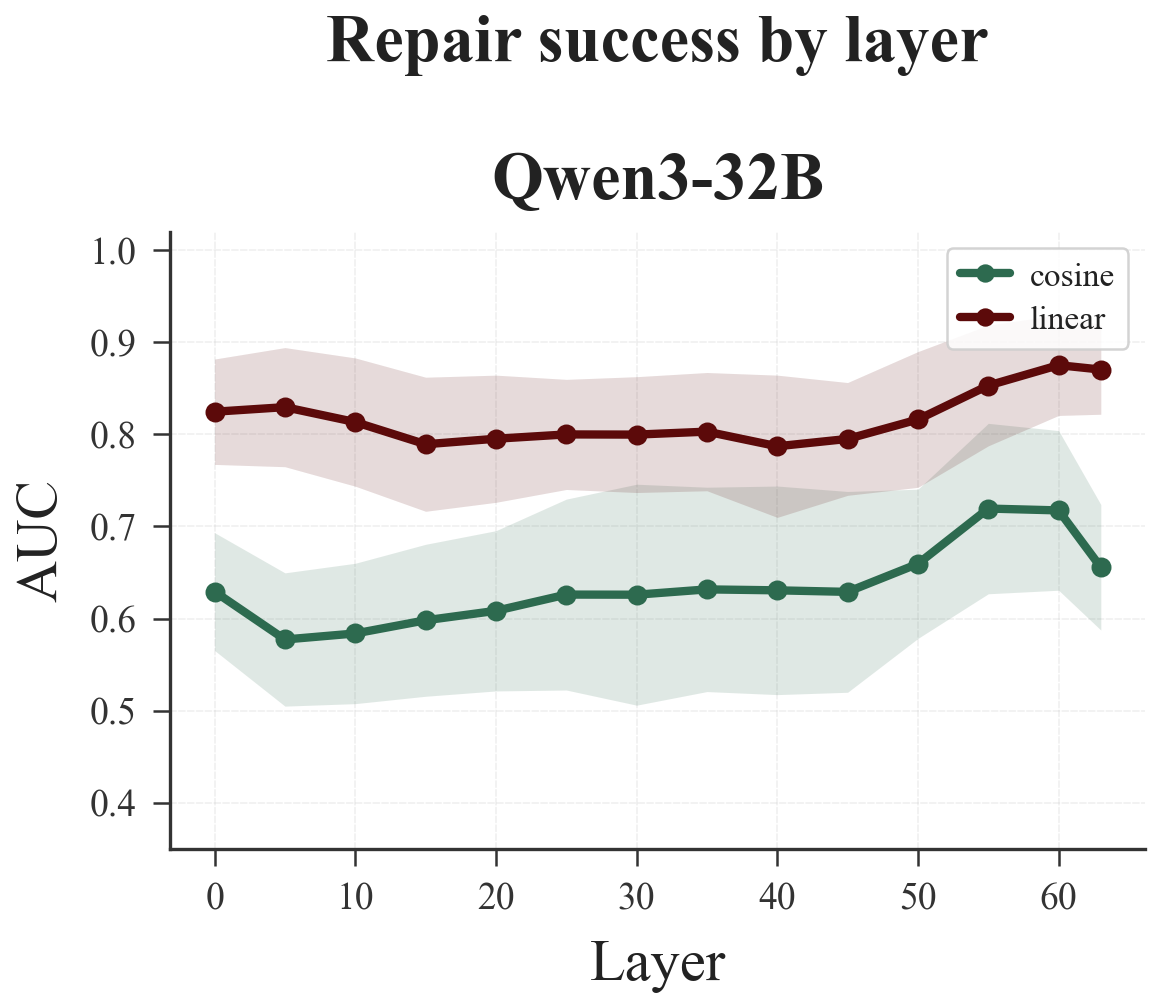}
    \par
    \includegraphics[width=0.7\columnwidth]{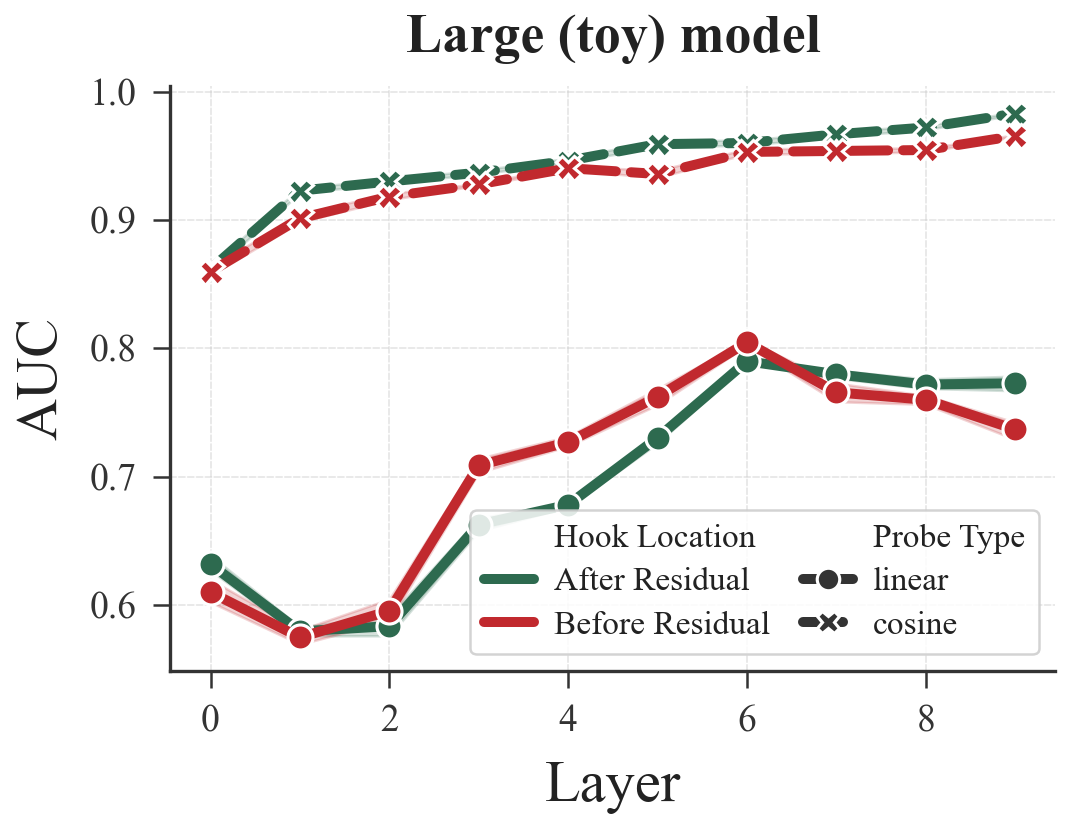}
    \caption{Hidden states predict repair outcome. Top: Qwen3-32B on ARC under replacement corruption. Bottom: controlled model under in-range ablation.}
    \label{fig:repair_probe_comparison}
\end{figure}

The cosine similarity between corrupted and clean activations at the prediction position separates repaired from failed examples. In the 10-layer controlled model, the repaired examples maintain cosine similarity $\cos_\ell \approx 0.95$ across layers, while failed examples diverge to ${\approx} 0.40$ by the final layer (Supplementary Figure 5.13). The gap grows with depth, indicating that successful repair preserves the clean residual direction, while failure corresponds to angular drift away from it. This cosine alignment of activations is predictive of repair at the per-example level. The within-ablation AUC (computed by training and evaluating the probe separately within each corruption rate) for cosine similarity reaches 0.80--0.95 across layers and corruption modes in the controlled setting (Figure~\ref{fig:repair_probe_comparison} bottom). This controls for corruption severity: because more corruption makes repair less likely, a probe trained across rates could predict repair merely by reading off how corrupted the input is. Training within a fixed rate removes this confounding factor, so the above-chance AUC reflects genuine repair-predictive structure rather than severity. See Supplementary Figure 5.14. 

In pretrained models, the pattern becomes task- and corruption-dependent. Cosine similarity probes still reveal late-layer drift for failed examples, especially under destructive corruptions such as dropout and replacement on factual QA. However, for milder corruptions such as distractor insertion and spelling, and for ARC and CommonsenseQA more broadly, correct and failed trajectories often remain close in cosine space. Thus, global angular alignment captures part of the information predictive of repair, but is not sufficient in all settings (Supplementary Figure 5.16).

Corruption detection is near-ceiling (AUC $\approx 0.96-1.00$) from the earliest layers across pretrained models and corruption modes (Supplementary Table 4.6) indicating that the weaker and more task-dependent separation of repaired from failed examples reflects distinct hidden-state information rather than mere corruption detection. Controlled-model corruption-detection results show the same distinction (Supplementary Figure 5.15).

The linear repair-success probe outperforms cosine in pretrained models compared to controlled models  (Figure~\ref{fig:repair_probe_comparison}). 
This suggests that pretrained models contain information predictive of repair in feature subspaces beyond global angular alignment. The controlled model therefore isolates a cleaner geometric repair signature compared to pretrained models. This could be due to controlled models being attention-only, as they lack MLPs, so all repair-relevant information must be encoded in residual-stream directions that attention can read and write. We hypothesize that pretrained models encode repair-relevant features in representations shaped by both MLP layers and task priors acquired during pretraining, which a single angular metric cannot capture.

\begin{figure}[!htbp]
    \centering
    \includegraphics[width=0.95\columnwidth]{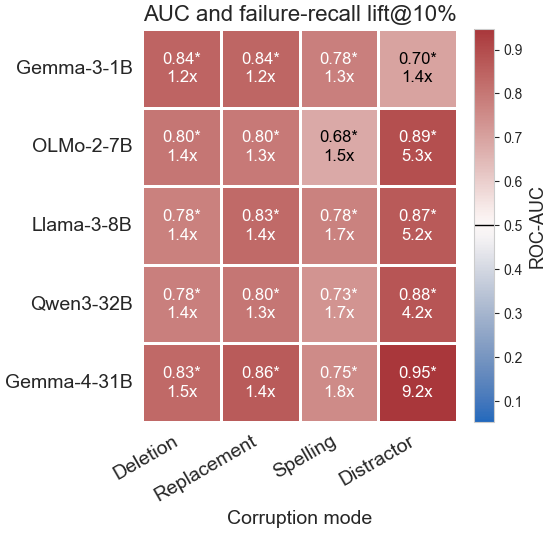}
    \caption{Earliest-layer linear-probe ROC-AUC for predicting repair failure on factual QA, by model and corruption mode. Each cell reports AUC (top) and recall lift over random flagging at the 10\% highest-risk threshold (bottom); asterisk indicates the bootstrap 95\% CI lies entirely above chance (AUC $>$ 0.5).}
    \label{fig:nlp_deployable_probe}
\end{figure}

\subsection{\emph{Can failure be predicted before generation?} Early Hidden States Support Deployable Failure Triage} 

\label{sec:deployability}

The preceding analyses establish that repair outcome is encoded in intermediate representations, but not every diagnostic is available in deployment. In particular, cosine similarity requires a clean version of the input, which is generally unavailable when a prompt is received. We therefore evaluate a deployable variant that uses only the hidden state produced by the corrupted prompt at the final prompt position. A linear probe reads this state from an early transformer layer and estimates the probability that the model will fail to recover the correct answer. It requires neither a clean reference input nor completion of output generation.

When corruption rates are pooled, the layer-0 probe achieves a mean ROC-AUC between 0.749 and 0.797 across the five pretrained models, with later early layers providing only small and inconsistent improvements (Figure ~\ref{fig:nlp_deployable_probe}, Supplementary Figure 5.21). This makes the first transformer layer the most economical operating point. Under a policy that flags the 10\% of inputs assigned the highest failure risk by the probe, the probe identifies 16.4--33.4\% of all failures, corresponding to a 1.64--3.34$\times$ recall lift over random selection. The flagged subset has 57.6--76.6\% failure precision. Thus, a system could selectively request a cleaner input, rerun the prompt, or invoke a more expensive verification procedure for a small high-risk subset rather than treating every potentially corrupted input as unreliable.

The present result supports \emph{selective failure triage} under matched or partially shifted deployment conditions: a low-cost failure-risk score, reserving intervention for inputs predicted to be at greatest risk. 

\subsection{\emph{Can repair robustness be improved?} Moderate-Corruption Finetuning Improves Robustness}
\label{sec:finding_ft}

\begin{figure}[H]
    \centering
    \includegraphics[width=0.75\columnwidth]{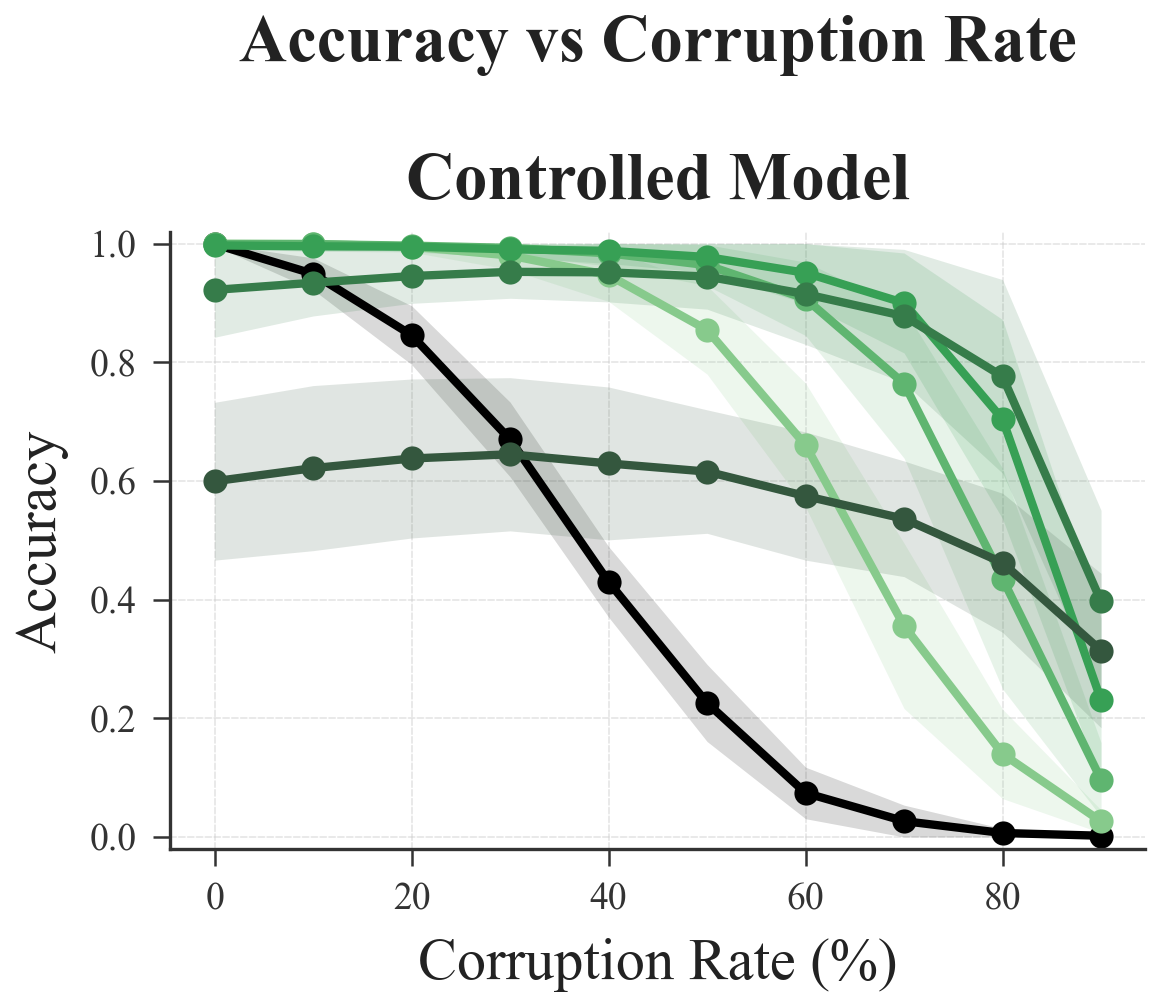}
    \includegraphics[width=0.75\columnwidth]{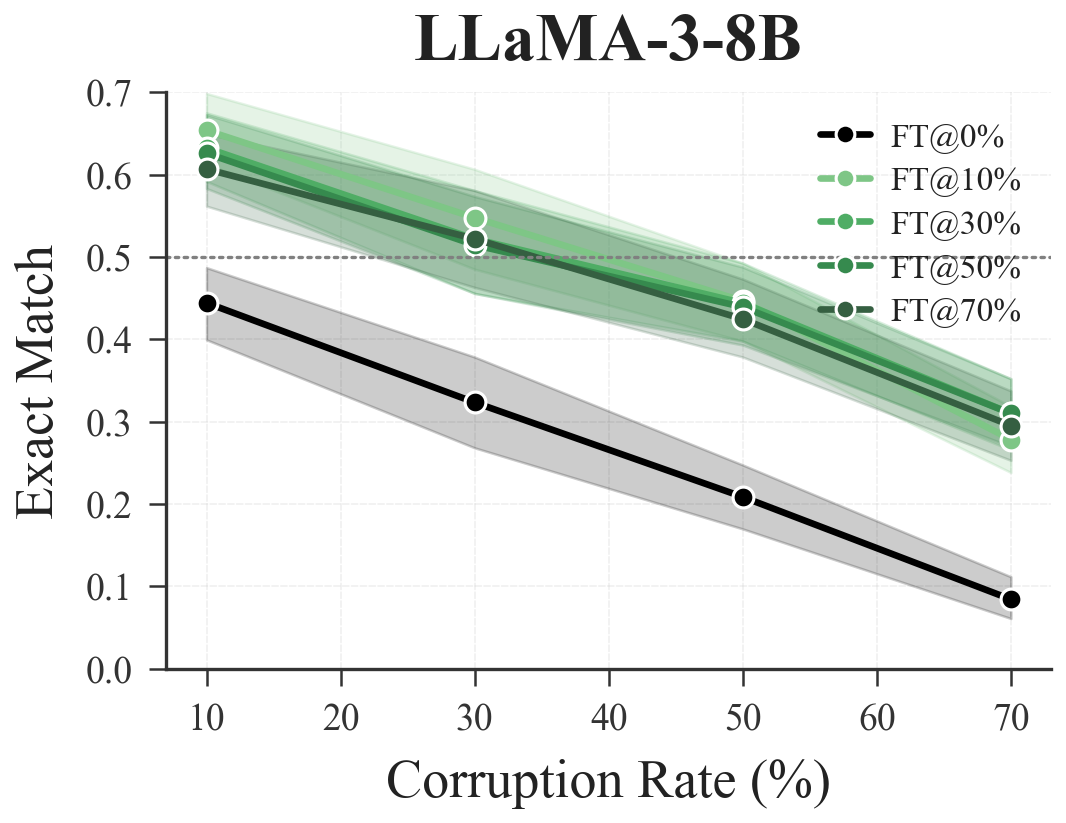}
    \caption{Finetuning on moderately corrupted sequences shifts the repair frontier outward for controlled model (Top) and upward for pretrained (Bottom).}
    \label{fig:toy_finetuning_frontier}
\end{figure}

We finetune the 10-layer model on corrupted sequences at five corruption rates (10\%, 30\%, 50\%, 70\%, 90\%) and evaluate across all ablation levels (Figure~\ref{fig:toy_finetuning_frontier} top). We denote a model finetuned at the corruption rate $r$ as FT @ $r$. FT@50 gives the best robustness-clean-accuracy trade-off, increasing the corruption rate tolerated at 50\% accuracy from 40\% in the baseline model to 90\%, while preserving clean accuracy (0.997 vs.\ 0.999). FT@70 gives a smaller robustness gain with a clean-accuracy cost (0.922), while FT@90 reduces clean accuracy to 0.600. Therefore, moderate corruption during finetuning expands the tolerance to corruption; excessive corruption sacrifices clean accuracy and gives weaker robustness gains. The behavioral gain is accompanied by a change in local geometry. Moderate-corruption finetuning reduces displacement-normalized linearization error along corruption directions, with the largest reduction around FT@50 (Supplementary Figure 5.22). Thus, the improvement in corruption tolerance is accompanied by a more nearly linear downstream response to corrupted representations.

We also finetune Llama-3-8B-Instruct on corrupted natural-language
examples at several corruption rates and evaluate corruption severities
across all four corruption modes (Figure~\ref{fig:toy_finetuning_frontier} bottom, Supplementary Figure 5.20).
Unlike the controlled setting, no single finetuning severity is uniformly
optimal. Corruption finetuning generally improves robustness, with the clearest
gains on factual QA and under spelling, replacement, and dropout corruption,
while gains on ARC and CommonsenseQA are smaller and more mode-dependent.
Distractor corruption shows comparatively little room for improvement because
the baseline is already robust. Thus, the result of the pretrained model supports the
qualitative benefit of finetuning corruption, but the severity of optimal training 
and the magnitude of improvement depend on the task and type of corruption.

\section{Conclusions}
\label{sec:limitations and concl.}
We set out to characterize how language models preserve correct task performance under corrupted inputs. \textbf{ In the controlled setting, the ability of context restoration emerges spontaneously: models trained exclusively on clean sequences retain substantial task performance under corruption without explicit denoising training. We then find that this has a structured depth-wise organization: the influence of patching varies across positions with depth, while repair outcome is reflected in residual-state alignment and downstream nonlinearity.} 

Across attention-only transformers and five pretrained LLMs (1B--32B parameters), context restoration follows a \textbf{shared two-phase spatial structure}: early layers localize repair effects at corrupted positions, and late layers concentrate the outcome at the prediction position. In the controlled model, head ablation leaves repair accuracy close to baseline, while the broader cross-model evidence show that repair outcome is associated with properties of the accumulated residual state, including alignment with the clean state and downstream nonlinearity.

\textbf{Repair and failure differ in two aspects of residual-state geometry:
alignment with the clean trajectory and nonlinearity of the downstream logit
map along the corruption direction.} Failed examples exhibit linearization residuals up to $6-8\times$ larger than repaired examples in the controlled analyses, with the disparity increasing along the corruption direction. In the controlled model, moderate-corruption finetuning increases the corruption level tolerated at 50\% accuracy from 40\% to 90\% while preserving clean performance. This behavioral improvement is accompanied by a reduction in displacement-normalized linearization error along the corruption directions, linking the increased robustness to a more nearly linear downstream response.

The divergence between controlled and pretrained settings is itself informative. Cosine similarity alone achieves within-corruption AUC 0.80--0.95 in attention-only models, but linear probes substantially outperform cosine in pretrained models. This difference may reflect the properties of hidden-state representation introduced by MLPs and pretraining, although isolating their respective contributions requires further study \cite{engels2024not}. %

The representation-level findings have two practical consequences. First, a linear probe using only the corrupted prompt's early hidden state can identify a small subset of inputs at elevated failure risk before generation, allowing
them to be routed for clarification, rerunning, or more expensive verification. Second, extending prior work on corruption finetuning, we vary training corruption severity and relate robustness gains to reduced linearization error in the controlled model, while showing that LLM gains depend on task and corruption type.

Together, these results connect context restoration to the depth-wise routing and geometry of residual representations. They provide representation-level diagnostics for repair failure and show that corruption-aware finetuning can modify both robustness and the local response to corrupted representations.

\section*{Limitations}
Our controlled symbolic setting deliberately removes MLP sublayers and uses procedurally generated arithmetic sequences so that restoration can be studied in a clean attention-only regime. This makes the mechanism tractable to study, but limits direct transfer of the quantitative findings to full-architecture transformers, multilingual settings, and tasks whose targets are not preserved under local content corruption. The divergence that we observe between controlled and pretrained models where linear probes substantially outperform cosine alignment, indicates that MLP-shaped features reshape the repair geometry in ways that our attention-only analysis cannot fully characterize. In the pretrained setting, we evaluate five instruction-tuned decoder-only LLMs (1B--32B parameters) on three English-language task families under four label-preserving corruption modes. We do not evaluate mixture-of-experts, state-space, encoder-decoder, or retrieval-augmented architectures, so the generality of these geometric analyses beyond the studied settings is untested. 

\bibliography{aaai2027}

\newpage

\setcounter{secnumdepth}{2}
\renewcommand{\thesection}{S\arabic{section}}
\renewcommand{\thesubsection}{\thesection.\arabic{subsection}}

\onecolumn
\begin{center}
    {\LARGE\bfseries Supplementary content}
    \vspace{5mm}
\end{center}
\section{Experiments and methods: Synthetic Symbolic Setting}
\label{app:methodology_symbolic}

\subsection{Problem Setup}
\label{app:problem_setup}

We study spontaneous context restoration in the symbolic setting. Given a clean input $x$, a label-preserving corrupted variant $\tilde{x} = c_r(x)$ that preserves the target $y$, and a model $f_\theta$, we ask whether $f_\theta(\tilde{x}) = y$, and trace where and how the network recovers the correct computation. At each layer $\ell$, the hidden state is $h_\ell \in \mathbb{R}^{T \times C}$, where $T$ is the token sequence length and $C$ is the hidden dimension. All predictions are read from the final position $p = T$.

\subsection{Data Generation}
\label{app:data_generation}

\paragraph{Sequence families.} We generate integer sequences from four arithmetic families: constant increment, constant decrement (subtract), add-subtract (alternating addition and subtraction with separate step sizes $q_{\mathrm{add}}$ and $q_{\mathrm{sub}}$, where $q_{\mathrm{add}} > q_{\mathrm{sub}}$), and variable arithmetic (step increases by 1 each turn). For each family, starting value $p$ and step size $q$ are drawn from Gaussian distributions with family-specific means and standard deviations (see Table~\ref{tab:sampling_params}), then rounded: $p = \lfloor |Z_p| \rfloor$, $q = \max\{1, \lfloor |Z_q| \rfloor\}$. Sequences whose values violate the constraint $0 \leq s_i \leq S_{\max}$ for all $i$ are rejected.

\paragraph{Arithmetic sequence selection.} We use procedurally generated arithmetic sequences as a controlled setting because they are simple, tractable, and have unambiguous next-token targets while still requiring the model to infer structure from context. Their generative rules can be specified exactly, corruption can be applied at known positions while preserving the target, and clean and corrupted internal trajectories can therefore be compared directly. This makes the setting suitable for mechanistic analyses of context restoration while minimizing confounds from linguistic ambiguity or external knowledge.

\begin{table}[h]
\centering
\small
\caption{Sampling parameters for each sequence family.}
\label{tab:sampling_params}
\begin{tabular}{lcccc}
\toprule
Family & $\mu_p$ & $\sigma_p$ & $\mu_q$ & $\sigma_q$ \\
\midrule
Increment & 6500 & 3500 & 100 & 200 \\
Add-subtract & 9000 & 1000 & 100 & 200 \\
Variable arithmetic & 15 & 100 & 15 & 100 \\
Subtract & 10000 & 1000 & 20 & 50 \\
\bottomrule
\end{tabular}
\end{table}

\paragraph{Train/test split.} For each family, we draw a pool of unique pairs $(p, q)$ (500K draws for most families; 10M for variable arithmetic due to lower acceptance rate). From the unique pairs, we sample 15,000 for training (4,200 for variable arithmetic) and 100 for testing, ensuring the two sets are disjoint. All sequences in the training set use a fixed length of $T_{\mathrm{num}} = 150$ numbers. The test set uses variable lengths $T_{\mathrm{num}} \in \{40, 60, 80, 100, 120, 140\}$.

\paragraph{Prompt format.} Each sequence is rendered as a comma-separated list of integers followed by a trailing comma (e.g.\ \texttt{46, 47, 49, 52,}). The prediction target is the next integer in the sequence, which is never included in the input.

\paragraph{Training corpus size.} The training set contains 39,360 sequences across all four families (15,000 per family except 4,200 for variable arithmetic, plus a small number from the add-subtract family). After tokenisation, this corresponds to approximately 14.76M tokens. The validation set is a 20\% held-out split (stratified by sequence family) drawn from the same $(p, q)$ pool as the training set.

\subsection{Tokeniser}
\label{app:tokeniser}

We train a byte-pair encoding (BPE) tokeniser on a minimal corpus of comma-separated digit strings, ensuring that the comma character is always segmented as its own token. We then add all integers from 0 to 13,000 as explicit single tokens, so every number in our sequences is guaranteed to be a single token. The integer token indices are randomly permuted at tokeniser construction time using the experiment seed; this forces the model to infer numerical structure from context rather than from embedding-index order. The final vocabulary contains 13,008 tokens: 13,001 integer tokens, a comma token, and 6 special tokens (\texttt{[PAD]}, \texttt{[UNK]}, \texttt{[CLS]}, \texttt{[SEP]}, \texttt{[MASK]}, and a BPE merge artifact).

\subsection{Model Architecture}
\label{app:architecture}

We train two decoder-only, attention-only transformers from scratch. Both use multi-head causal self-attention with no MLP layers:

\begin{table}[h]
\centering
\small
\caption{Model configurations.}
\label{tab:model_configs}
\begin{tabular}{lccccc}
\toprule
Name & Layers & Heads & $d_{\mathrm{model}}$ & Dropout & Parameters \\
\midrule
\textsc{Small} & 4 & 8 & 64 & 0.1 & 1.75M \\
\textsc{Large} & 10 & 8 & 128 & 0.1 & 4.01M \\
\bottomrule
\end{tabular}
\end{table}

\begin{figure}[!htbp]
    \centering
    \includegraphics[width=0.3\textwidth]{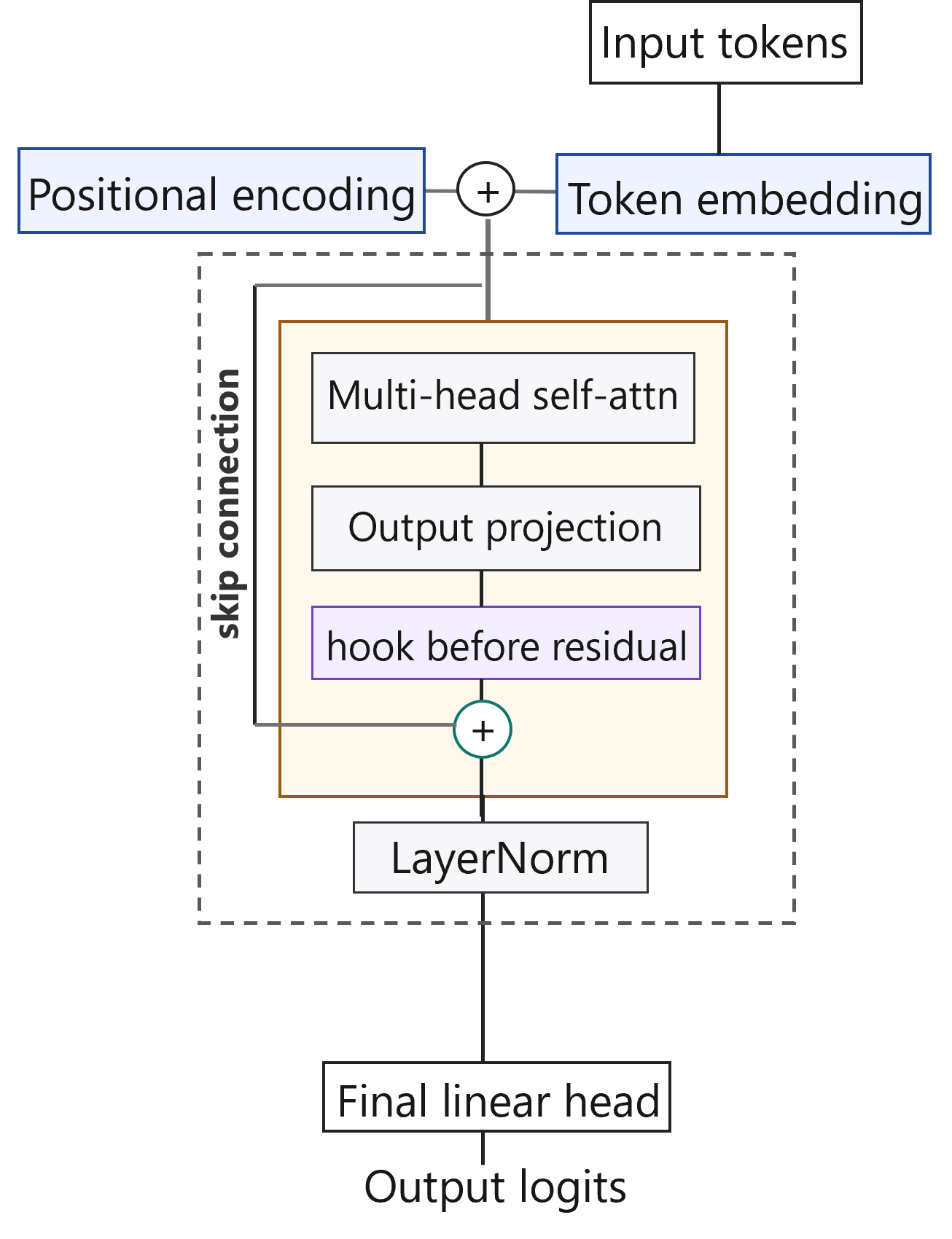}
    \caption{Attention-only transformer architecture.}
    \label{fig:custom_model}
\end{figure}

Each layer consists of: multi-head self-attention $\to$ residual addition $\to$ LayerNorm. There are no feedforward/MLP sublayers. The unembedding matrix $W_u$ projects from $d_{\mathrm{model}}$ to the vocabulary. The 10-layer model is used for the main mechanistic analyses; the 4-layer model is used to verify that observed effects are not artifacts of a particular depth.

\subsection{Training}
\label{app:training}

Both models are trained with the standard next-token prediction objective: for input sequence $(s_1, \ldots, s_T)$, the loss is the cross-entropy between the model's prediction at position $t$ and the ground-truth token $s_{t+1}$, summed over all positions $t \in \{1, \ldots, T-1\}$.

\paragraph{Optimiser.} Adam with initial learning rate $4 \times 10^{-4}$. Learning rate is reduced by a factor of 0.5 on plateau (patience 3 epochs, monitoring validation loss).

\paragraph{Hyperparameter selection.} During development, we performed a small manual exploratory sweep over learning rates $\{1\times10^{-4},\,4\times10^{-4},\,1\times10^{-3}\}$. We selected $4\times10^{-4}$ based on validation loss and stable convergence. For the logistic-regression probes used in subsequent analyses, we additionally explored regularisation strengths $C \in \{0.01,\,0.1,\,1,\,10\}$ and found the results qualitatively insensitive to this choice; we use $C=1$ throughout the reported experiments.

\paragraph{Batch size.} We manually explored batch sizes of $\{8,12,24\}$ during development and used 12 sequences per batch based on validation performance and training stability. 

\paragraph{Convergence.} Training is run for the number of epochs needed to converge ($\approx 60$ for the \textsc{Small} model, $\approx 35$ for the \textsc{Large} model), with checkpoints saved at regular intervals. All diagnostic experiments use the final converged checkpoint.

\subsection{Corruption Regimes (Test Time)}
\label{app:corruption}

For a corruption rate $r \in \{10, 20, 30, 40, 50, 60, 70, 80, 90\}\%$, we sample $\lceil r \cdot T_{\mathrm{num}} / 100 \rceil$ number positions uniformly without replacement, excluding the final observed element (which determines the prediction target). We apply one of three substitution schemes:

\begin{enumerate}
    \item \textbf{Zero ablation:} selected entries are replaced with the literal token \texttt{0}.
    \item \textbf{In-range substitution:} selected entries are replaced with integers sampled uniformly from $(\min s, \max s)$, where $s$ is the clean sequence. A single replacement value is drawn per sequence and applied to all corrupted positions in that sequence.
    \item \textbf{Out-of-range substitution:} selected entries are replaced with integers sampled from $[0, \min s) \cup (\max s, S_{\max}]$.
\end{enumerate}

In all cases the prediction target (the next number after the displayed sequence) is kept fixed, so all corruptions are label-preserving. Corrupted token positions and number positions are recorded for use in position-group patching analyses.

\subsection{Evaluation Protocol}
\label{app:evaluation}

For each combination of sequence family ($\times 4$), corruption regime ($\times 3$), corruption rate ($\times 9$), and sequence length ($\times 6$), we evaluate 100 held-out sequences whose $(p, q)$ pairs are disjoint from those used during training. Accuracy is measured by exact match between the model's argmax predicted next token and the clean ground-truth target. All evaluations use greedy decoding (no sampling).

\subsection{Fine-tuning on Corrupted Data}
\label{app:finetuning}

\paragraph{Dataset construction.} Starting from the clean training sequences, we sample 20\% for fine-tuning and 10\% for fine-tuning validation, stratified by sequence family. For each target corruption rate $r \in \{10, 30, 50, 70, 90\}\%$, we apply zero-ablation corruption to the sampled sequences. The input to the model is the corrupted sequence; the label is the original clean sequence (i.e.\ the model is trained to predict clean next tokens from corrupted inputs).

\paragraph{Training procedure.} Fine-tuning starts from the converged base checkpoint. All parameters are trainable (full-model fine-tuning, no freezing). We use AdamW with learning rate $5 \times 10^{-5}$, weight decay $0.01$, batch size 12, and ReduceLROnPlateau scheduling (factor 0.5, patience 3). Training runs for up to 100 epochs; we select the checkpoint with the lowest validation loss.

\paragraph{Evaluation.} Each fine-tuned model is evaluated on the same held-out test sets used for the base model, across all three corruption regimes (zero, in-range, out-of-range), all ablation rates (0\%--90\%), and all sequence lengths. We report accuracy-vs-ablation curves, $\tau_{50}$ (interpolated ablation rate at 50\% accuracy), mean accuracy across ablation levels, and clean accuracy (performance at 0\% corruption).

\subsection{Activation Caching}
\label{app:caching}

All analyses require access to intermediate hidden states. We extract activations using forward hooks registered on specific modules within each transformer layer. Two caching targets are used:

\begin{enumerate}
    \item \textbf{Residual-stream activations (after LayerNorm):} Hooks are registered on the full layer module (i.e.\ the output of \texttt{layers[$\ell$]}), which corresponds to the hidden state after the attention output has been added to the residual stream and passed through LayerNorm. These are denoted $h_\ell^{\mathrm{resid}} \in \mathbb{R}^{T \times C}$.
    
    \item \textbf{Layer-only activations (before residual addition):} Hooks are registered on the \texttt{self\_attention} submodule, capturing the attention output before it enters the residual connection. These are denoted $h_\ell^{\mathrm{layer}} \in \mathbb{R}^{T \times C}$.
\end{enumerate}

Hooks clone and detach all captured tensors to prevent gradient graph retention. For analyses requiring gradients (the linearisation test), the hook instead calls \texttt{retain\_grad()} on the output tensor without detaching.

\subsection{Logit Readout and Mean-Centring}
\label{app:logit_readout}

At any layer $\ell$, we read out logits by applying the model's unembedding matrix $W_u$ to the cached activation:
\[
\mathrm{Logit}_\ell = W_u \cdot h_\ell^{\mathrm{resid}}[p, :] \in \mathbb{R}^{V},
\]
where $V$ is the vocabulary size and $p$ is the prediction position. Following standard practice~\citep{elhage2021mathematical,rushing2024explorations,mcgrath2023hydra}, all logit vectors are mean-centred (subtract the mean across the vocabulary dimension) before any comparisons. This preserves the softmax distribution while removing a global offset that varies across layers.

\subsection{Activation Patching}
\label{app:patching}

Activation patching replaces hidden states from one forward pass with those from another and measures the effect on the target-token logit.

\paragraph{Setup.} For each example, we run a clean forward pass (input $x$) and a corrupted forward pass (input $\tilde{x}$), caching all layer activations $\{h_\ell^{\mathrm{clean}}\}$ and $\{h_\ell^{\mathrm{corr}}\}$ and recording the target-token logits $g^{\mathrm{clean}}$ and $g^{\mathrm{corr}}$.

\paragraph{Position-group patching.} At each layer $\ell$, we run the corrupted input but replace activations in all the corrupted positions with their counterparts from uncorrupted run, then replace the activations separately in all clean positions. This yields two scalars per layer: the mean absolute logit change $|\Delta g|$ per position from patching corrupted positions and from patching clean positions. Token positions are partitioned by whether the corresponding number token was corrupted. 

\paragraph{Per-position patching.} We also patch individual positions one at a time across all layers, producing a full $(L \times T)$ patching map per example. The cumulative repair curve sorts positions by patching effect to show that a small subset of positions carries most of the total repair signal.

\paragraph{All-position interpolation.} We interpolate between clean and corrupted activations at all positions simultaneously. At layer $\ell$ and interpolation strength $\alpha$:
\[
h_\ell^{\mathrm{patched}}[:, :] = h_\ell^{\mathrm{clean}}[:, :] + \alpha \cdot \bigl(h_\ell^{\mathrm{corr}}[:, :] - h_\ell^{\mathrm{clean}}[:, :]\bigr).
\]
We evaluate $\alpha \in \{-0.5, 0.0, 0.5, 1.0, 1.5, 2.0\}$. The base forward pass uses the clean input. The normalised logit change $\Delta g(\alpha) / \Delta g(1)$ is plotted across layers.

\subsection{Cosine Similarity and Repair Probes}
\label{app:cosine_probes}

\paragraph{Cosine similarity.} For each example and each layer $\ell$, we compute cosine similarity between the clean and corrupted residual-stream activations at the prediction position:
\[
\cos_\ell = \frac{h_\ell^{\mathrm{clean}}[p, :] \cdot h_\ell^{\mathrm{corr}}[p, :]}{\|h_\ell^{\mathrm{clean}}[p, :]\| \cdot \|h_\ell^{\mathrm{corr}}[p, :]\| + \epsilon},
\]
where $\epsilon = 10^{-10}$. The same computation is applied to layer-only activations (before residual addition), yielding two cosine trajectories per example.

\paragraph{Repair label.} An example is labelled as ``repaired'' ($y = 1$) if the model's argmax prediction on the corrupted input matches the ground-truth target token, and ``failed'' ($y = 0$) otherwise.

\paragraph{Linear probe.} We train a logistic regression classifier (scikit-learn, L-BFGS solver, $C = 1.0$, max 1000 iterations) to predict the repair label from the corrupted activation vector $h_\ell^{\mathrm{corr}}[p, :] \in \mathbb{R}^C$. Features are standardised (zero mean, unit variance) per fold. We report 5-fold stratified cross-validated ROC-AUC.

\paragraph{Cosine probe.} The same logistic regression is trained on a single scalar feature: the cosine similarity $\cos_\ell$. This tests whether angular alignment alone is predictive.

\paragraph{Within-ablation evaluation.} To control for the trivial correlation between corruption severity and repair success, we also evaluate both probes restricted to examples at a single corruption rate $\alpha$. This yields within-ablation AUC values that measure whether the probe discriminates repaired from failed examples at the same corruption level.

\paragraph{Corruption-detection probe.} A separate logistic regression is trained to predict whether an input was corrupted at all, using a balanced dataset of all unique clean activations (label 0) and all corrupted activations (label 1). This probe is evaluated per-layer to determine at what depth corruption identity becomes linearly decodable.

\subsection{Linearisation Scale Test}
\label{app:linearization}

The linearisation test measures functional curvature of the mapping from layer-$\ell$ hidden states to a target logit, along the direction of corruption-induced displacement.

\paragraph{Setup.} Let $g: \mathbb{R}^C \to \mathbb{R}$ denote the function that maps the layer-$\ell$ hidden state at position $p$ to the scalar logit of the target token, with all other positions and all downstream computation held fixed. We compute:

\begin{enumerate}[leftmargin=*]
    \item \textbf{Clean forward pass} on input $x$. A hook on layer $\ell$ captures the output $h_\ell^{\mathrm{clean}}$ with gradients retained. We read the target-token logit $g(h_\ell^{\mathrm{clean}})$ and backpropagate to obtain the gradient $\nabla_h g \big|_{h_\ell^{\mathrm{clean}}} \in \mathbb{R}^C$ at position $p$.

    \item \textbf{Corrupted forward pass} on input $\tilde{x}$. A hook captures $h_\ell^{\mathrm{corr}}$.
    
    \item \textbf{Error direction:} $e = h_\ell^{\mathrm{corr}}[p, :] - h_\ell^{\mathrm{clean}}[p, :]$, with $\|e\|$ recorded.
\end{enumerate}

\paragraph{Predicted vs.\ true logit change.} For each $\alpha \in \{0.25, 0.5, 0.75, 1.0\}$, we construct the perturbed state
\[
\tilde{h}_\ell(\alpha)[p, :] = h_\ell^{\mathrm{clean}}[p, :] + \alpha \cdot e,
\]
leaving all other positions at their clean values. The \textbf{predicted} (first-order) logit change is
\[
\Delta g_{\mathrm{pred}}(\alpha) = \alpha \cdot \nabla_h g^{\top} e,
\]
and the \textbf{true} logit change is obtained by a patched forward pass:
\[
\Delta g_{\mathrm{true}}(\alpha) = g\bigl(\tilde{h}_\ell(\alpha)\bigr) - g\bigl(h_\ell^{\mathrm{clean}}\bigr).
\]
The patching is implemented via a forward hook that modifies only position $p$ at layer $\ell$, then allows the forward pass to continue normally through all subsequent layers.

\paragraph{Residual and metrics.} The linearisation residual is $r(\alpha) = |\Delta g_{\mathrm{true}}(\alpha) - \Delta g_{\mathrm{pred}}(\alpha)|$. By Taylor's theorem,
\[
\Delta g_{\mathrm{true}}(\alpha) = \alpha \, \nabla g^{\top} e + \tfrac{1}{2}\alpha^2 \, e^{\top} H \, e + \mathcal{O}(\alpha^3),
\]
where $H$ is the Hessian of $g$ with respect to $h_\ell$ at $h_\ell^{\mathrm{clean}}$. Small $\alpha$ probes linearity; larger $\alpha$ reveals curvature along $e$. We report:
\begin{itemize}
    \item Per-$\alpha$ absolute error: $|r(\alpha)|$.
    \item Directional curvature proxy: $\kappa(\alpha) = 2 \cdot r(\alpha) / \alpha^2$, which approximates $|e^{\top} H e|$ when higher-order terms are small.
    \item RMSE across all $\alpha$ values.
    \item Score at $\alpha = 1$: $-|r(1)|$ (higher = more linear = more repairable).
\end{itemize}

\paragraph{Outcome conditioning.} Examples are first filtered to those where the clean-run argmax prediction matches the ground-truth target (ensuring the model possesses the relevant knowledge). Among these, each example is then labelled as ``repaired'' ($y=1$) if the corrupted-run argmax also matches the target, and ``failed'' ($y=0$) otherwise. All residual plots are split by this corrupted-run label to compare curvature between repaired and failed examples.

\paragraph{Per-layer execution.} The test is run independently at every layer $\ell \in \{0, \ldots, L-1\}$, for every example in the evaluation set, across all corruption regimes (zero, in-range, out-of-range) and all ablation rates (10\%--90\%).

\subsection{Head Ablation}
\label{app:head_ablation}

To test whether repair depends on specific attention heads, we mean-ablate subsets of heads at a target layer while running inference on corrupted inputs.

\paragraph{Procedure.} For a target layer $\ell$ and a set of head indices $\mathcal{H} \subseteq \{0, \ldots, H-1\}$, we replace the post-softmax attention weights of each head $h \in \mathcal{H}$ with uniform causal attention:
\[
A_{h}[i, j] = \begin{cases} 1/(i+1) & \text{if } j \leq i \\ 0 & \text{otherwise} \end{cases}
\]
This preserves the causal mask structure while removing all content-dependent attention for the ablated heads. Non-ablated heads are unmodified. The intervention is implemented as a context manager that monkey-patches the attention module's forward method during inference.

\paragraph{Ablation schedule.} We ablate subsets of size $|\mathcal{H}| \in \{4, 6, 8\}$ (out of 8 total heads), sampling 3 random subsets per size, at every layer independently. For each configuration, we record layer-wise residual token agreement and final accuracy on corrupted inputs at ablation rates 10\%, 30\%, and 50\%.

\section{Experiments and Methods: Natural-Language Setting}
\label{app:methodology_nlp}

\subsection{Problem Setup}

Natural-language experiments extend the symbolic setting to pretrained instruction-tuned LLMs evaluated on standard NLP benchmarks. The same core question applies: given a clean prompt $x$ and a label-preserving corrupted variant $\tilde{x} = c_r(x)$ that preserves the target $y$, we ask whether $f_\theta(\tilde{x}) = y$ and trace where and how recovery occurs internally. All models are evaluated with greedy decoding and no system prompt.

\subsection{Models}
\label{app:nlp_models}

We evaluate five instruction-tuned pretrained language models:

\begin{table}[h]
\centering
\small
\caption{Pretrained models evaluated in the natural-language setting.}
\label{tab:nlp_models}
\begin{tabular}{lc}
\toprule
Model & Parameters \\
\midrule
Gemma-3-1B & 1B \\
OLMo-2-1124-7B-Instruct & 7B \\
Meta-LLaMA-3-8B-Instruct & 8B \\
Qwen3-32B & 32B \\
Gemma-4-31B-IT & 31B \\
\bottomrule
\end{tabular}
\end{table}

All models are loaded via HuggingFace Transformers with \texttt{output\_hidden\_states=True}. Models up to 8B are run in bfloat16; larger models use 8-bit quantization (\texttt{bitsandbytes}). Prompts are rendered using each model's native chat template via \texttt{apply\_chat\_template} without system prompt, to avoid introducing a prompting confound. The padding uses the EOS token on the left side.

\subsection{Tasks and Prompt Construction}
\label{app:nlp_tasks}

We evaluated three task families, each cast into a unified prompt-to-answer format with up to 500 base examples per task. The examples are filtered to have at most 250 tokens after the chat-template rendering.

\paragraph{ARC-Challenge.} Multiple-choice science questions from the AI2 Reasoning Challenge~\citep{allenai:arc}. The prompt contains the question text and the labeled answer choices (A--D). The model must return only the exact answer text.

\paragraph{Factual QA.} Extractive reading-comprehension examples derived from SQuAD~\citep{rajpurkar2016squad}. The prompt contains a context passage and a question. The model must return only the short answer span copied from the context.

\paragraph{CommonsenseQA.} Common sense Multiple-choice reasoning questions~\citep{talmor-etal-2019-commonsenseqa}. The prompt contains the question and five answer choices (A--E). The model must return only the exact answer text.

\paragraph{Prompt structure.} Each prompt is decomposed into three parts:
x = (\texttt{user\_prefix},\; \texttt{corruption\_region},\;  \texttt{user\_suffix}).

Only \texttt{corruption\_region} is modified under corruption. For ARC and CommonsenseQA, the corruption region is the question text; answer choices remain intact. For Factual QA, the corruption region contains both the context and question; the instruction specifying the answer format is protected. Scaffold tokens (\texttt{Context:}, \texttt{Question:}, \texttt{Answer:}, \texttt{Choices:}) are never corrupted. This ensures that task format and gold answer are unchanged, so that all corruptions are label-preserving.

\paragraph{Dataset selection.} We select ARC-Challenge, CommonsenseQA, and a SQuAD-derived factual QA task to cover complementary forms of language-model reasoning and context use. ARC-Challenge evaluates multiple-choice scientific reasoning, CommonsenseQA evaluates multiple-choice commonsense reasoning, and factual QA evaluates extractive reading comprehension in which the answer must be recovered from an explicit context passage. Together, these tasks allow us to test whether context restoration generalizes across different task structures and sources of task-relevant information, rather than being specific to a single benchmark or response format.

\subsection{Corruption Modes}
\label{app:nlp_corruption}

We apply four corruption modes at rates $r \in \{10, 30, 50, 70, 90, 100\}\%$. The corruption rate is defined relative to the number of tokeniser tokens in the clean corruption region: if the region contains $N$ tokens, then rate $\alpha$ modifies approximately $\alpha N$ tokens' worth of content. For dropout, replacement, and spelling, eligible words are selected until the corresponding token mass reaches the target budget. For distractor corruption, $\alpha$ denotes the ratio of inserted distractor tokens to the original token count.

\begin{enumerate}
    \item \textbf{Dropout:} Selected eligible words are deleted from the corruption region. Scaffold words and punctuation are protected.

    \item \textbf{Replacement:} Selected eligible words are replaced with random high-frequency function words drawn from a fixed set (\texttt{"the"}, \texttt{"a"}, \texttt{"of"}, \texttt{"and"}, etc.). Replacements that collide with the original word are re-sampled up to 10 times.

    \item \textbf{Spelling:} Selected eligible words are misspelled by shuffling their internal characters while preserving the first and last characters. Words with $\leq 3$ characters or uniform middle characters are skipped. This corruption preserves approximate word identity while disrupting exact tokenisation.

    \item \textbf{Distractor:} Semantically irrelevant but syntactically well-formed sentences (e.g.\ \texttt{"Fish swim."}, \texttt{"Trees grow."}) are interleaved into the corruption region at random positions until the target token budget is reached. Sentences are drawn from a fixed pool of 100 factual/common-sense statements.
\end{enumerate}

All corruption functions operate on word-level tokenisation, track the achieved corruption rate in model-specific tokens, and record which word positions were modified. Only dropout and replacement alter the number of tokens; spelling preserves token count approximately; distractor strictly adds tokens.

\subsection{Evaluation Metrics}
\label{app:nlp_eval}

We evaluate model outputs using greedy decoding with \texttt{max\_new\_tokens=32}. Clean and corrupted predictions are generated in batches of 32 with left-padding. Clean predictions are cached per unique prompt to avoid redundant generation. We compute:

\begin{itemize}
    \item \textbf{Exact match (EM):} Whether the normalised prediction exactly matches any gold answer. For multiple-choice tasks, we also accept the correct option letter (A/B/C/D/E). Normalisation lowercases, removes punctuation and articles, and collapses whitespace.

    \item \textbf{Token-level F1:} Precision and recall computed over whitespace-tokenised normalised prediction and gold answer, taking the maximum over all gold answers.

    \item \textbf{Token agreement:} Whether the normalised corrupted prediction matches the normalised clean prediction. This provides a label-free measure of whether corruption changed the model's output.

    \item \textbf{Repair label:} An example is labelled as ``repaired'' if F1 $\geq 0.5$. This threshold is used for all probe-based analyses.
\end{itemize}

\subsection{Position-Group Activation Patching}
\label{app:nlp_patching}

We adapt the patching methodology from the symbolic setting to pretrained models. Because dropout and distractor corruptions change the token count (making position-aligned patching impossible), we restrict patching analyses to \textbf{replacement} and \textbf{spelling} corruptions, which preserve sequence length.

\paragraph{Corrupted position identification.} For replacement and spelling modes, we tokenize both the clean and corrupted rendered prompts and compare token-by-token. The positions where the tokens differ are labeled ``corrupted''; all other positions are ``clean''. Examples where the clean and corrupted tokenisations have different lengths are excluded.

\paragraph{Procedure.} For each example and each target layer $\ell$:
\begin{enumerate}
    \item Run a clean forward pass, cache the layer-$\ell$ activation $h_\ell^{\mathrm{clean}} \in \mathbb{R}^{T \times C}$ (moved to CPU to manage GPU memory).
    \item Run a corrupted forward pass, record the corrupted-target logit.
    \item \textbf{Patch corrupted positions:} Run the corrupted input through the model, but at layer $\ell$ replace the activations in all the corrupted positions with the corresponding activations from the uncorrupted run. Record the patched-target logit and the raw logit change.
    \item \textbf{Patch clean positions:} Same procedure, but replace activations at all clean (non-corrupted) positions.
\end{enumerate}

This yields two scalars per (example, layer): the logit changes from patching corrupted positions to patching clean positions. We report the mean absolute logit change per position, averaged within each group.

\paragraph{Layer selection.} For computational efficiency, patching is run on every fifth layer plus the second-to-last layer.

\subsection{Cosine Similarity and Repair Probes}
\label{app:nlp_probes}

\paragraph{Activation collection.} For each example, we run one forward pass on the clean prompt and one on the corrupted prompt with \texttt{output\_hidden\_states=True}. At each probed layer $\ell$, we extract the hidden state at the final token position as a single vector $h_\ell[p, :] \in \mathbb{R}^C$, cast to float32. Clean activations are cached per unique prompt text. Probes are run at every 5th layer plus the final layer.

\paragraph{Cosine similarity.} For each example and layer, we compute $\cos_\ell$ between the clean and corrupted hidden states at the final position, identically to the symbolic setting.

\paragraph{Repair probe.} A logistic regression classifier (scikit-learn, L-BFGS, $C = 1.0$, max 1000 iterations, standard-scaled features) is trained to predict the repair label (F1 $\geq 0.5$) from the corrupted activation vector $h_\ell^{\mathrm{corr}}[p, :] \in \mathbb{R}^C$. We report 5-fold stratified cross-validated ROC-AUC. The cosine probe uses $\cos_\ell$ as a single scalar feature.

\paragraph{Within-corruption evaluation.} Both probes are also evaluated within each (corruption mode, ablation rate) combination to control for trivial severity--repair correlations.

\paragraph{Corruption-detection probe.} A separate logistic regression is trained to distinguish clean from corrupted activations using the paired clean and corrupted vectors. This is evaluated per-layer to determine when corruption identity becomes linearly decodable.

\subsection{Linearization Scale Test}
\label{app:nlp_linearization}

The linearization test follows the same procedure as in the symbolic setting (Appendix~\ref{app:linearization}), with the following adaptations for HuggingFace models:

\paragraph{Gradient computation.} A forward hook on the target layer captures the hidden state and sets \texttt{requires\_grad=True} on the output tensor. The target-token logit (argmax of the clean prediction at the final position) is backpropagated to obtain the gradient $\nabla_h g \in \mathbb{R}^C$.

\paragraph{Batched patched forwards.} For efficiency, all four $\alpha$ values ($\{0.25, 0.5, 0.75, 1.0\}$) are evaluated in a single batched forward pass. The clean input is replicated 4 times and a hook adds the perturbation $\alpha \cdot e$ at the final position for each batch element. This avoids 4 separate forward passes per example.

\paragraph{Attention masks.} Both clean and corrupted forward passes use their respective attention masks. The corrupted pass uses the corrupted attention mask; the patched passes use the clean attention mask (since patching modifies hidden states within the clean computation graph).

\paragraph{Layer selection.} For computational efficiency, the linearization test is run at every 10th layer plus the second-to-last layer.

\paragraph{Outcome labels.} Repair labels (correct/incorrect) are drawn from the pre-computed evaluation metrics (EM and F1), not from the linearization test's own predictions. This ensures consistency across all analyses.

\newpage
\section{Supplementary Boxes}

\definecolor{repaircleanbg}{HTML}{E2F0D9}
\definecolor{repaircorrbg}{HTML}{FBE5D6}
\definecolor{repairtarget}{HTML}{FF0000}
\definecolor{repairborder}{HTML}{D9D9D9}
\definecolor{repairtitlebg}{HTML}{D0D0D0}

\newtcolorbox{repairbox}[3][]{
    width=\linewidth,
    height=6.0cm,
    colback=#2,
    colframe=repairborder,
    colbacktitle=repairtitlebg,
    coltitle=white,
    boxrule=0.5pt,
    arc=1.2mm,
    left=1.5mm,right=1.5mm,top=0.8mm,bottom=0.8mm,
    fonttitle=\bfseries\footnotesize,
    title={#3},
    valign=top,
    #1
}

\begin{figure*}[!htbp]
\centering
\setlength{\tabcolsep}{6pt}
\renewcommand{\arraystretch}{1}
\begin{tabular}{m{0.32\textwidth} m{0.32\textwidth} m{0.32\textwidth}}

\begin{repairbox}{repaircleanbg}{ARC clean ($p=0$)\\ Corruption: none}
\scriptsize
\textbf{Question:} George wants to warm his hands quickly by rubbing them. Which skin surface will produce the most heat?\\[0.5mm]
\textbf{Choices / instruction:} Choices:\\
A) dry palms\\
B) wet palms\\
C) palms covered with oil\\
D) palms covered with lotion\\
Return only the exact answer text from the choices. Do not explain.\\[0.5mm]
\textbf{Gold answer:} \textcolor{repairtarget}{dry palms}
\end{repairbox}
&
\begin{repairbox}{repaircorrbg}{ARC corrupted ($p=0.5$)\\ Corruption: spelling}
\scriptsize
\textbf{Question:} George wnats to wram his hndas quickly by rubbing tehm. Wihch sikn surface wlil produce the most haet?\\[0.5mm]
\textbf{Choices / instruction:} Choices:\\
A) dry palms\\
B) wet palms\\
C) palms covered with oil\\
D) palms covered with lotion\\
Return only the exact answer text from the choices. Do not explain.\\[0.5mm]
\textbf{Gold answer:} \textcolor{repairtarget}{dry palms}
\end{repairbox}
&
\begin{repairbox}{repaircorrbg}{ARC corrupted ($p=0.5$)\\ Corruption: dropout}
\scriptsize
\textbf{Question:} George wants to hands rubbing. skin will the?\\[0.5mm]
\textbf{Choices / instruction:} Choices:\\
A) dry palms\\
B) wet palms\\
C) palms covered with oil\\
D) palms covered with lotion\\
Return only the exact answer text from the choices. Do not explain.\\[0.5mm]
\textbf{Gold answer:} \textcolor{repairtarget}{dry palms}
\end{repairbox}
\\[1.8mm]

\begin{repairbox}{repaircorrbg}{ARC corrupted ($p=0.5$)\\ Corruption: replacement}
\scriptsize
\textbf{Question:} an wants to warm or a the by rubbing them. and an and will produce an but of?\\[0.5mm]
\textbf{Choices / instruction:} Choices:\\
A) dry palms\\
B) wet palms\\
C) palms covered with oil\\
D) palms covered with lotion\\
Return only the exact answer text from the choices. Do not explain.\\[0.5mm]
\textbf{Gold answer:} \textcolor{repairtarget}{dry palms}
\end{repairbox}
&
\begin{repairbox}{repaircorrbg}{ARC corrupted ($p=0.5$)\\ Corruption: distractor}
\scriptsize
\textbf{Question:} George wants to warm his hands quickly by rubbing them. Dolphins jump water. Monkeys climb. Which skin surface will produce the most heat?\\[0.5mm]
\textbf{Choices / instruction:} Choices:\\
A) dry palms\\
B) wet palms\\
C) palms covered with oil\\
D) palms covered with lotion\\
Return only the exact answer text from the choices. Do not explain.\\[0.5mm]
\textbf{Gold answer:} \textcolor{repairtarget}{dry palms}
\end{repairbox}
&
\mbox{}
\end{tabular}
\caption{ARC multiple-choice prompts under all corruption families present in the uploaded task file. Green denotes the clean prompt. Orange denotes corrupted prompts at the sampled corruption level $p=0.5$. Red highlights the gold answer.}
\label{fig:arc_all_corruptions_appendix}
\end{figure*}

\definecolor{repaircleanbg}{HTML}{E2F0D9}
\definecolor{repaircorrbg}{HTML}{FBE5D6}
\definecolor{repairtarget}{HTML}{FF0000}
\definecolor{repairborder}{HTML}{D9D9D9}
\definecolor{repairtitlebg}{HTML}{D0D0D0}

\newtcolorbox{repairbox4}[3][]{
    width=\linewidth,
    height=6.0cm,
    colback=#2,
    colframe=repairborder,
    colbacktitle=repairtitlebg,
    coltitle=white,
    boxrule=0.5pt,
    arc=1.2mm,
    left=1.5mm,right=1.5mm,top=0.8mm,bottom=0.8mm,
    fonttitle=\bfseries\footnotesize,
    title={#3},
    valign=top,
    #1
}

\begin{figure*}[!htbp]
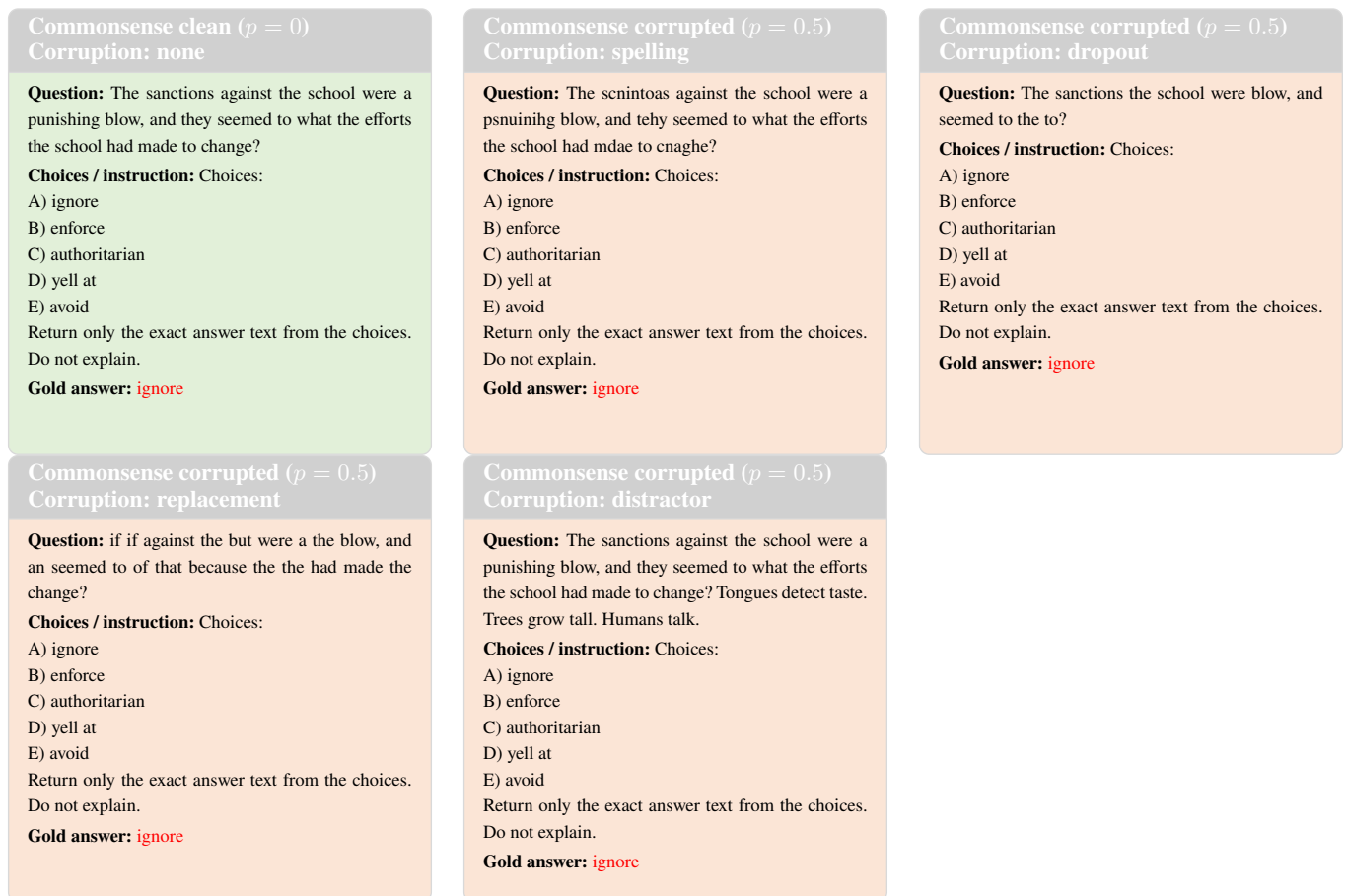

\centering
\setlength{\tabcolsep}{6pt}
\renewcommand{\arraystretch}{1}
\begin{tabular}{m{0.32\textwidth} m{0.32\textwidth} m{0.32\textwidth}}

\begin{repairbox4}{repaircleanbg}{Commonsense clean ($p=0$)\\ Corruption: none}
\scriptsize
\textbf{Question:} The sanctions against the school were a punishing blow, and they seemed to what the efforts the school had made to change?\\[0.5mm]
\textbf{Choices / instruction:} Choices:\\
A) ignore\\
B) enforce\\
C) authoritarian\\
D) yell at\\
E) avoid\\
Return only the exact answer text from the choices. Do not explain.\\[0.5mm]
\textbf{Gold answer:} \textcolor{repairtarget}{ignore}
\end{repairbox4}
&
\begin{repairbox4}{repaircorrbg}{Commonsense corrupted ($p=0.5$)\\ Corruption: spelling}
\scriptsize
\textbf{Question:} The scnintoas against the school were a psnuinihg blow, and tehy seemed to what the efforts the school had mdae to cnaghe?\\[0.5mm]
\textbf{Choices / instruction:} Choices:\\
A) ignore\\
B) enforce\\
C) authoritarian\\
D) yell at\\
E) avoid\\
Return only the exact answer text from the choices. Do not explain.\\[0.5mm]
\textbf{Gold answer:} \textcolor{repairtarget}{ignore}
\end{repairbox4}
&
\begin{repairbox4}{repaircorrbg}{Commonsense corrupted ($p=0.5$)\\ Corruption: dropout}
\scriptsize
\textbf{Question:} The sanctions the school were blow, and seemed to the to?\\[0.5mm]
\textbf{Choices / instruction:} Choices:\\
A) ignore\\
B) enforce\\
C) authoritarian\\
D) yell at\\
E) avoid\\
Return only the exact answer text from the choices. Do not explain.\\[0.5mm]
\textbf{Gold answer:} \textcolor{repairtarget}{ignore}
\end{repairbox4}
\\[1.8mm]

\begin{repairbox4}{repaircorrbg}{Commonsense corrupted ($p=0.5$)\\ Corruption: replacement}
\scriptsize
\textbf{Question:} if if against the but were a the blow, and an seemed to of that because the the had made the change?\\[0.5mm]
\textbf{Choices / instruction:} Choices:\\
A) ignore\\
B) enforce\\
C) authoritarian\\
D) yell at\\
E) avoid\\
Return only the exact answer text from the choices. Do not explain.\\[0.5mm]
\textbf{Gold answer:} \textcolor{repairtarget}{ignore}
\end{repairbox4}
&
\begin{repairbox4}{repaircorrbg}{Commonsense corrupted ($p=0.5$)\\ Corruption: distractor}
\scriptsize
\textbf{Question:} The sanctions against the school were a punishing blow, and they seemed to what the efforts the school had made to change? Tongues detect taste. Trees grow tall. Humans talk.\\[0.5mm]
\textbf{Choices / instruction:} Choices:\\
A) ignore\\
B) enforce\\
C) authoritarian\\
D) yell at\\
E) avoid\\
Return only the exact answer text from the choices. Do not explain.\\[0.5mm]
\textbf{Gold answer:} \textcolor{repairtarget}{ignore}
\end{repairbox4}
&
\mbox{}

\end{tabular}
\caption{Commonsense multiple-choice prompts under all corruption families present in the uploaded task file. Green denotes the clean prompt. Orange denotes corrupted prompts at the sampled corruption level $p=0.5$ from the CSV artifact. Red highlights the gold answer.}
\label{fig:commonsense_all_corruptions_appendix}
\end{figure*}

\definecolor{repaircleanbg}{HTML}{E2F0D9}
\definecolor{repaircorrbg}{HTML}{FBE5D6}
\definecolor{repairtarget}{HTML}{FF0000}
\definecolor{repairborder}{HTML}{D9D9D9}
\definecolor{repairtitlebg}{HTML}{D0D0D0}

\newtcolorbox{repairbox2}[3][]{
    width=\linewidth,
    height=10.5cm,
    colback=#2,
    colframe=repairborder,
    colbacktitle=repairtitlebg,
    coltitle=white,
    boxrule=0.5pt,
    arc=1.2mm,
    left=1.5mm,right=1.5mm,top=0.8mm,bottom=0.8mm,
    fonttitle=\bfseries\footnotesize,
    title={#3},
    valign=top,
    #1
}

\begin{figure*}[!htbp]
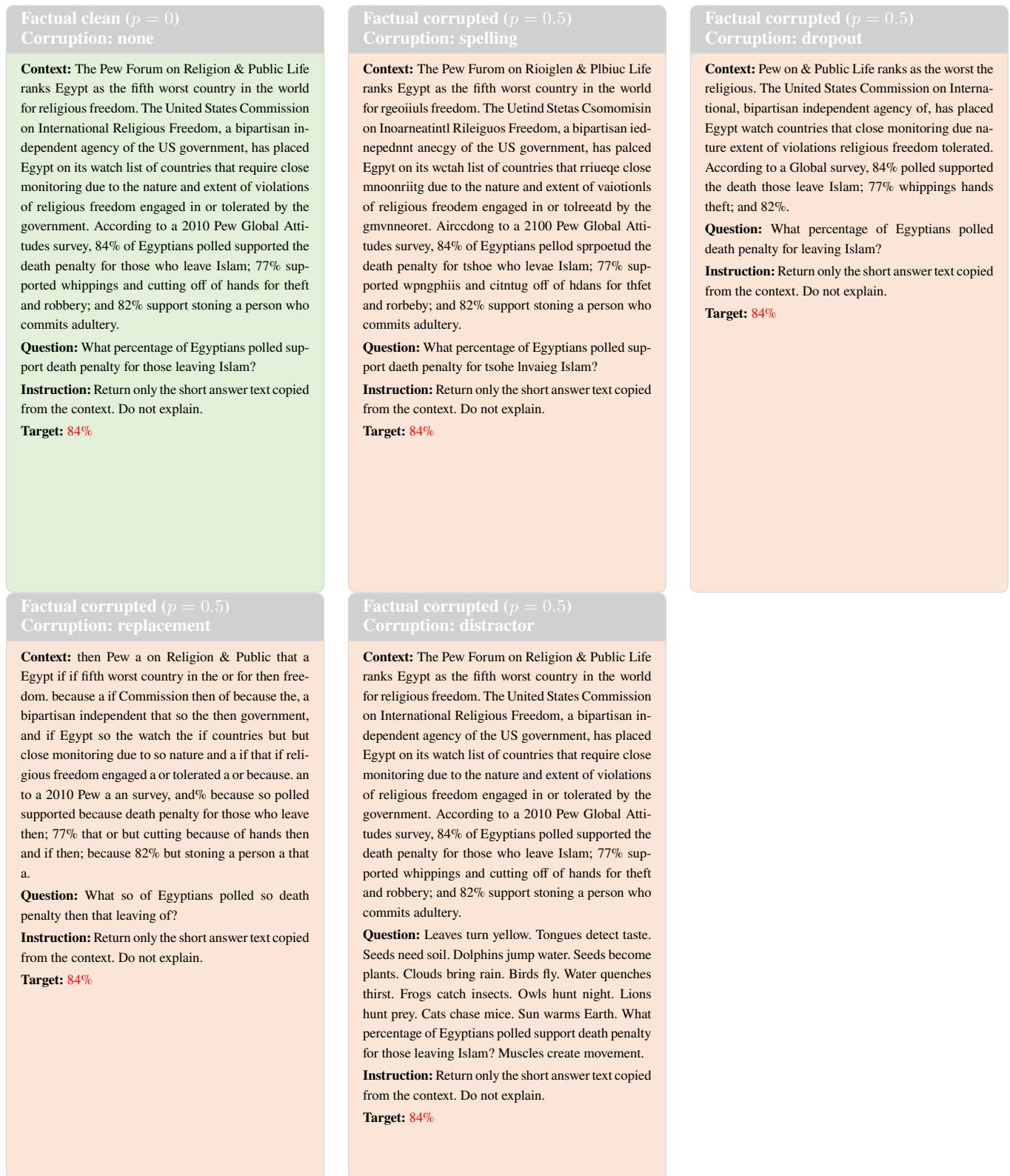

\centering
\setlength{\tabcolsep}{6pt}
\renewcommand{\arraystretch}{1}
\begin{tabular}{m{0.32\textwidth} m{0.32\textwidth} m{0.32\textwidth}}

\begin{repairbox2}{repaircleanbg}{Factual clean ($p=0$)\\ Corruption: none}
\scriptsize
\textbf{Context:} The Pew Forum on Religion \& Public Life ranks Egypt as the fifth worst country in the world for religious freedom. The United States Commission on International Religious Freedom, a bipartisan independent agency of the US government, has placed Egypt on its watch list of countries that require close monitoring due to the nature and extent of violations of religious freedom engaged in or tolerated by the government. According to a 2010 Pew Global Attitudes survey, 84\% of Egyptians polled supported the death penalty for those who leave Islam; 77\% supported whippings and cutting off of hands for theft and robbery; and 82\% support stoning a person who commits adultery.\\[0.5mm]
\textbf{Question:} What percentage of Egyptians polled support death penalty for those leaving Islam?\\[0.5mm]
\textbf{Instruction:} Return only the short answer text copied from the context. Do not explain.\\[0.5mm]
\textbf{Target:} \textcolor{repairtarget}{84\%}
\end{repairbox2}
&
\begin{repairbox2}{repaircorrbg}{Factual corrupted ($p=0.5$)\\ Corruption: spelling}
\scriptsize
\textbf{Context:} The Pew Furom on Rioiglen \& Plbiuc Life ranks Egypt as the fifth worst country in the world for rgeoiiuls freedom. The Uetind Stetas Csomomisin on Inoarneatintl Rileiguos Freedom, a bipartisan iednepednnt anecgy of the US government, has palced Egpyt on its wctah list of countries that rriueqe close mnoonriitg due to the nature and extent of vaiotionls of religious freodem engaged in or tolreeatd by the gmvnneoret. Airccdong to a 2100 Pew Global Attitudes survey, 84\% of Egyptians pellod sprpoetud the death penalty for tshoe who levae Islam; 77\% supported wpngphiis and citntug off of hdans for thfet and rorbeby; and 82\% support stoning a person who commits adultery.\\[0.5mm]
\textbf{Question:} What percentage of Egyptians polled support daeth penalty for tsohe lnvaieg Islam?\\[0.5mm]
\textbf{Instruction:} Return only the short answer text copied from the context. Do not explain.\\[0.5mm]
\textbf{Target:} \textcolor{repairtarget}{84\%}
\end{repairbox2}
&
\begin{repairbox2}{repaircorrbg}{Factual corrupted ($p=0.5$)\\ Corruption: dropout}
\scriptsize
\textbf{Context:} Pew on \& Public Life ranks as the worst the religious. The United States Commission on International, bipartisan independent agency of, has placed Egypt watch countries that close monitoring due nature extent of violations religious freedom tolerated. According to a Global survey, 84\% polled supported the death those leave Islam; 77\% whippings hands theft; and 82\%.\\[0.5mm]
\textbf{Question:} What percentage of Egyptians polled death penalty for leaving Islam?\\[0.5mm]
\textbf{Instruction:} Return only the short answer text copied from the context. Do not explain.\\[0.5mm]
\textbf{Target:} \textcolor{repairtarget}{84\%}
\end{repairbox2}
\\[1.8mm]

\begin{repairbox2}{repaircorrbg}{Factual corrupted ($p=0.5$)\\ Corruption: replacement}
\scriptsize
\textbf{Context:} then Pew a on Religion \& Public that a Egypt if if fifth worst country in the or for then freedom. because a if Commission then of because the, a bipartisan independent that so the then government, and if Egypt so the watch the if countries but but close monitoring due to so nature and a if that if religious freedom engaged a or tolerated a or because. an to a 2010 Pew a an survey, and\% because so polled supported because death penalty for those who leave then; 77\% that or but cutting because of hands then and if then; because 82\% but stoning a person a that a.\\[0.5mm]
\textbf{Question:} What so of Egyptians polled so death penalty then that leaving of?\\[0.5mm]
\textbf{Instruction:} Return only the short answer text copied from the context. Do not explain.\\[0.5mm]
\textbf{Target:} \textcolor{repairtarget}{84\%}
\end{repairbox2}
&
\begin{repairbox2}{repaircorrbg}{Factual corrupted ($p=0.5$)\\ Corruption: distractor}
\scriptsize
\textbf{Context:} The Pew Forum on Religion \& Public Life ranks Egypt as the fifth worst country in the world for religious freedom. The United States Commission on International Religious Freedom, a bipartisan independent agency of the US government, has placed Egypt on its watch list of countries that require close monitoring due to the nature and extent of violations of religious freedom engaged in or tolerated by the government. According to a 2010 Pew Global Attitudes survey, 84\% of Egyptians polled supported the death penalty for those who leave Islam; 77\% supported whippings and cutting off of hands for theft and robbery; and 82\% support stoning a person who commits adultery.\\[0.5mm]
\textbf{Question:} Leaves turn yellow. Tongues detect taste. Seeds need soil. Dolphins jump water. Seeds become plants. Clouds bring rain. Birds fly. Water quenches thirst. Frogs catch insects. Owls hunt night. Lions hunt prey. Cats chase mice. Sun warms Earth. What percentage of Egyptians polled support death penalty for those leaving Islam? Muscles create movement.\\[0.5mm]
\textbf{Instruction:} Return only the short answer text copied from the context. Do not explain.\\[0.5mm]
\textbf{Target:} \textcolor{repairtarget}{84\%}
\end{repairbox2}

\end{tabular}
\caption{Factual prompts under all corruption families present in the uploaded task file. Green denotes the clean prompt. Orange denotes corrupted prompts at the sampled corruption level $p=0.5$. Red highlights the target answer.}
\label{fig:factual_all_corruptions_appendix}
\end{figure*}

\definecolor{repaircleanbg}{HTML}{E2F0D9}
\definecolor{repaircorrbg}{HTML}{FBE5D6}
\definecolor{repairtarget}{HTML}{FF0000}
\definecolor{repairborder}{HTML}{D9D9D9}
\definecolor{repairtitlebg}{HTML}{D0D0D0}
\definecolor{repairbad}{HTML}{8B0000}

\newcommand{\corr}[1]{\textcolor{repairbad}{\textbf{#1}}}

\newtcolorbox{repairbox3}[3][]{
    width=\linewidth,
    height=5.2cm,
    colback=#2,
    colframe=repairborder,
    colbacktitle=repairtitlebg,
    coltitle=black,
    boxrule=0.5pt,
    arc=1.2mm,
    left=1.2mm,right=1.2mm,top=0.8mm,bottom=0.8mm,
    fonttitle=\bfseries\footnotesize,
    title={#3},
    valign=top,
    #1
}

\begin{figure*}[!htbp]
\centering
\setlength{\tabcolsep}{4pt}
\renewcommand{\arraystretch}{1}

\resizebox{\textwidth}{!}{%
\begin{tabular}{m{0.245\textwidth} m{0.245\textwidth} m{0.245\textwidth} m{0.245\textwidth}}

\begin{repairbox3}{repaircleanbg}{Add-subtract clean\\Corruption: none}
\scriptsize
\textbf{Sequence:}\\
10185,\ 10220,\ 10205,\ 10240,\ 10225,\ 10260,\ 10245,\ 10280,\ 10265,\ 10300,\ 10285,\ 10320,\ 10305,\ 10340,\ 10325,\ 10360,\ 10345,\ 10380,\ 10365,\ 10400,\ 10385,\ 10420,\ 10405,\ 10440,\ 10425,\ 10460,\ 10445,\ 10480,\ 10465,\ 10500,\ 10485,\ 10520,\ 10505,\ 10540,\ 10525,\ 10560,\ 10545,\ 10580,\ 10565,\ 10600\\[0.5mm]
\textbf{Target next number:} \textcolor{repairtarget}{10585}
\end{repairbox3}
&
\begin{repairbox3}{repaircorrbg}{Add- subtract corrupted\\Corruption: in-range ablation}
\scriptsize
\textbf{Sequence:}\\
\corr{10476},\ 10220,\ 10205,\ 10240,\ 10225,\ 10260,\ 10245,\ 10280,\ \corr{10476},\ 10300,\ 10285,\ 10320,\ 10305,\ 10340,\ 10325,\ 10360,\ \corr{10476},\ 10380,\ 10365,\ 10400,\ 10385,\ 10420,\ 10405,\ \corr{10476},\ 10425,\ 10460,\ 10445,\ 10480,\ 10465,\ 10500,\ 10485,\ 10520,\ 10505,\ 10540,\ 10525,\ 10560,\ 10545,\ 10580,\ 10565,\ 10600\\[0.5mm]
\textbf{Target next number:} \textcolor{repairtarget}{10585}
\end{repairbox3}
&
\begin{repairbox3}{repaircorrbg}{Add- subtract corrupted\\Corruption: out-of-range ablation}
\scriptsize
\textbf{Sequence:}\\
10185,\ 10220,\ 10205,\ 10240,\ 10225,\ 10260,\ 10245,\ 10280,\ 10265,\ 10300,\ 10285,\ 10320,\ 10305,\ 10340,\ 10325,\ 10360,\ 10345,\ 10380,\ \corr{10002},\ \corr{10002},\ 10385,\ 10420,\ 10405,\ 10440,\ 10425,\ 10460,\ 10445,\ 10480,\ \corr{10002},\ 10500,\ \corr{10002},\ 10520,\ 10505,\ 10540,\ 10525,\ 10560,\ 10545,\ 10580,\ 10565,\ 10600\\[0.5mm]
\textbf{Target next number:} \textcolor{repairtarget}{10585}
\end{repairbox3}
&
\begin{repairbox3}{repaircorrbg}{Add- subtract corrupted\\Corruption: zero ablation}
\scriptsize
\textbf{Sequence:}\\
10185,\ 10220,\ 10205,\ \corr{0},\ 10225,\ 10260,\ 10245,\ 10280,\ \corr{0},\ 10300,\ 10285,\ 10320,\ 10305,\ 10340,\ 10325,\ 10360,\ 10345,\ 10380,\ 10365,\ \corr{0},\ \corr{0},\ 10420,\ 10405,\ 10440,\ 10425,\ 10460,\ 10445,\ 10480,\ 10465,\ 10500,\ 10485,\ 10520,\ 10505,\ 10540,\ 10525,\ 10560,\ 10545,\ 10580,\ 10565,\ 10600\\[0.5mm]
\textbf{Target next number:} \textcolor{repairtarget}{10585}
\end{repairbox3}

\\[1.8mm]

\begin{repairbox3}{repaircleanbg}{Variable arithmetic clean\\Corruption: none}
\scriptsize
\textbf{Sequence:}\\
46,\ 47,\ 49,\ 52,\ 56,\ 61,\ 67,\ 74,\ 82,\ 91,\ 101,\ 112,\ 124,\ 137,\ 151,\ 166,\ 182,\ 199,\ 217,\ 236,\ 256,\ 277,\ 299,\ 322,\ 346,\ 371,\ 397,\ 424,\ 452,\ 481,\ 511,\ 542,\ 574,\ 607,\ 641,\ 676,\ 712,\ 749,\ 787,\ 826\\[0.5mm]
\textbf{Target next number:} \textcolor{repairtarget}{866}
\end{repairbox3}
&
\begin{repairbox3}{repaircorrbg}{Variable arithmetic corrupted\\Corruption: in-range ablation}
\scriptsize
\textbf{Sequence:}\\
46,\ 47,\ 49,\ 52,\ 56,\ 61,\ 67,\ 74,\ 82,\ 91,\ 101,\ 112,\ 124,\ 137,\ 151,\ 166,\ 182,\ 199,\ 217,\ 236,\ 256,\ 277,\ 299,\ 322,\ 346,\ 371,\ 397,\ 424,\ 452,\ 481,\ 511,\ 542,\ 574,\ 607,\ 641,\ 676,\ \corr{152},\ 749,\ 787,\ 826\\[0.5mm]
\textbf{Target next number:} \textcolor{repairtarget}{866}
\end{repairbox3}
&
\begin{repairbox3}{repaircorrbg}{Variable arithmetic corrupted\\Corruption: out-of-range ablation}
\scriptsize
\textbf{Sequence:}\\
46,\ 47,\ 49,\ 52,\ 56,\ 61,\ 67,\ 74,\ 82,\ \corr{2547},\ 101,\ 112,\ 124,\ 137,\ 151,\ 166,\ 182,\ 199,\ 217,\ 236,\ 256,\ 277,\ 299,\ 322,\ 346,\ 371,\ 397,\ 424,\ 452,\ 481,\ 511,\ 542,\ 574,\ 607,\ 641,\ 676,\ 712,\ 749,\ \corr{2547},\ 826\\[0.5mm]
\textbf{Target next number:} \textcolor{repairtarget}{866}
\end{repairbox3}
&
\begin{repairbox3}{repaircorrbg}{Variable arithmetic corrupted\\Corruption: zero ablation}
\scriptsize
\textbf{Sequence:}\\
46,\ 47,\ 49,\ 52,\ 56,\ 61,\ 67,\ 74,\ 82,\ \corr{0},\ 101,\ 112,\ 124,\ 137,\ 151,\ 166,\ 182,\ 199,\ 217,\ 236,\ 256,\ 277,\ 299,\ 322,\ 346,\ 371,\ 397,\ 424,\ 452,\ 481,\ 511,\ 542,\ 574,\ 607,\ 641,\ 676,\ 712,\ 749,\ \corr{0},\ 826\\[0.5mm]
\textbf{Target next number:} \textcolor{repairtarget}{866}
\end{repairbox3}

\\[1.8mm]

\begin{repairbox3}{repaircleanbg}{Subtract clean\\Corruption: none}
\scriptsize
\textbf{Sequence:}\\
12787,\ 12757,\ 12727,\ 12697,\ 12667,\ 12637,\ 12607,\ 12577,\ 12547,\ 12517,\ 12487,\ 12457,\ 12427,\ 12397,\ 12367,\ 12337,\ 12307,\ 12277,\ 12247,\ 12217,\ 12187,\ 12157,\ 12127,\ 12097,\ 12067,\ 12037,\ 12007,\ 11977,\ 11947,\ 11917,\ 11887,\ 11857,\ 11827,\ 11797,\ 11767,\ 11737,\ 11707,\ 11677,\ 11647,\ 11617\\[0.5mm]
\textbf{Target next number:} \textcolor{repairtarget}{11587}
\end{repairbox3}
&
\begin{repairbox3}{repaircorrbg}{Subtract corrupted\\Corruption: in-range ablation}
\scriptsize
\textbf{Sequence:}\\
12787,\ 12757,\ 12727,\ 12697,\ 12667,\ 12637,\ 12607,\ 12577,\ 12547,\ 12517,\ 12487,\ 12457,\ 12427,\ 12397,\ 12367,\ 12337,\ 12307,\ 12277,\ 12247,\ \corr{11914},\ 12187,\ 12157,\ 12127,\ 12097,\ 12067,\ 12037,\ 12007,\ 11977,\ 11947,\ 11917,\ 11887,\ 11857,\ 11827,\ 11797,\ 11767,\ 11737,\ 11707,\ 11677,\ 11647,\ 11617\\[0.5mm]
\textbf{Target next number:} \textcolor{repairtarget}{11587}
\end{repairbox3}
&
\begin{repairbox3}{repaircorrbg}{Subtract corrupted\\Corruption: out-of-range ablation}
\scriptsize
\textbf{Sequence:}\\
12787,\ 12757,\ 12727,\ 12697,\ 12667,\ 12637,\ 12607,\ 12577,\ 12547,\ \corr{12833},\ 12487,\ \corr{12833},\ 12427,\ 12397,\ 12367,\ 12337,\ 12307,\ 12277,\ 12247,\ 12217,\ 12187,\ 12157,\ 12127,\ 12097,\ 12067,\ 12037,\ 12007,\ 11977,\ 11947,\ 11917,\ 11887,\ 11857,\ 11827,\ 11797,\ 11767,\ 11737,\ 11707,\ 11677,\ 11647,\ 11617\\[0.5mm]
\textbf{Target next number:} \textcolor{repairtarget}{11587}
\end{repairbox3}
&
\begin{repairbox3}{repaircorrbg}{Subtract corrupted\\Corruption: zero ablation}
\scriptsize
\textbf{Sequence:}\\
12787,\ 12757,\ 12727,\ 12697,\ 12667,\ 12637,\ 12607,\ 12577,\ 12547,\ \corr{0},\ 12487,\ \corr{0},\ 12427,\ 12397,\ 12367,\ 12337,\ 12307,\ 12277,\ 12247,\ 12217,\ 12187,\ 12157,\ 12127,\ 12097,\ 12067,\ 12037,\ 12007,\ 11977,\ 11947,\ 11917,\ 11887,\ 11857,\ 11827,\ 11797,\ 11767,\ 11737,\ 11707,\ 11677,\ 11647,\ 11617\\[0.5mm]
\textbf{Target next number:} \textcolor{repairtarget}{11587}
\end{repairbox3}

\end{tabular}%
}

\caption{
Representative symbolic toy-task prompts with full sequences. Green boxes denote clean inputs. Red boxes denote corrupted inputs under in-range, out-of-range, and zero ablations. Corrupted entries are highlighted in dark red, while the gold next-number target remains unchanged.
}
\label{fig:symbolic_toy_all_corruptions}
\end{figure*}
\FloatBarrier

\section{Supplementary tables}
\label{app:generated_tables}

\begin{table}[!htbp]
\centering
\label{tab:head_ablation}
\resizebox{0.9\columnwidth}{!}{\begin{tabular}{cccc}
\toprule
Heads ablated & Corruption rate & No ablation & Mean accuracy over ablated-head layer \\
\midrule
4 & 10\% & 0.949 & 0.950 (0.218) \\
4 & 30\% & 0.670 & 0.683 (0.465) \\
4 & 50\% & 0.226 & 0.189 (0.391) \\
6 & 10\% & 0.949 & 0.939 (0.240) \\
6 & 30\% & 0.670 & 0.653 (0.476) \\
6 & 50\% & 0.226 & 0.200 (0.400) \\
8 & 10\% & 0.949 & 0.953 (0.212) \\
8 & 30\% & 0.670 & 0.650 (0.477) \\
8 & 50\% & 0.226 & 0.225 (0.418) \\
\bottomrule
\end{tabular}
}
\caption{Controlled-model head ablation under zero ablation. The no-ablation column reports the baseline at the same corruption rate; the ablated column reports final post-ablation accuracy, averaged over the ten possible layers in which the head subset is ablated, with standard deviation in parentheses.}
\end{table}

\begin{table}[!htbp]
\centering
\label{tab:iscorrupted_toy}
\resizebox{0.6\columnwidth}{!}{\begin{tabular}{llccc}
\toprule
Model & Layer & In-range & Out-of-range & Zero \\
\midrule
Large, 10L & 0 & 0.599 & 0.647 & 0.978 \\
Large, 10L & 1 & 0.545 & 0.632 & 0.964 \\
Large, 10L & 2 & 0.554 & 0.658 & 0.969 \\
Large, 10L & 3 & 0.703 & 0.761 & 0.978 \\
Large, 10L & 4 & 0.723 & 0.771 & 0.976 \\
Large, 10L & 5 & 0.789 & 0.839 & 0.975 \\
Large, 10L & 6 & 0.839 & 0.869 & 0.970 \\
Large, 10L & 7 & 0.812 & 0.852 & 0.972 \\
Large, 10L & 8 & 0.801 & 0.835 & 0.976 \\
Large, 10L & 9 & 0.802 & 0.839 & 0.977 \\
Small, 4L & 0 & 0.586 & 0.671 & 0.995 \\
Small, 4L & 1 & 0.586 & 0.659 & 0.944 \\
Small, 4L & 2 & 0.665 & 0.720 & 0.972 \\
Small, 4L & 3 & 0.673 & 0.777 & 0.976 \\
\bottomrule
\end{tabular}
}
\caption{Controlled-model is-corrupted probe AUC. Values average the two recorded hook sites; the generated table comment records the maximum absolute hook-site difference.}
\end{table}

\begin{table}[!htbp]
\centering
\label{tab:iscorrupted_pretrained}
\resizebox{\columnwidth}{!}{\begin{tabular}{lccccc}
\toprule
Layer & gemma-3-1b-it & OLMo-2-1124-7B-Instruct & Meta-Llama-3-8B-Instruct & Qwen3-32B & gemma-4-31B-it \\
\midrule
Layer 0 & 0.952 (0.037) & 0.969 (0.023) & 0.965 (0.028) & 0.966 (0.021) & 0.969 (0.020) \\
Layer 5 & 0.978 (0.018) & 0.994 (0.008) & 0.994 (0.008) & 0.990 (0.010) & 0.994 (0.008) \\
Layer 10 & 0.981 (0.017) & 0.997 (0.005) & 0.997 (0.006) & 0.995 (0.006) & 0.995 (0.006) \\
Layer 15 & 0.987 (0.015) & 0.996 (0.005) & 0.995 (0.007) & 0.997 (0.003) & 0.996 (0.005) \\
Layer 20 & 0.987 (0.013) & 0.995 (0.007) & 0.996 (0.006) & 0.998 (0.003) & 0.997 (0.005) \\
Layer 25 & 0.986 (0.016) & 0.993 (0.009) & 0.995 (0.008) & 0.998 (0.003) & 0.997 (0.006) \\
Layer 30 & -- & 0.992 (0.010) & 0.995 (0.008) & 0.997 (0.004) & 0.996 (0.004) \\
Layer 31 & -- & 0.991 (0.011) & 0.994 (0.009) & -- & -- \\
Layer 35 & -- & -- & -- & 0.998 (0.003) & 0.995 (0.005) \\
Layer 40 & -- & -- & -- & 0.998 (0.003) & 0.995 (0.004) \\
Layer 45 & -- & -- & -- & 0.998 (0.003) & 0.995 (0.005) \\
Layer 50 & -- & -- & -- & 0.998 (0.003) & 0.996 (0.005) \\
Layer 55 & -- & -- & -- & 0.998 (0.003) & 0.995 (0.006) \\
Layer 59 & -- & -- & -- & -- & 0.993 (0.007) \\
Layer 60 & -- & -- & -- & 0.996 (0.006) & -- \\
Layer 63 & -- & -- & -- & 0.995 (0.007) & -- \\
\bottomrule
\end{tabular}
}
\caption{Pretrained is-corrupted probe AUC on ARC, compressed across corruption modes. Rows are probed layers, columns are models, and each cell reports mean AUC with standard deviation in parentheses over distractor, replacement, spelling, and dropout.}
\end{table}

\begin{table}[!htbp]
\centering
\label{tab:rmse_slope}
\resizebox{0.7\columnwidth}{!}{\begin{tabular}{lcc}
\toprule
Model & Repaired slope ($R^2$) & Failed slope ($R^2$) \\
\midrule
OLMo-2-1124-7B-Instruct & 0.064 (0.73) & 0.219 (0.72) \\
Meta-Llama-3-8B-Instruct & 0.051 (0.90) & 0.221 (0.86) \\
Qwen3-32B & 0.049 (0.98) & 0.188 (0.98) \\
gemma-4-31B-it & 0.021 (1.00) & 0.186 (1.00) \\
\bottomrule
\end{tabular}
}
\caption{Linearization residual diagnostics by model and task family. The slope is measured from residual norm to linearization RMSE.}
\end{table}
\FloatBarrier

\section{Additional figures}
\label{app:additional_figures}

\begin{figure}[htbp!]
    \centering
    \includegraphics[width=0.85\columnwidth]{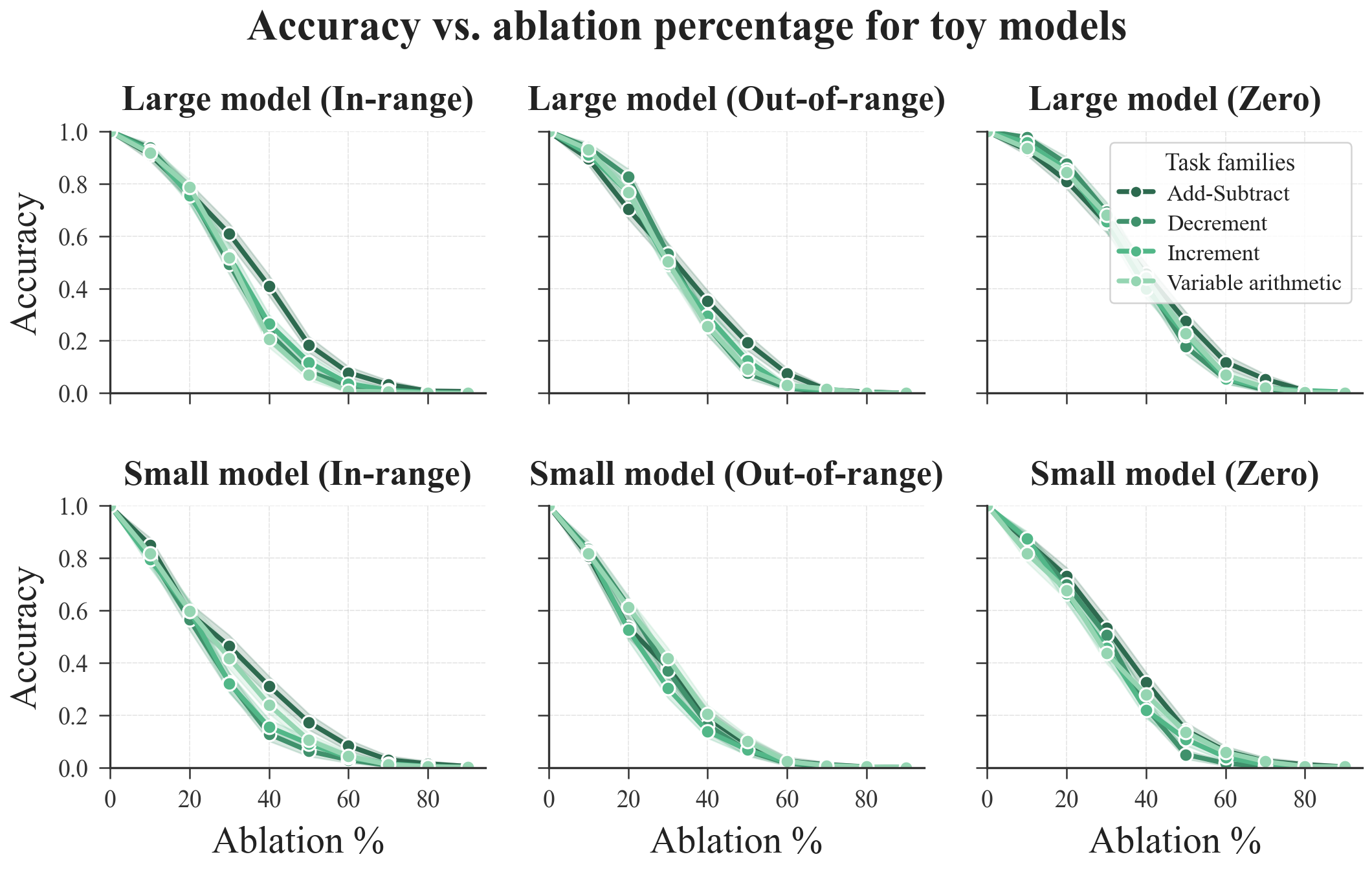}
    \caption{Accuracy versus ablation percentage across toy model sizes, corruption modes and task families.}
    \label{fig:toy_accuracy_degradation}
    \end{figure}
    
\begin{figure}[htbp!]
\centering
\includegraphics[width=0.45\columnwidth]{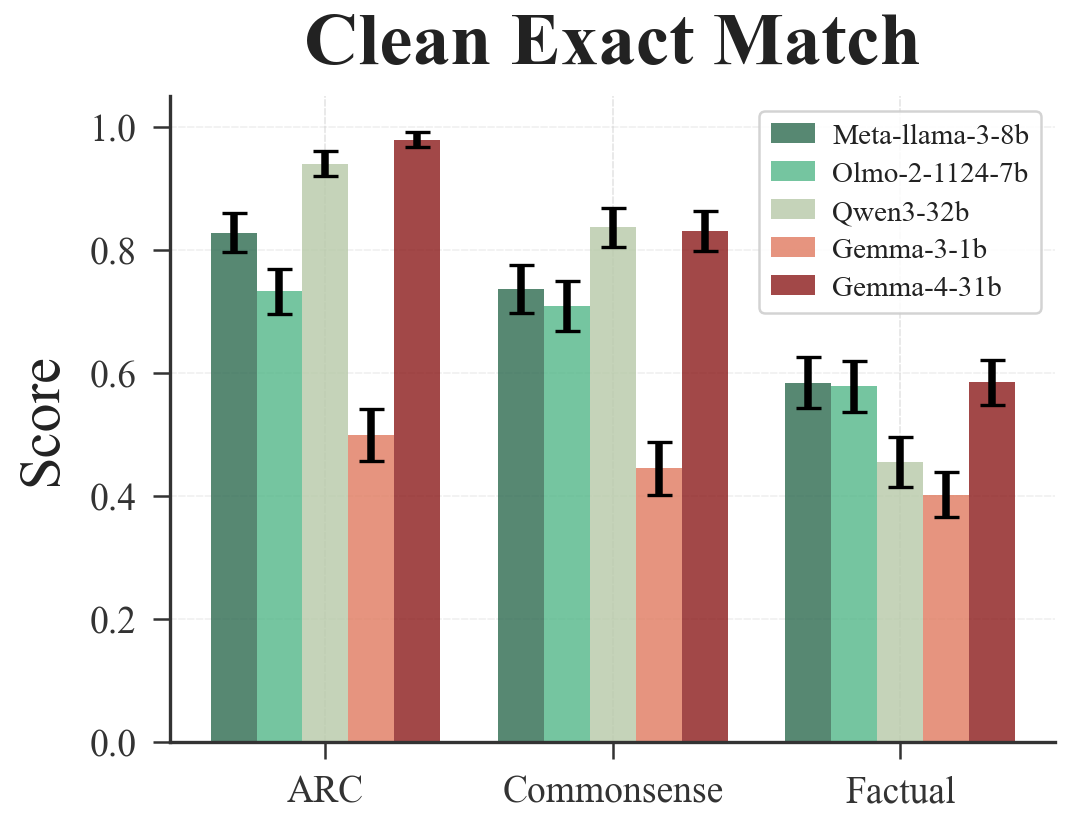}
\caption{Clean baseline exact-match performance of the pretrained models across the factual QA, ARC, and CommonsenseQA task families before any corruption is applied.}
\label{nlpfig:clean_baseline}
\end{figure}

\begin{figure}[htbp!]
\centering
\includegraphics[width=0.65\columnwidth]{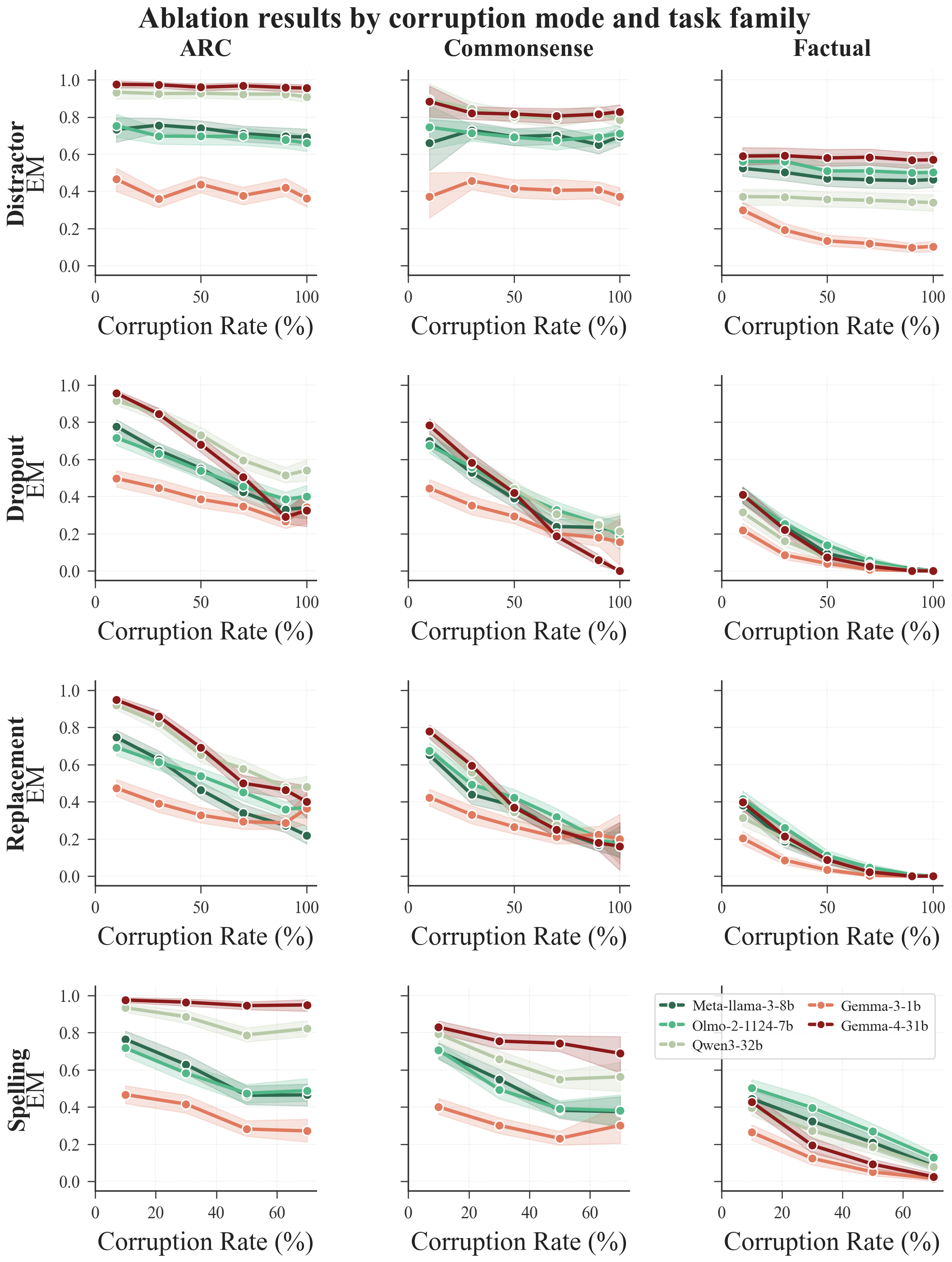}
\caption{Representative degradation curves for pretrained models. Exact match (EM) is plotted versus ablation rate across task families and model sizes, illustrating that dropout and replacement are generally the most damaging corruptions while distractor-style corruption is typically milder.}
\label{nlpfig:degradation_representative}
\end{figure}

\begin{figure}[t]
\centering
\includegraphics[width=0.85\columnwidth]{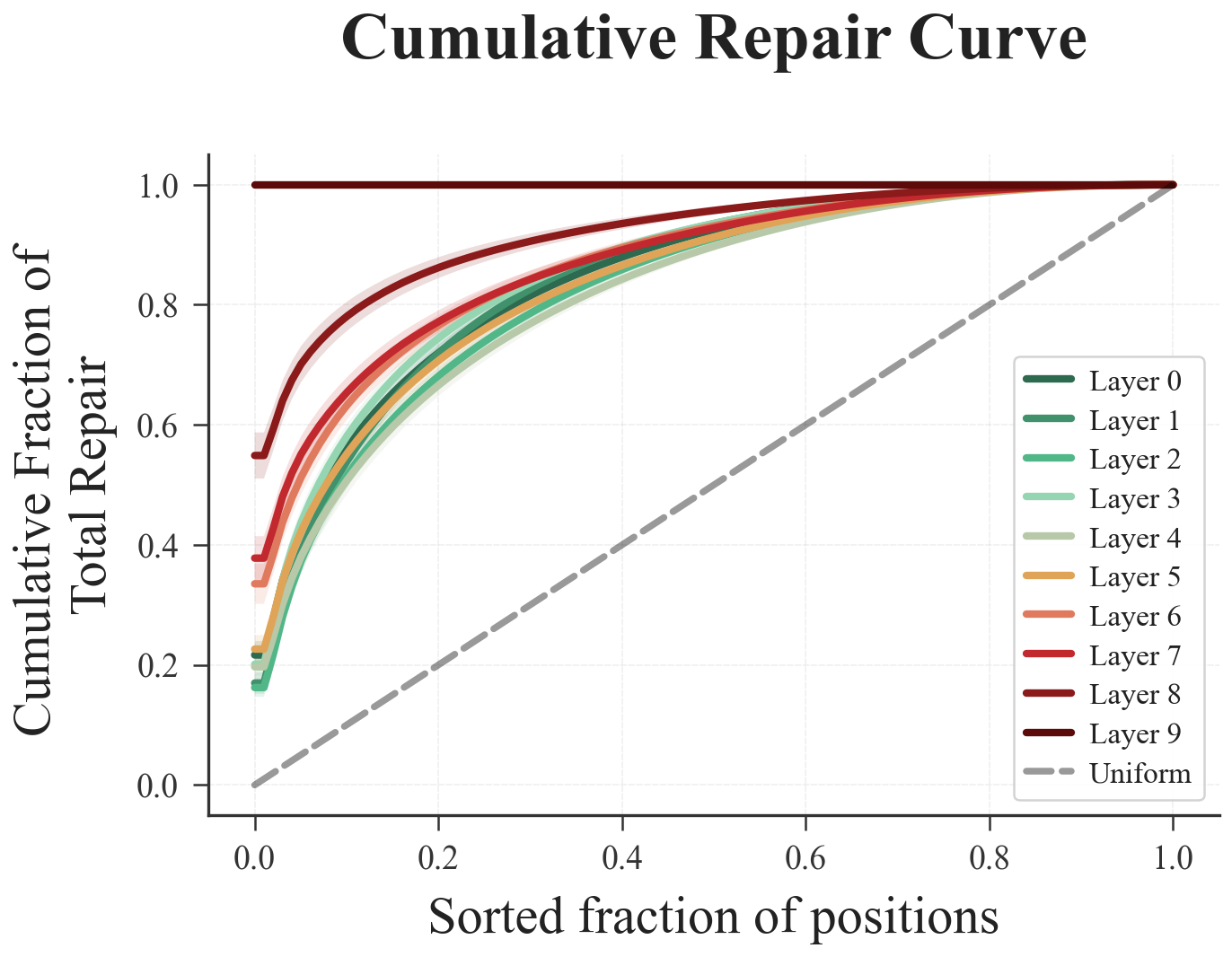}
\caption{Cumulative repair concentration in the large toy model for sequence length 80, the zero corruption mode, and 30\% ablation. Positions are sorted by their per-position patching effect before the residual addition; the steep initial rise shows that a small subset of positions carries most of the total repair signal.}
\label{fig:toy_patching_cumulative}
\end{figure}

\begin{figure}[t]
\centering
\includegraphics[width=0.95\columnwidth]{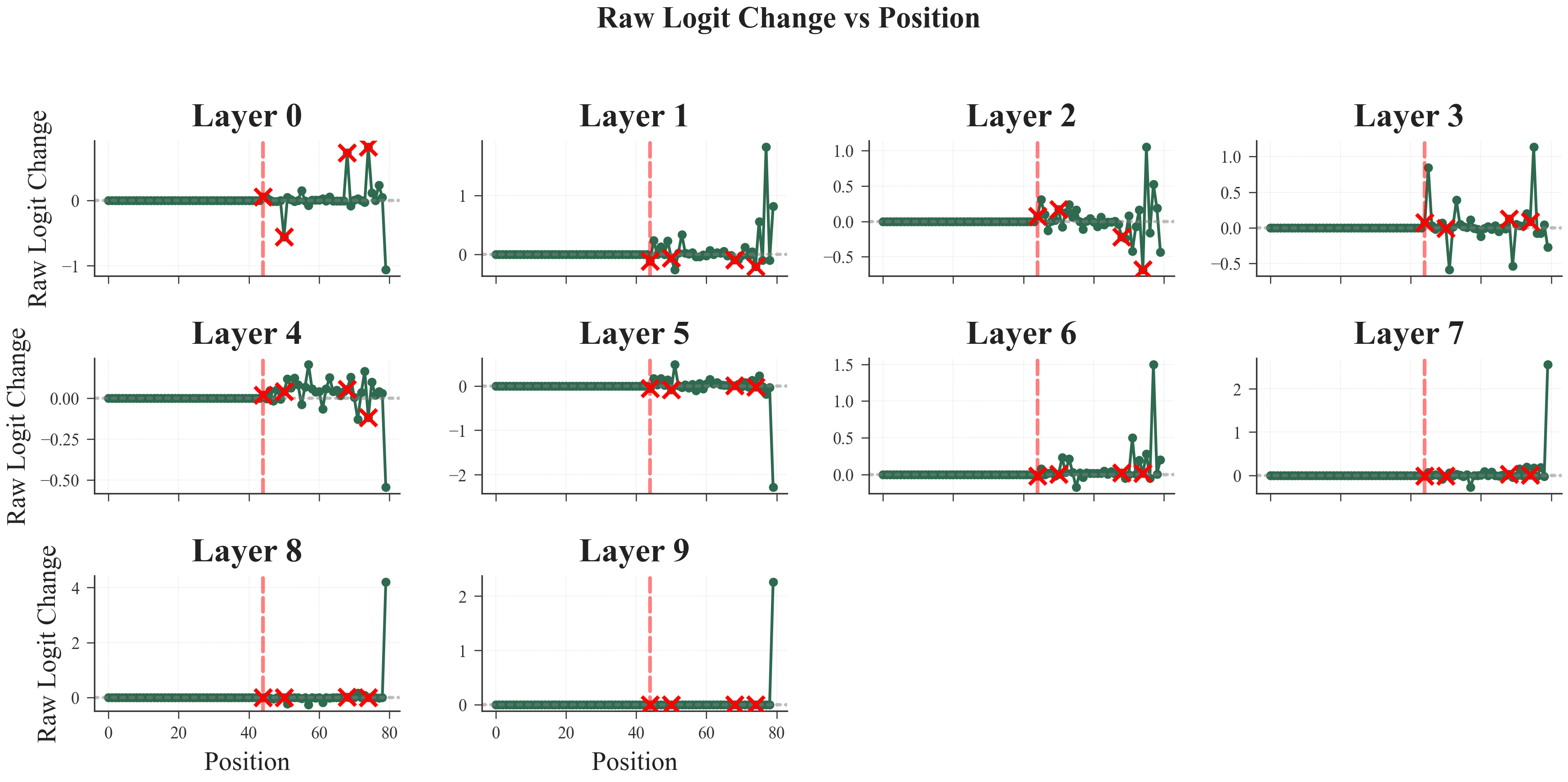}
\caption{Example-level per-position patching map for the large toy model before the residual addition. Each panel shows the raw logit change across positions at a different layer, illustrating how early effects stay near corrupted locations and later effects spread toward output-relevant positions.}
\label{fig:toy_patching_position_map}
\end{figure}

\begin{figure}[t]
\centering
\includegraphics[width=0.95\columnwidth]{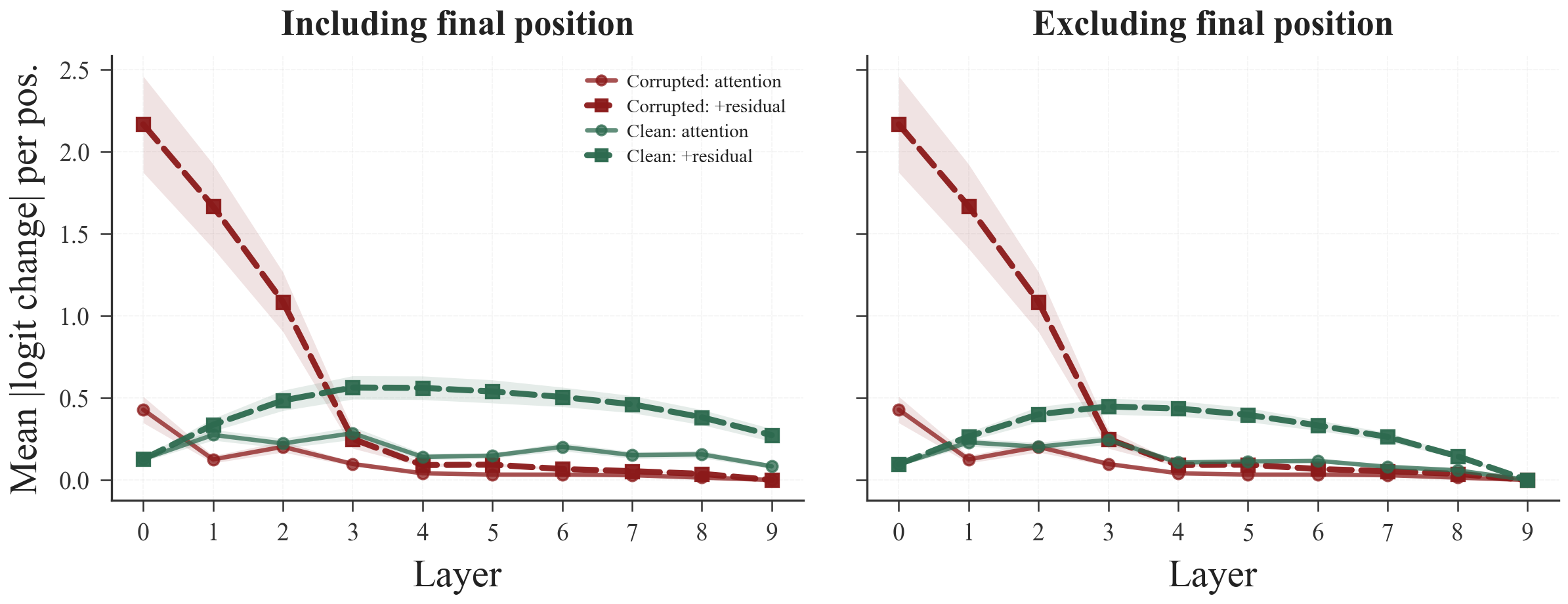}
\caption{Repair signal with and without the final position. Mean absolute logit change from patching attention outputs at corrupted and clean positions across layers. Left: all eligible positions, including the final sequence position. Right: the final position is excluded.
}
\label{fig:toy_attention_residual_stack}
\end{figure}

\begin{figure}[htbp!]
\centering

\includegraphics[width=0.95\columnwidth]{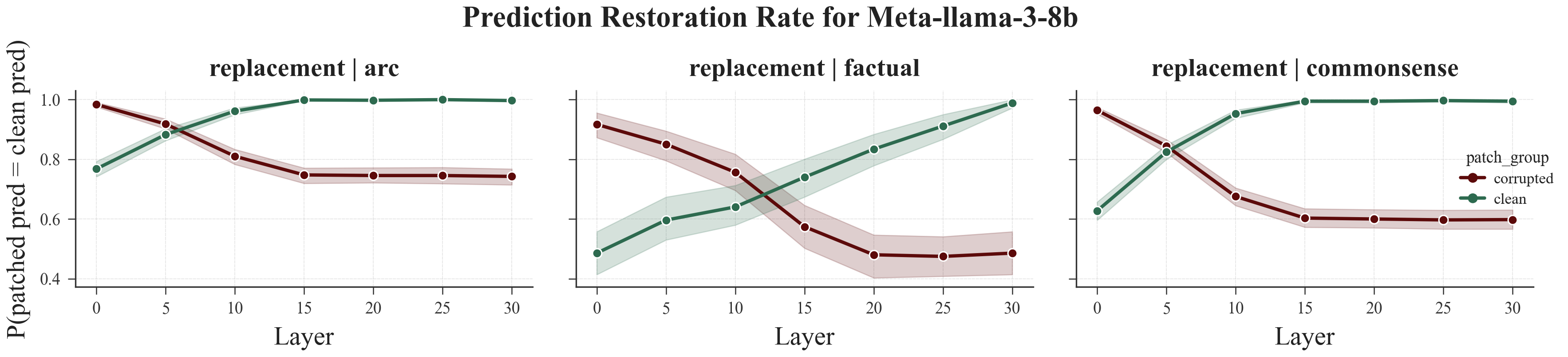}

\vspace{0.5em}

\includegraphics[width=0.95\columnwidth]{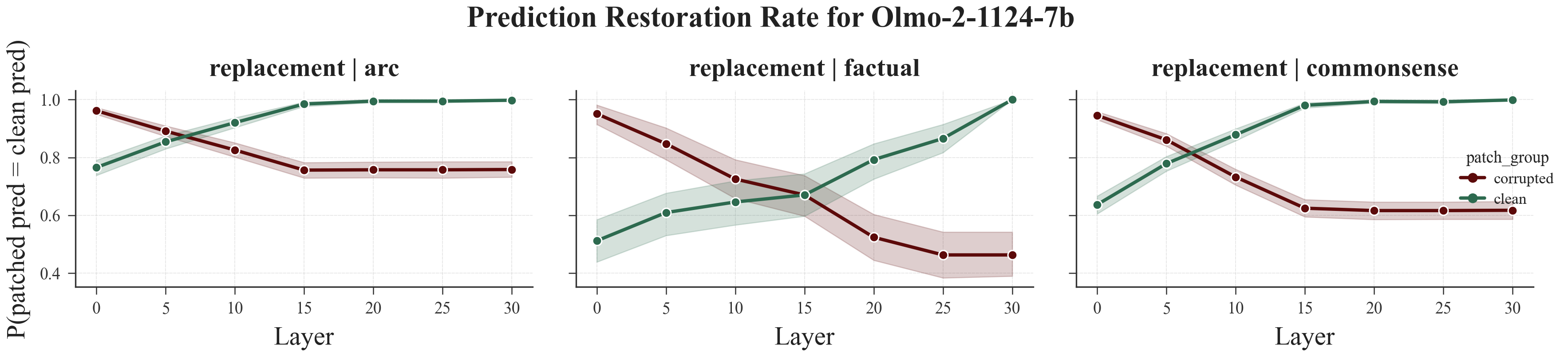}

\vspace{0.5em}

\includegraphics[width=0.95\columnwidth]{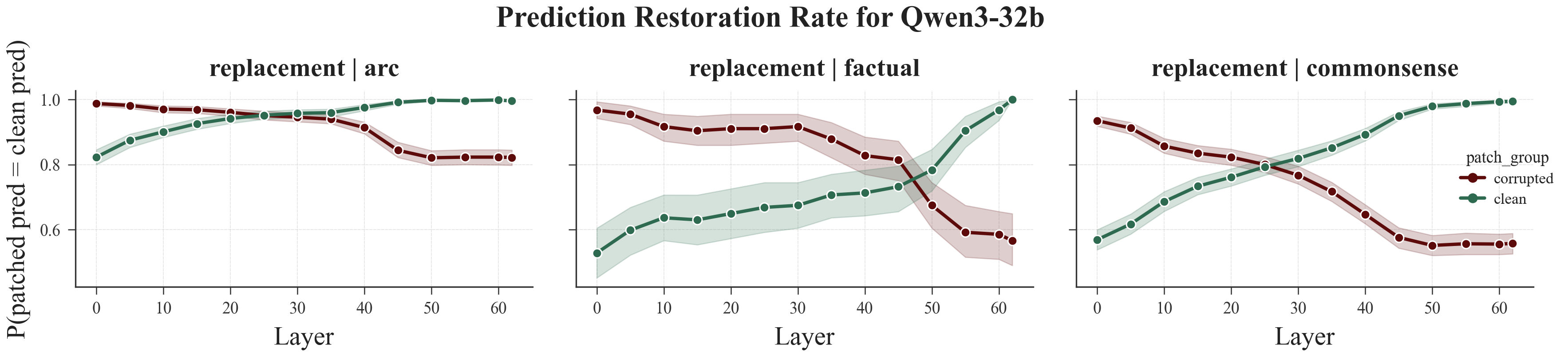}

\vspace{0.5em}

\includegraphics[width=0.95\columnwidth]{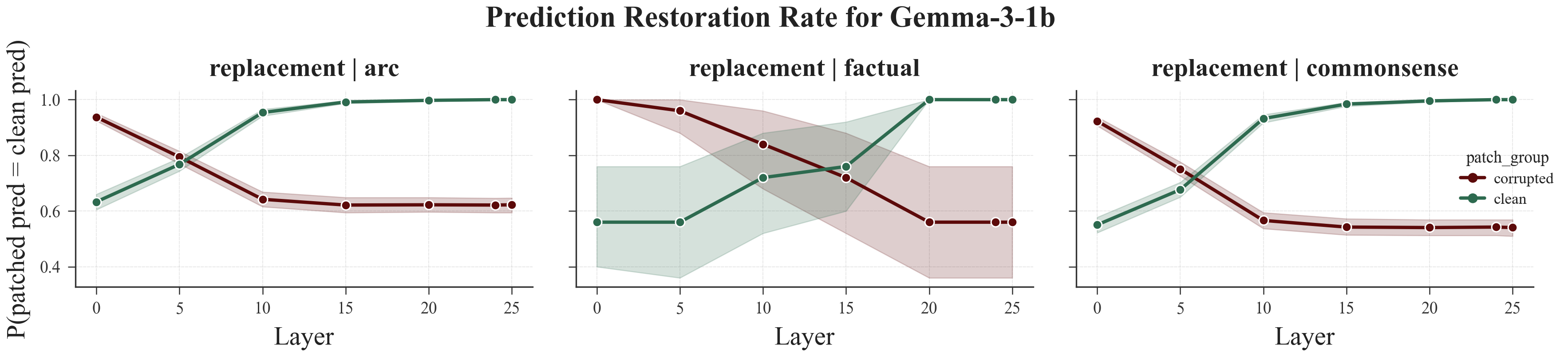}

\vspace{0.5em}

\includegraphics[width=0.95\columnwidth]{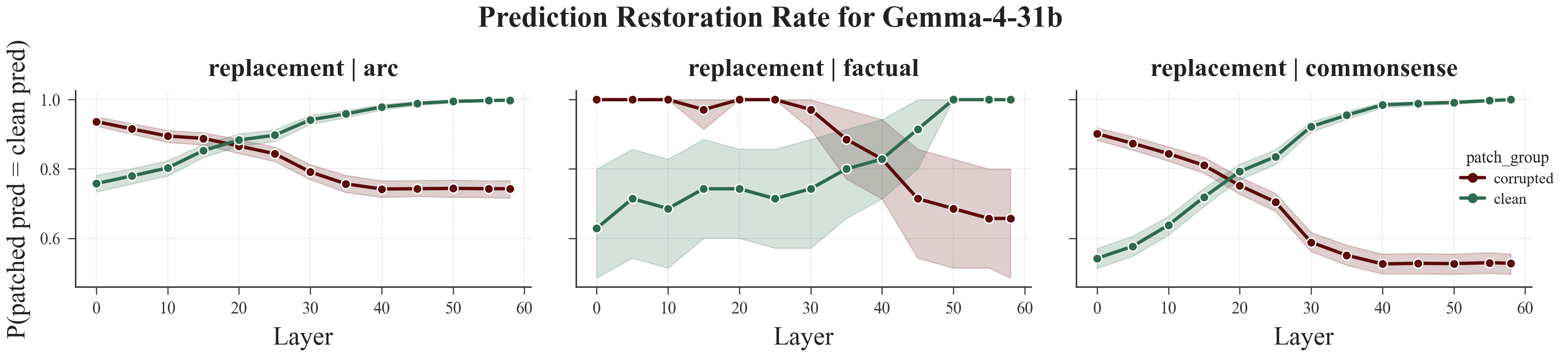}

\caption{Prediction restoration under position-group patching in pretrained models. This reports the probability that patching corrupted or clean positions restores the clean prediction, for each model across corruption modes and task families.}
\label{nlpfig:prediction_restoration_all_models}
\end{figure}

\begin{figure}[t]
\centering
\includegraphics[width=0.95\columnwidth]{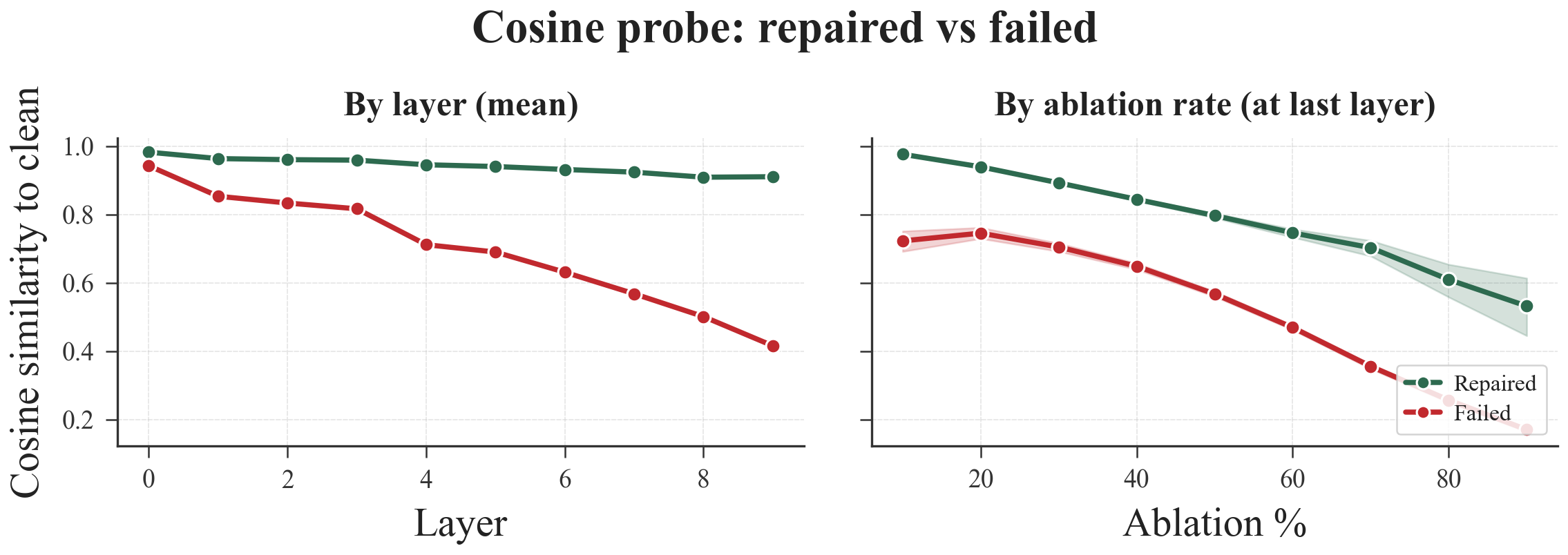}
\caption{Cosine similarity between clean and corrupted activations (hooked after LayerNorm block) in the large toy model. Repaired examples remain closely aligned across depth and as corruption increases, whereas failed examples drift away, especially at the final layer.}
\label{fig:toy_cosine_by_outcome}
\end{figure}

\begin{figure}[htbp!]
\centering

\includegraphics[width=0.95\columnwidth]{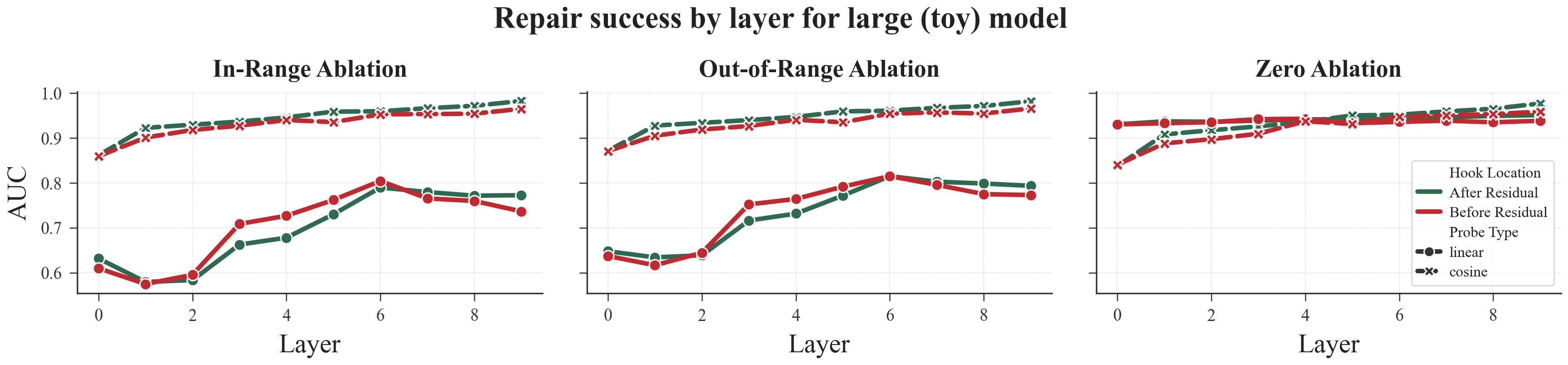}

\vspace{0.5em}

\includegraphics[width=0.95\columnwidth]{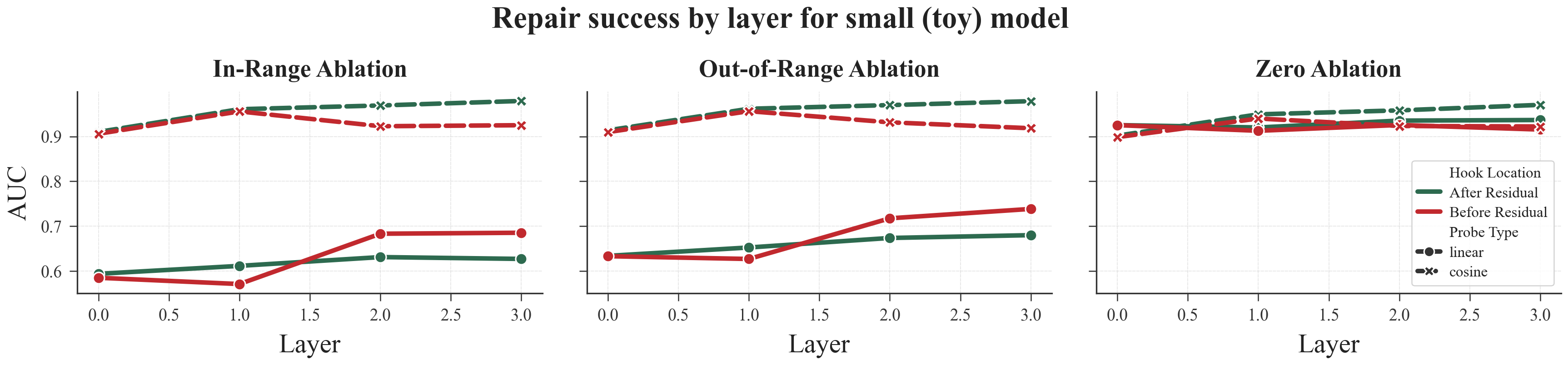}

\caption{Repair-success probes in toy models. This shows layer-wise AUC for predicting whether a corrupted example is ultimately answered correctly in the large and small models, respectively, comparing linear and cosine readouts across zero, in-range, and out-of-range ablation modes.}
\label{fig:toy_repair_probe_all}
\end{figure}
\begin{figure}[htbp!]
\centering

\includegraphics[width=0.95\columnwidth]{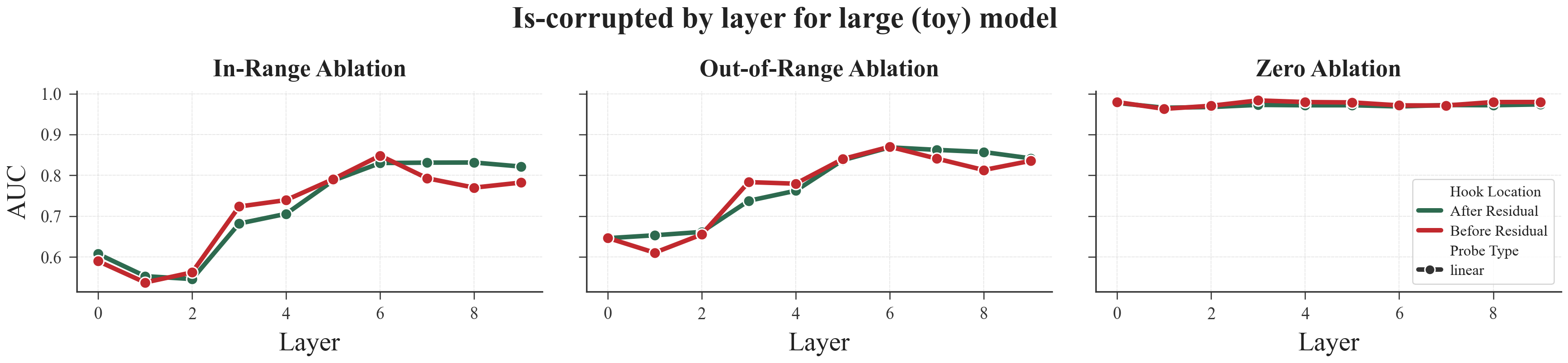}

\vspace{0.5em}

\includegraphics[width=0.95\columnwidth]{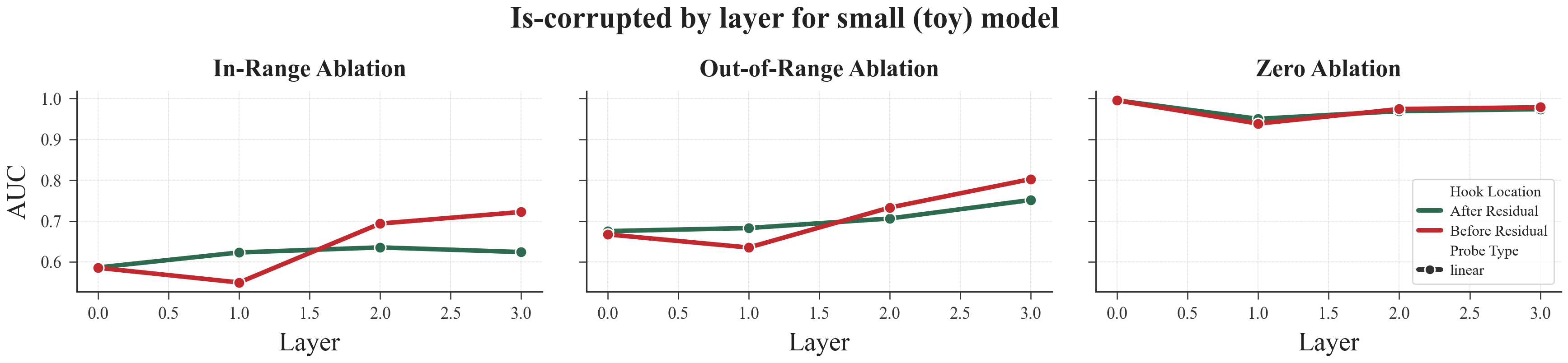}

\caption{Corruption-detection probes in toy models. This shows layer-wise AUC for predicting whether an input was corrupted in the large and small models, respectively; unlike repair-success prediction, corruption identity becomes strongly linearly decodable only after several layers.}
\label{fig:toy_corruption_probe_all}
\end{figure}

\begin{figure}[htbp!]
\centering
\includegraphics[width=0.95\columnwidth]{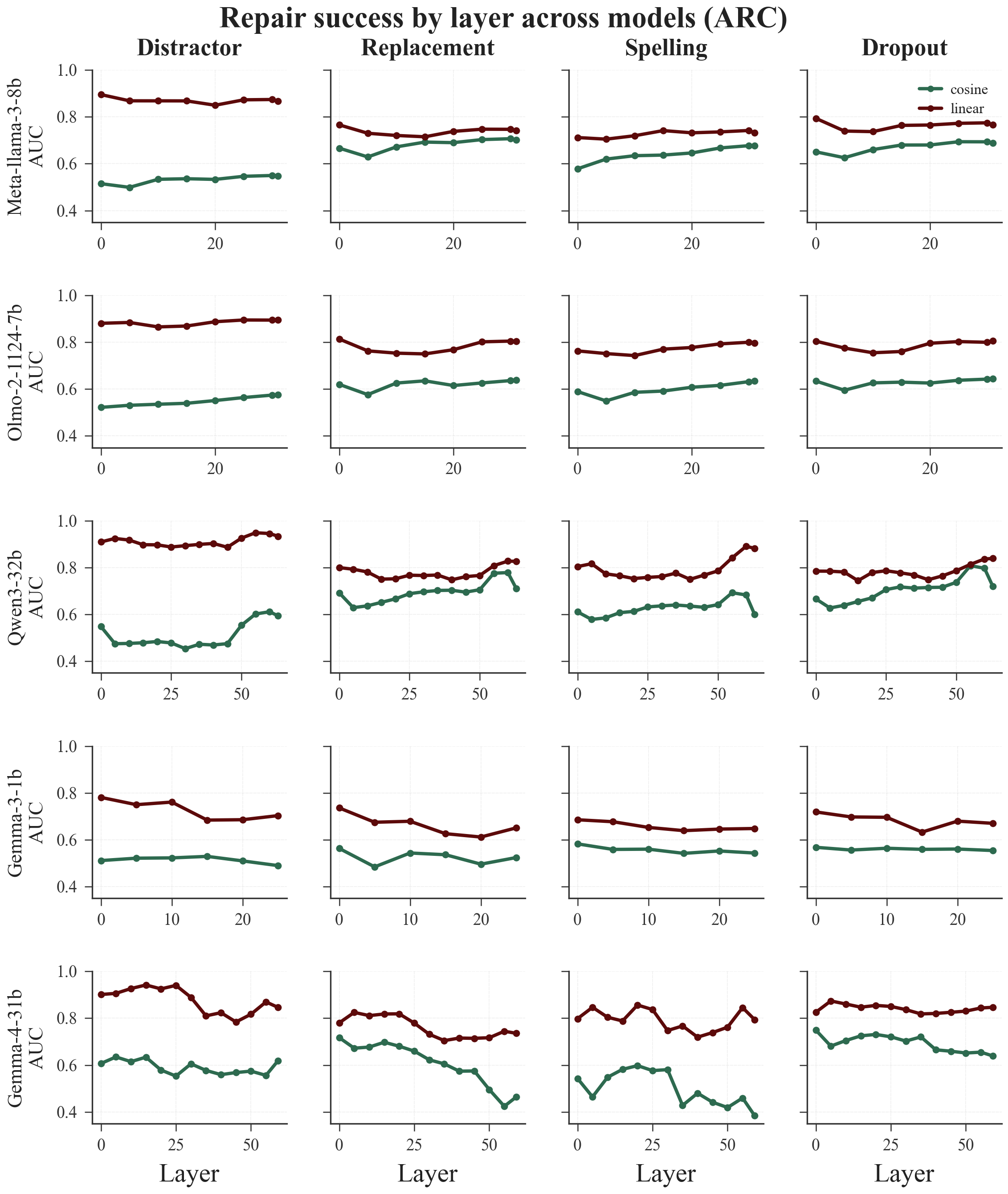}
\caption{Representative repair-success probes in pretrained models. Layer-wise AUC predicts whether a corrupted example will still be answered correctly, comparing linear and cosine probes across corruption modes for ARC.}
\label{nlpfig:repair_probe_representative}
\end{figure}

\begin{figure}[t]
    \centering
    \includegraphics[width=0.75\columnwidth]{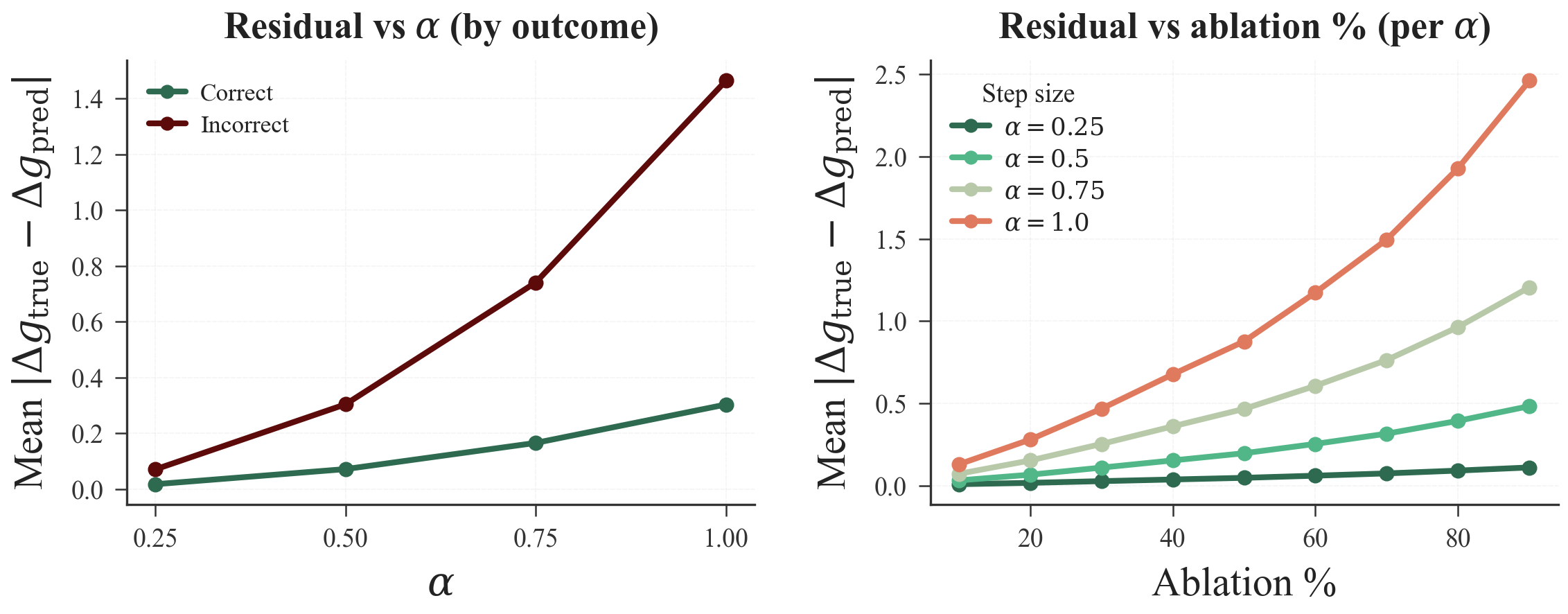}
    \caption{Linearization residuals in the large toy model. Repaired examples stay close to locally linear behavior near the clean trajectory, while failed examples show much larger residuals at larger interpolation steps and higher corruption levels.}
    \label{fig:linear}
\end{figure}
\begin{figure}[htbp!]
\centering

\includegraphics[width=0.95\columnwidth]{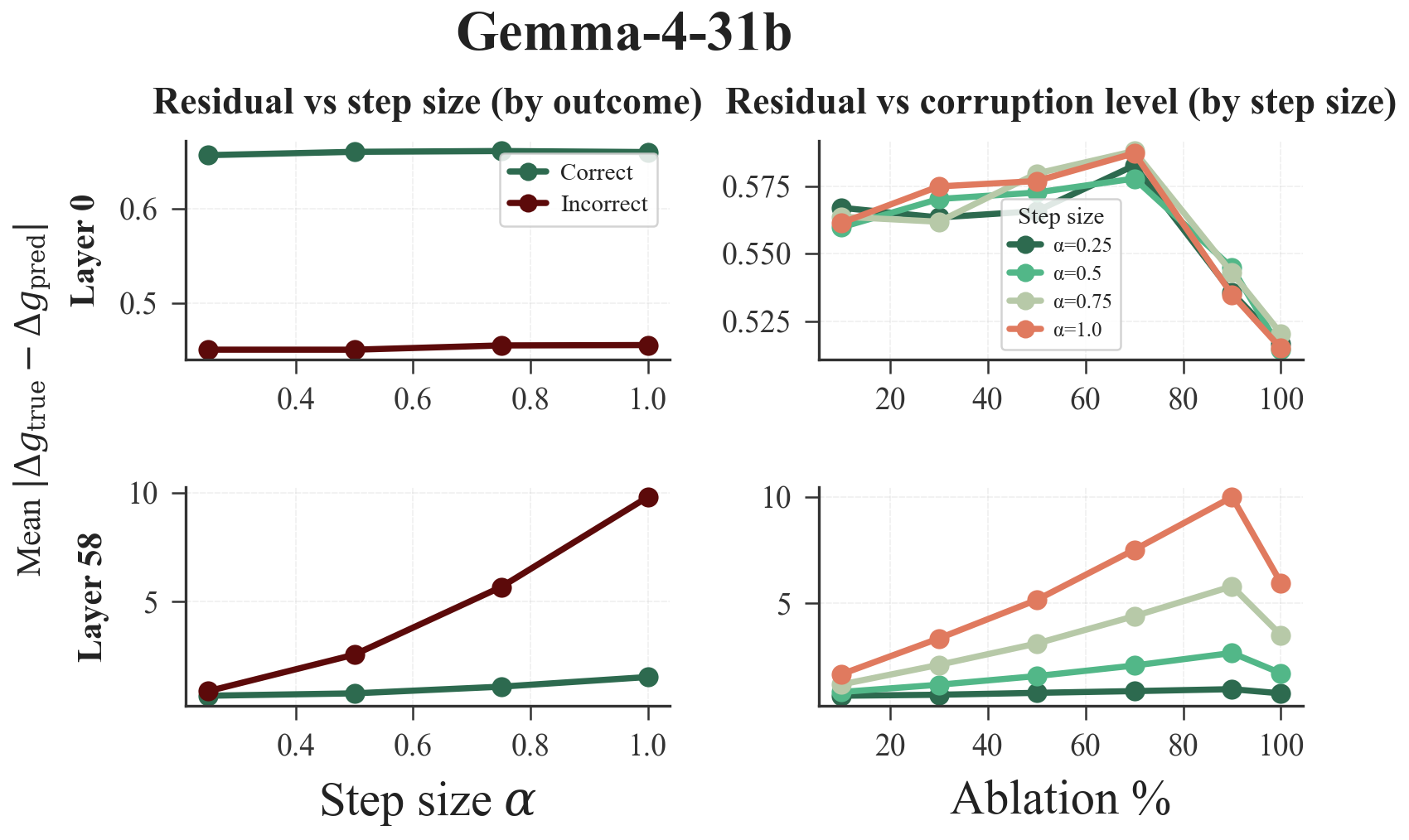}

\vspace{0.5em}

\includegraphics[width=0.95\columnwidth]{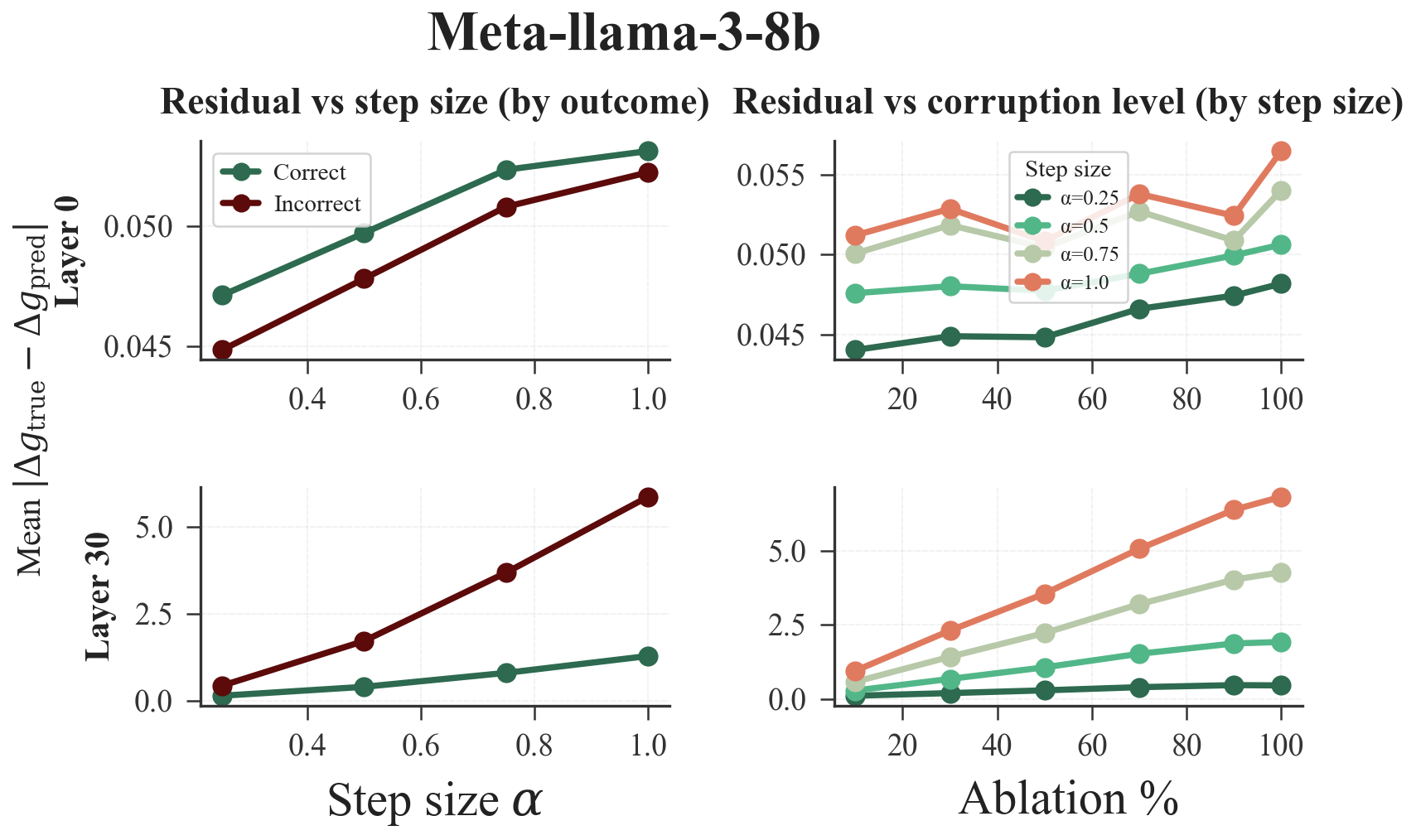}

\caption{Linearization residuals in pretrained models. This shows mean absolute linearization error as a function of interpolation step size and corruption level; repaired and failed examples separate most clearly at large $\alpha$ and in later layers.}
\label{nlpfig:linearization_early_models}
\label{fig:linearization_nlp}
\end{figure}

\begin{figure}[htbp!]
\centering

\includegraphics[width=0.95\columnwidth]{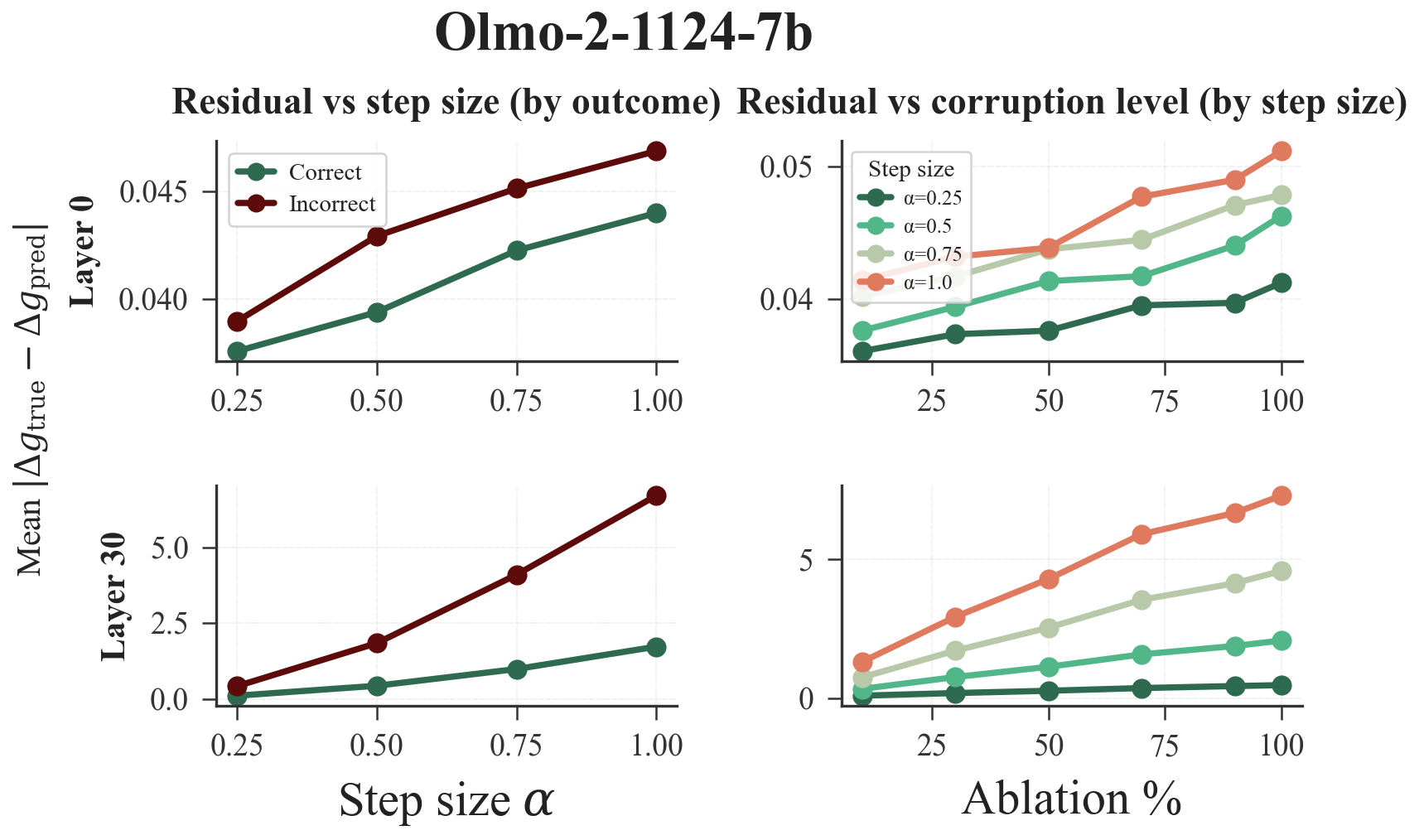}

\vspace{0.5em}

\includegraphics[width=0.95\columnwidth]{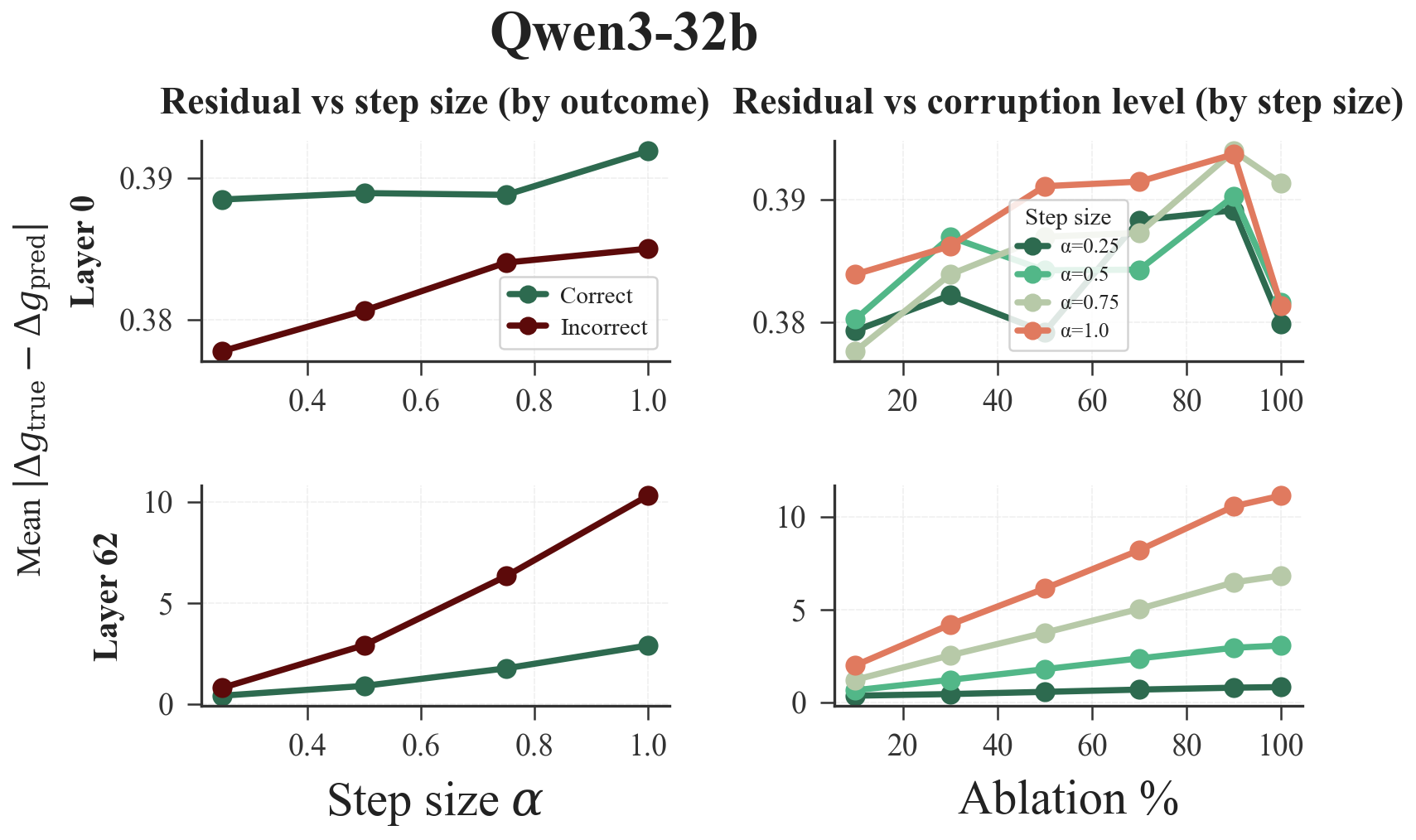}

\caption{Additional pretrained-model linearization residuals. This shows mean absolute linearization error as a function of interpolation step size and corruption level for the remaining evaluated models.}
\label{nlpfig:linearization_late_models}
\end{figure}

\begin{figure}[htbp!]
\centering

\includegraphics[width=0.95\columnwidth]{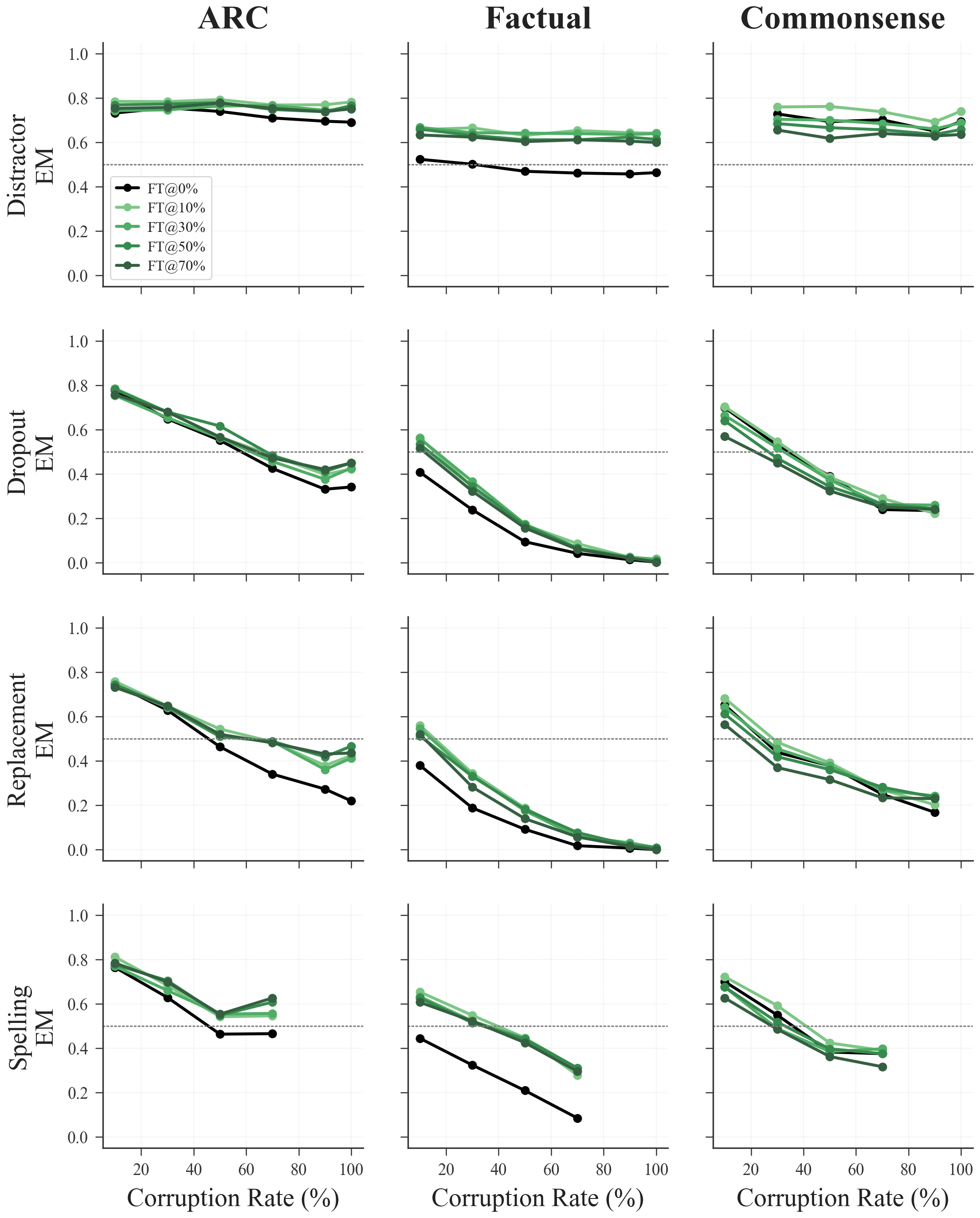}

\caption{Finetuning on corrupted inputs differentially improves the robustness of an LLM.}
\label{nlpfig:linearization_early_models}
\label{fig:linearization_nlp}
\end{figure}

\begin{figure}[htbp!]
\centering

\includegraphics[width=0.95\columnwidth]{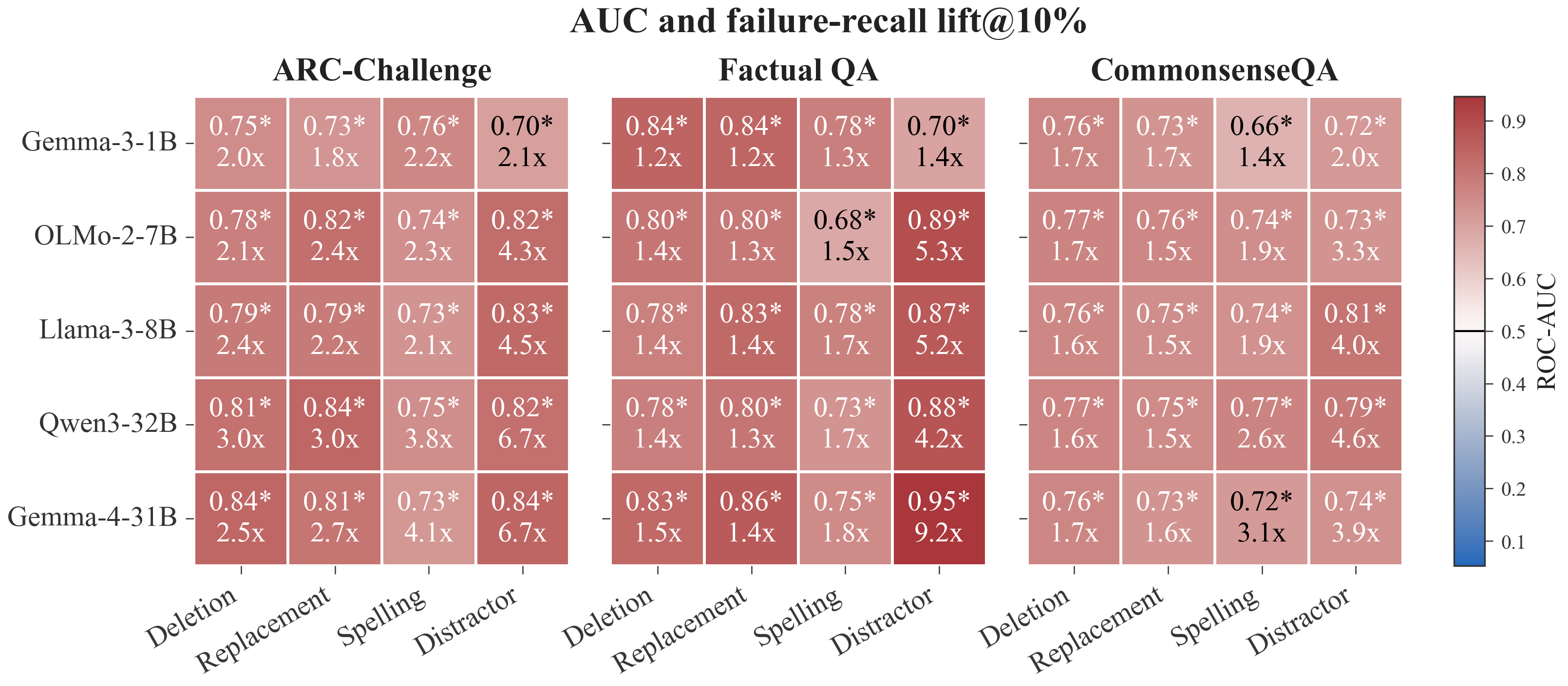}

\caption{Earliest-layer probe performance by model and corruption mode. Cells show ROC-AUC and 10\%-budget recall lift; asterik indicates bootstrap 95\% CI above chance.}
\label{nlpfig:deployable_probe_appendix}
\end{figure}

\begin{figure}[t]
\centering
\includegraphics[width=0.65\columnwidth]{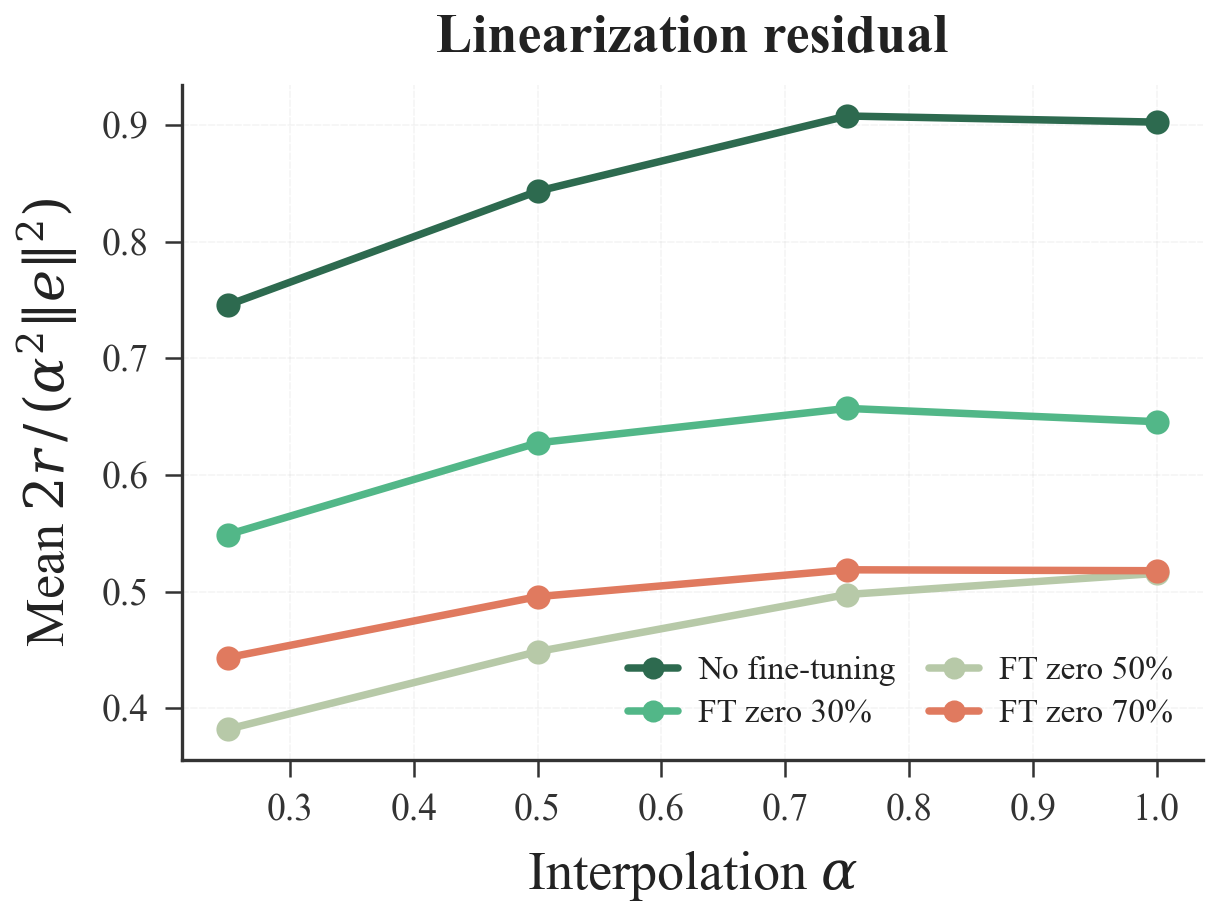}
\caption{Fine-tuning reduces displacement-normalized linearization error across interpolation steps, with the largest reduction at moderate corruption levels.
}
\label{fig:linearization_ft}
\end{figure}

\end{document}